\documentclass[10pt,journal,compsoc]{IEEEtran}
\usepackage{graphicx}
\usepackage{subfigure}
\usepackage{amsmath}
\usepackage{amssymb}
\usepackage{tabularx}
\usepackage{multirow}
\usepackage{color}
\usepackage{colortbl}
\usepackage{epstopdf}

\definecolor{dbcolor}{rgb}{0,0,1}

\definecolor{srcolor}{rgb}{1,0,0}

\newcommand{\figref}[1]{Fig. \ref{#1}}
\newcommand{\tabref}[1]{Table \ref{#1}}
\newcommand{\equref}[1]{(\ref{#1})}
\newcommand{\secref}[1]{Sec. \ref{#1}}

\definecolor{red}{rgb}{1,0,0}

\definecolor{green}{rgb}{0,1,0}

\definecolor{blue}{rgb}{0,0,1}

\makeatletter
\def\hlinewd#1{%
\noalign{\ifnum0=`}\fi\hrule \@height #1 \futurelet
\reserved@a\@xhline}

\ifCLASSOPTIONcompsoc
  \usepackage[nocompress]{cite}
\else
  \usepackage{cite}
\fi

\ifCLASSINFOpdf

\else

\fi

\renewcommand{\IEEEQED}{\IEEEQEDclosed}

\begin{document}
\bstctlcite{IEEEexample:BSTcontrol}

\title{DASC: Robust Dense Descriptor for Multi-modal and Multi-spectral Correspondence Estimation}

\author{Seungryong~Kim,~\IEEEmembership{Student Member,~IEEE,}
        Dongbo~Min,~\IEEEmembership{Senior Member,~IEEE,}
        Bumsub~Ham,~\IEEEmembership{Member,~IEEE,}
        Minh~N.~Do,~\IEEEmembership{Fellow,~IEEE,}
        and~Kwanghoon~Sohn,~\IEEEmembership{Senior Member,~IEEE}
\IEEEcompsocitemizethanks{\IEEEcompsocthanksitem S. Kim and K. Sohn are with the School of Electrical and Electronic Engineering, Yonsei University, Seoul 120-749, Korea. \protect\\
E-mail: \{srkim89, khsohn\}@yonsei.ac.kr
\IEEEcompsocthanksitem D. Min is with the Department of Computer Science and Engineering,
Chungnam National University, Daejeon 305-764, Korea. \protect\\
E-mail: dbmin@cnu.ac.kr
\IEEEcompsocthanksitem B. Ham is with Willow Team, INRIA, Paris 75013, France. \protect\\
E-mail: bumsub.ham@inria.fr \protect
\IEEEcompsocthanksitem M. N. Do is
with the Department of Electrical and Computer Engineering and the
Coordinated Science Laboratory, University of Illinois at
Urbana-Champaign, Urbana, IL 61801 USA. E-mail: minhdo@illinois.edu \protect}}

\IEEEtitleabstractindextext{%
\begin{abstract}
Establishing dense correspondences between multiple images is
a fundamental task in many applications. However, finding a reliable
correspondence in multi-modal or multi-spectral images still remains
unsolved due to their challenging photometric and geometric
variations. In this paper, we propose a novel dense
descriptor, called dense adaptive self-correlation (DASC), to
estimate multi-modal and multi-spectral dense correspondences. Based
on an observation that self-similarity existing within images is
robust to imaging modality variations, we define the descriptor with
a series of an adaptive self-correlation similarity measure between
patches sampled by a randomized receptive field pooling, in which a
sampling pattern is obtained using a discriminative learning. The
computational redundancy of dense descriptors is dramatically
reduced by applying fast edge-aware filtering. 
Furthermore, in order to address geometric variations including scale
and rotation, we propose a geometry-invariant DASC (GI-DASC)
descriptor that effectively leverages the DASC through a
superpixel-based representation. For a quantitative evaluation of
the GI-DASC, we build a novel multi-modal benchmark as varying photometric and geometric
conditions. Experimental results demonstrate the outstanding
performance of the DASC and GI-DASC in many cases of
multi-modal and multi-spectral dense correspondences.
\end{abstract}

}

\maketitle

\IEEEdisplaynontitleabstractindextext
\IEEEpeerreviewmaketitle

\section{Introduction}\label{sec:1}
\IEEEPARstart{R}{ecently}, many computer vision and computational photography
problems have been reformulated to overcome their inherent limitations
by leveraging multi-modal and multi-spectral images. 
Typical examples of other imaging modalities include near-infrared (NIR) image \cite{Brown11,yan13} 
and dark flash image \cite{Krishnan09}. 
More broadly, flash and no-flash images
\cite{Petschnigg04}, blurred images \cite{HaCohen13,Lee13}, and images taken
under different radiometric conditions \cite{Sen12} can also be considered as multi-modal \cite{Shen14}.

\begin{figure}[!t]
\centering
\renewcommand{\thesubfigure}{}
\subfigure[]
{\includegraphics[width=0.25\linewidth]{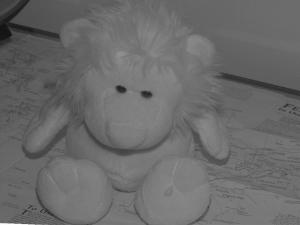}}\hfill
\subfigure[]
{\includegraphics[width=0.25\linewidth]{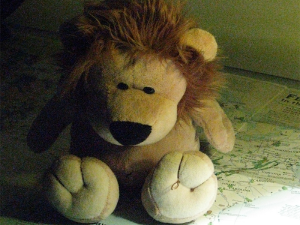}}\hfill
\subfigure[]
{\includegraphics[width=0.25\linewidth]{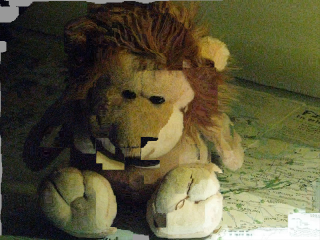}}\hfill
\subfigure[]
{\includegraphics[width=0.25\linewidth]{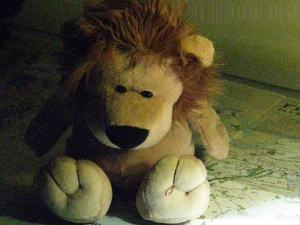}}\hfill
\vspace{-22pt}
\subfigure[]
{\includegraphics[width=0.25\linewidth]{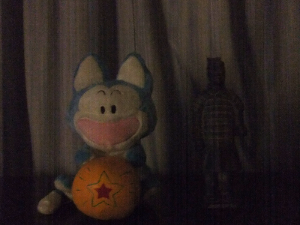}}\hfill
\subfigure[]
{\includegraphics[width=0.25\linewidth]{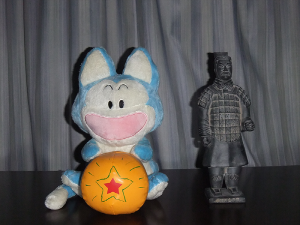}}\hfill
\subfigure[]
{\includegraphics[width=0.25\linewidth]{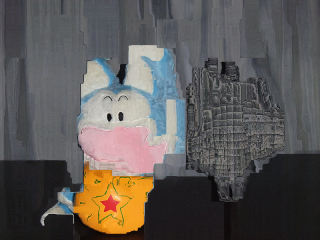}}\hfill
\subfigure[]
{\includegraphics[width=0.25\linewidth]{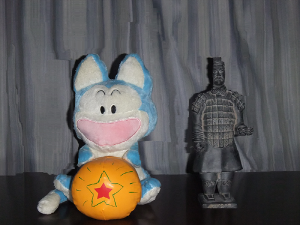}}\hfill
\vspace{-22pt}
\subfigure[]
{\includegraphics[width=0.25\linewidth]{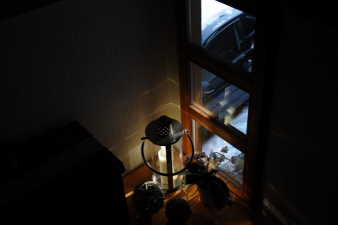}}\hfill
\subfigure[]
{\includegraphics[width=0.25\linewidth]{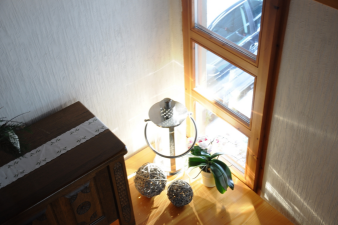}}\hfill
\subfigure[]
{\includegraphics[width=0.25\linewidth]{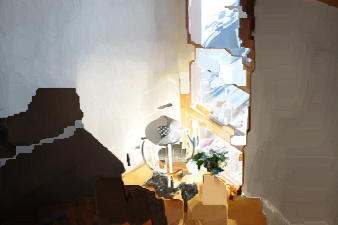}}\hfill
\subfigure[]
{\includegraphics[width=0.25\linewidth]{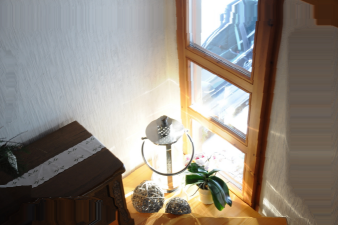}}\hfill
\vspace{-22pt}
\subfigure[(a) Image 1]
{\includegraphics[width=0.25\linewidth]{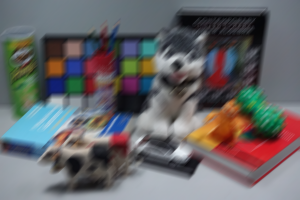}}\hfill
\subfigure[(b) Image 2]
{\includegraphics[width=0.25\linewidth]{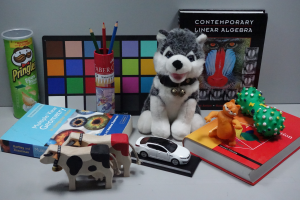}}\hfill
\subfigure[(c) DAISY \cite{Tola10}]
{\includegraphics[width=0.25\linewidth]{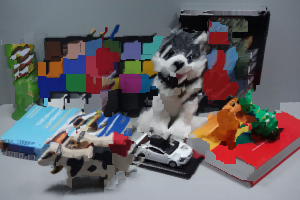}}\hfill
\subfigure[(d) DASC]
{\includegraphics[width=0.25\linewidth]{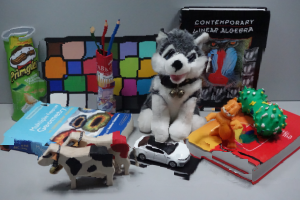}}\hfill
\vspace{-10pt}
\caption{Some challenging multi-modal and
multi-spectral images such as (from top to bottom) RGB-NIR, flash-noflash images, two
images with different exposures, and blur-sharp images. The images
in the third and fourth column are the results obtained by warping images in
the second column to images in the first column with dense
correspondence maps estimated by using DAISY \cite{Tola10} and our DASC descriptor, respectively.}
\label{img:1}\vspace{-10pt}
\end{figure}

\begin{figure}[!t]
\centering
\renewcommand{\thesubfigure}{}
\subfigure[(a) RGB image]
{\includegraphics[width=0.495\linewidth]{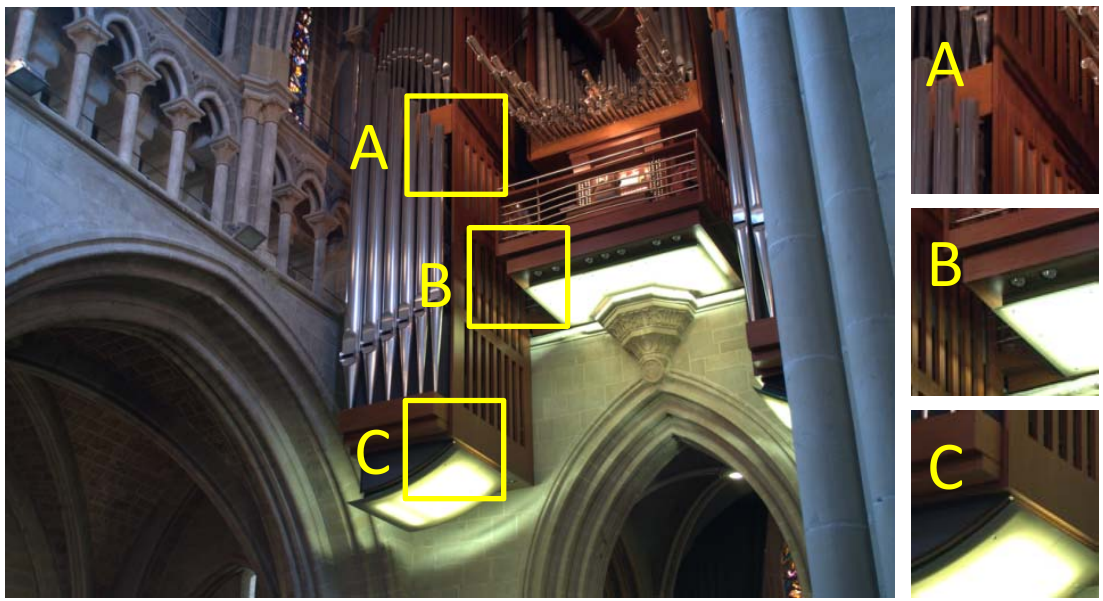}}
\subfigure[(b) NIR image]
{\includegraphics[width=0.495\linewidth]{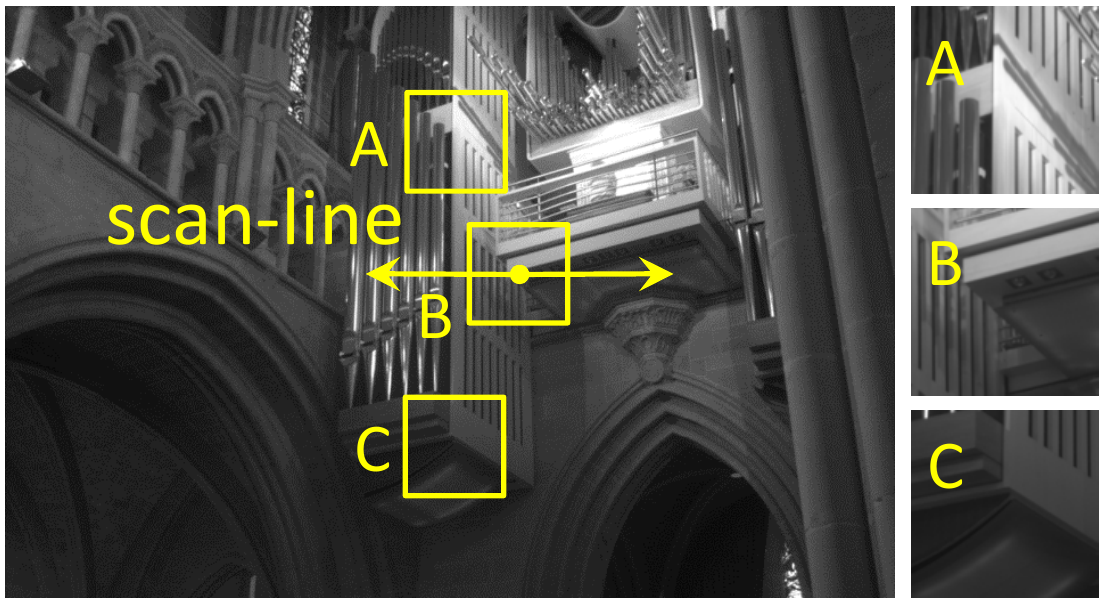}}\\
\vspace{-8pt}
\subfigure[(c) Matching cost in A]
{\includegraphics[width=0.333\linewidth]{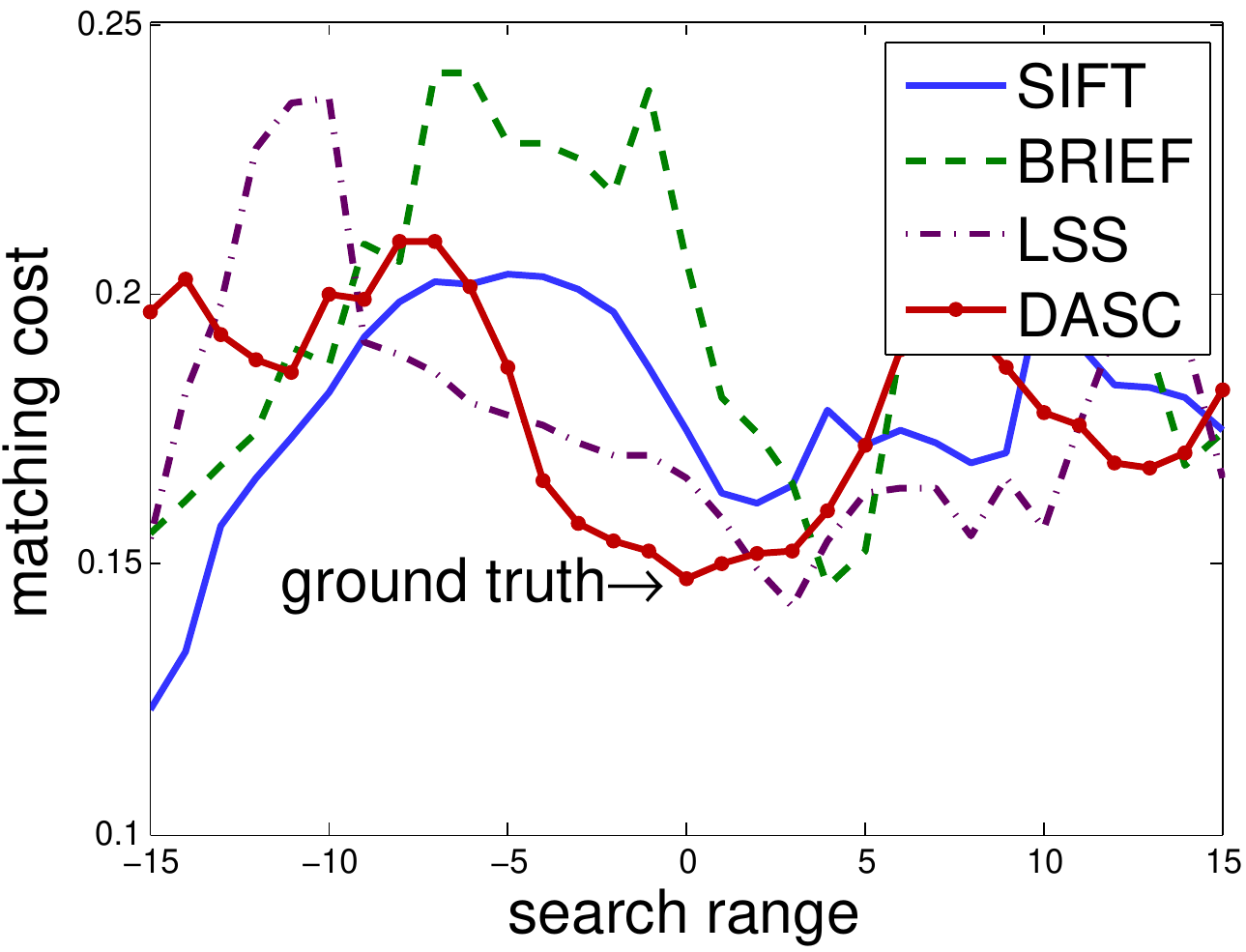}}\hfill
\subfigure[(d) Matching cost in B]
{\includegraphics[width=0.333\linewidth]{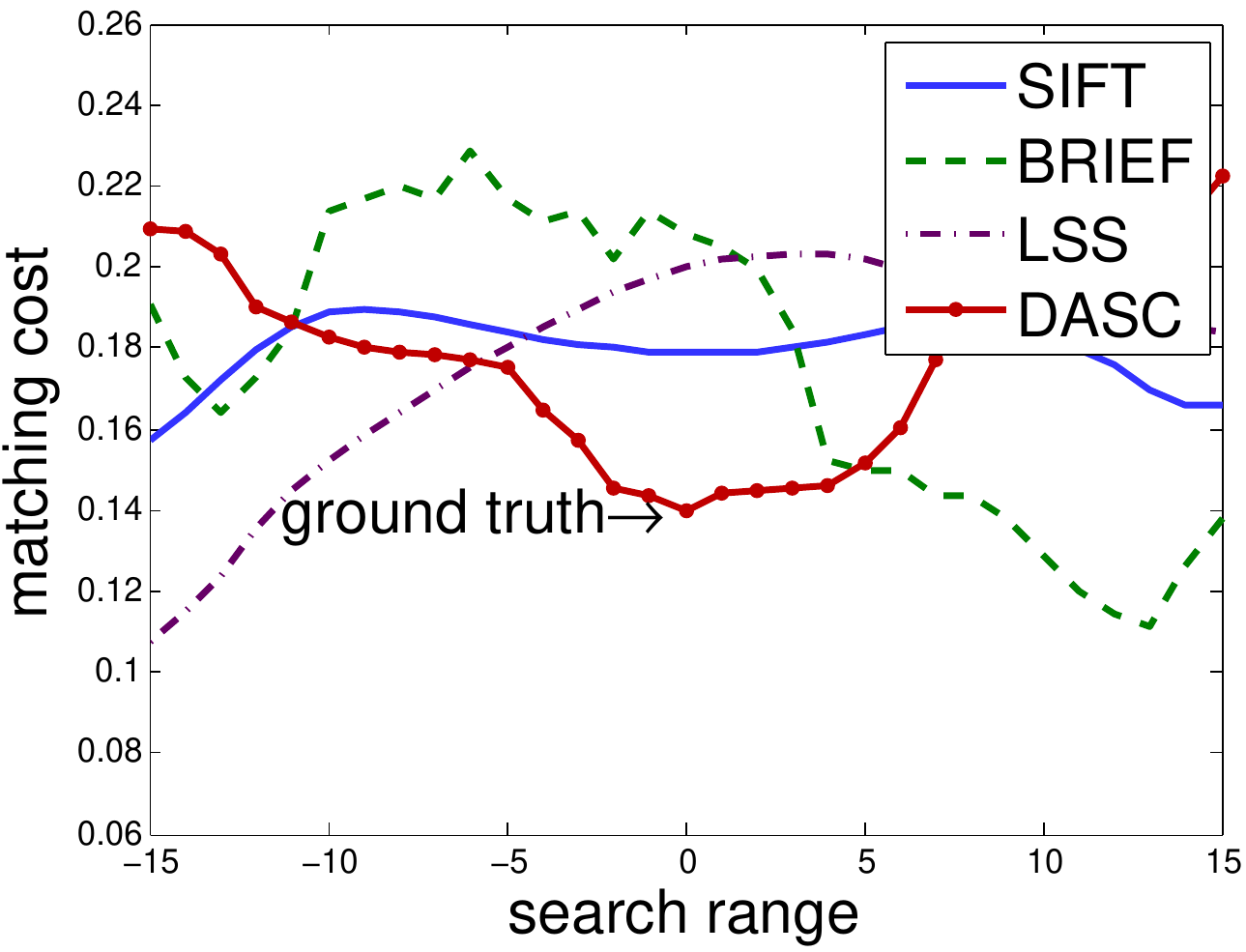}}\hfill
\subfigure[(e) Matching cost in C]
{\includegraphics[width=0.333\linewidth]{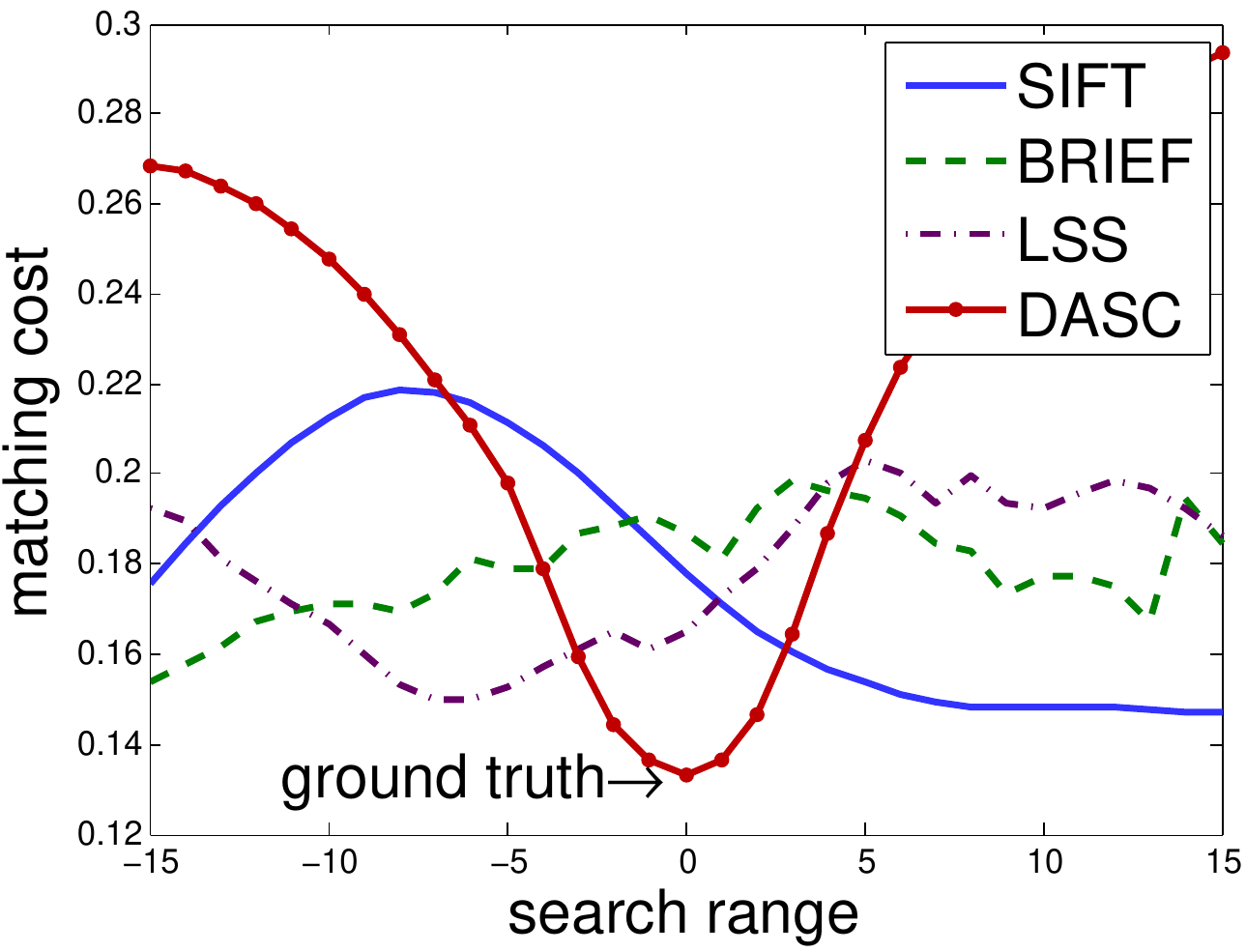}}\hfill
\vspace{-10pt}
\caption{Examples of matching cost comparison. Multi-spectral RGB and NIR images have locally non-linear deformation as depicted in A, B, and C.
Matching costs computed with different descriptors along A, B, and C's scan-lines are plotted in (c)-(e). Unlike conventional descriptors,
the proposed DASC descriptor yields a reliable global minimum.}\label{img:2}\vspace{-10pt}
\end{figure}

Establishing dense visual correspondences for multi-modal and multi-spectral images is a key enabler for realizing
such tasks. In general, the performance of correspondence algorithms relies primarily on two components: appearance
descriptor and optimization scheme. Traditional dense
correspondence methods for estimating depth \cite{Scharstein02} or
optical flow \cite{Butler12,Sun10} fields, in which input images are
acquired in a similar imaging condition, have been dramatically
advanced in recent studies. To define a matching fidelity term, they
typically assume that multiple images share a similar visual
pattern, \emph{e.g.}, color, gradient, and structural similarity.
However, when it comes to multi-spectral and multi-modal images,
such properties do not hold as shown in \figref{img:1}, and thus conventional
descriptors or similarity measures often fail to capture reliable
matching evidence.
This leads to a poor matching quality as shown in
\figref{img:2}. Furthermore, substantial geometric
variations, which often appear in images captured under wide-baseline conditions,
make the matching task even more challenging. Although
employing a powerful optimization technique could help estimate a
reliable solution with a spatial context \cite{Liu11,Kim13,Li15}, an
optimizer itself cannot address an inherent limitation without
suitable matching descriptors \cite{Pinggera12}.

Our method starts from an observation that a local internal
layout of self-similarities is less sensitive to photometric
distortions, even when an intensity distribution of an anatomical
structure is not maintained across different imaging modalities
\cite{Schechtman07}. That is, the local self-similarity (LSS) descriptor
would be beneficial to overcoming inherent limitations
of existing descriptors in establishing correspondences between
multi-modal or multi-spectral images. Several approaches based on
the LSS have been presented for multi-modal and multi-spectral image
registration \cite{Mattias12,Torabi13}, but they do not scale well
to estimating dense correspondences for multi-modal and
multi-spectral images, and their matching performance is still poor.

In this paper, we propose a novel local descriptor, called dense
adaptive self-correlation (DASC), designed for establishing dense
multi-modal and multi-spectral correspondences. It is defined with a
series of patch-wise similarities within a local support window. The
similarity is computed with an adaptive self-correlation measure,
which encodes an intrinsic structure while providing the robustness
against modality variations. To further improve the matching quality
and runtime efficiency, we propose a randomized receptive field
pooling strategy using sampling patterns that select two patches
within the local support window. A linear discriminative learning is
employed for obtaining an optimal sampling pattern.
The computational redundancy that arises when computing densely sampled
descriptors over an entire image is dramatically reduced by applying
fast edge-aware filtering \cite{He13}.

Furthermore, in order to address geometric variation problems
such as the scale and rotation, we propose the geometry-invariant DASC
(GI-DASC) descriptor that leverages the efficiency and effectiveness
of the DASC through a superpixel-based representation. Specifically,
we infer an initial geometric field with corresponding scale and
rotation of reliable sparse key-points obtained using weighted
maximally self-dissimilarity (WMSD), and then
propagate the initial geometric field on a superpixel graph.
After transforming sampling patterns according to geometric fields on each superpixel,
the DASC is efficiently computed with the transformed sampling patterns on each superpixel extended subimage.
Compared to conventional geometry-invariant methods for dense correspondence \cite{Yang14,Hur15},
which have been focusing on employing powerful optimization schemes,
the GI-DASC provides geometric and photometric robustness on the descriptor itself.

Experimental results show that the DASC outperforms conventional
area-based and feature-based approaches on various benchmarks including modality variations; (1) Middlebury stereo benchmark containing
illumination and exposure variations \cite{middlebury}, (2)
multi-modal and multi-spectral dataset including RGB-NIR images
\cite{Shen14,Brown11}, different exposure \cite{Shen14,Sen12},
flash-noflash images \cite{Sen12}, and blurry images
\cite{HaCohen13,Lee13}, and (3) MPI optical flow benchmark containing
specular reflections, motion blur, and defocus blur \cite{Butler12}. We also show that the GI-DASC outperforms
existing geometry-invariant methods on a novel multi-modal
benchmark. \vspace{-5pt}
\subsection{Contribution}\label{sec:11}
The contributions of this paper can be summarized as follows.
First, to the best of our knowledge, our approach is the first attempt to
design an efficient, dense descriptor for matching multi-modal and
multi-spectral images, even under varying geometric conditions.
Second, unlike a center-biased dense max pooling, we propose a
randomized receptive field pooling with sampling patterns optimized
via a discriminative learning, making the descriptor more robust to
matching outliers incurred by different imaging modalities.
Third, we propose an efficient computational scheme that significantly
improves the runtime efficiency of the proposed dense descriptor.
Fourth, a geometry-invariant dense descriptor is also proposed,
which provides a geometric robustness as a descriptor itself.

This manuscript extends its preliminary version \cite{Kim15}. It
newly adds (1) a scale and rotation invariant extension of the DASC,
called GI-DASC; (2) a new multi-modal benchmark with a ground truth
annotation, captured under varying photometric and geometric
conditions; and (3) an intensive comparative study with existing
geometry invariant methods using various datasets. The source code
of our work (including DASC and GI-DASC) and the new multi-modal
benchmark are available at our project webpage \cite{dasc}.
\vspace{-5pt}
\section{Related Work}\label{sec:2}
\subsection{Feature Descriptors}\label{sec:21}
As a pioneering work, the scale invariant feature transform (SIFT)
was first introduced by Lowe \cite{Lowe04} to estimate robust sparse
feature correspondence under geometric and photometric variations.
Based on the intensity comparison, fast binary
descriptors, such as binary robust independent elementary features
(BRIEF) \cite{Calonder11} and fast retina keypoint (FREAK)
\cite{Alahi12}, have been proposed. Unlike these sparse
descriptors, Tola \emph{et al.} developed a dense descriptor, called
DAISY \cite{Tola10}, which re-designs conventional sparse
descriptors, \emph{i.e.}, SIFT, to efficiently compute densely
sampled descriptors over an entire image. Although these
conventional gradient-based and intensity comparison-based
descriptors show satisfactory performance for small photometric
deformation, they cannot properly describe multi-modal and
multi-spectral images that often exhibit severe non-linear
deformation.

To estimate correspondences in multi-modal and multi-spectral
images, some variants of the SIFT have been developed \cite{Saleem14},
but these gradient-based descriptors have an inherent limitation
similar to the SIFT, especially when an image gradient varies across
different modality images. Schechtman and Irani introduced the LSS
descriptor \cite{Schechtman07} for the purpose of template matching,
and achieved impressive results in object detection and retrieval.
Torabi \emph{et al.} employed the LSS as a multi-spectral similarity
metric to register human region of interests (ROIs) \cite{Torabi13}.
The LSS also has been applied to the registration of multi-spectral
remote sensing images \cite{Ye14}. For multi-modal medical image
registration, Heinrich \emph{et al.} proposed a modality independent
neighborhood descriptor (MIND) \cite{Mattias12} inspired by the LSS.
However, none of these approaches scale very well to dense matching
tasks for multi-modal and multi-spectral images due to a low
discriminative power and a huge complexity. 

Recently, several approaches started to employ deep
convolutional neural networks (CNNs) \cite{Alex12} for estimating correspondences. 
For designing explicit, discriminative feature descriptors, intermediate
activations from CNN architecture are extracted 
\cite{Simonyan14,Fischer14,Donahue14,Simo-Serra15b}, and they have
been shown to be effective for patch-level tasks. However, even
though CNN-based descriptors encode a discriminative structure with
a deep architecture, they have inherent limitations in multi-modal
images, since they use shared convolutional kernels 
across images which lead to inconsistent responses similar to conventional 
descriptor \cite{Dong15,Simo-Serra15b}. Furthermore,
they are unable to provide dense descriptors in the image due to a
prohibitively high computational complexity.
\vspace{-5pt}
\subsection{Area-based Similarity Measures}\label{sec:22}
As surveyed in \cite{Pluim03}, the mutual information (MI),
leveraging the entropy of the joint probability distribution
function (PDF), has been popularly applied to a registration
of multi-modal medical images. However, the MI is sensitive to
local radiometric variation since it formulates the intensity
variation in a global manner using the joint entropy computed over
an entire image. In \cite{Heo13}, this issue can be alleviated to some extent by
leveraging a locally adaptive weight obtained from SIFT matching, 
called MI+SIFT in this paper, but its
performance is still limited against the multi-modal variation \cite{Xu16}. 
Although cross-correlation based
methods such as an adaptive normalized cross-correlation (ANCC)
\cite{Heo} show satisfactory results for locally linear
variations, they show a limitation under severe modality
variations. Irani \emph{et al.} employed the
cross-correlation on the Laplacian energy map for measuring
multi-sensor image similarity \cite{Irani98},
but it also shows a limitation for general image matching tasks.
A robust selective
normalized cross-correlation (RSNCC) \cite{Shen14} was proposed for
the dense alignment between multi-modal images, but its performance
is still unsatisfactory due to an inherent limitation of intensity
based similarity measure. \vspace{-5pt}
\begin{figure}[!t]
	\centering
	\renewcommand{\thesubfigure}{}
	\subfigure[(a) LSS descriptor \cite{Schechtman07}]
	{\includegraphics[width=0.45\linewidth]{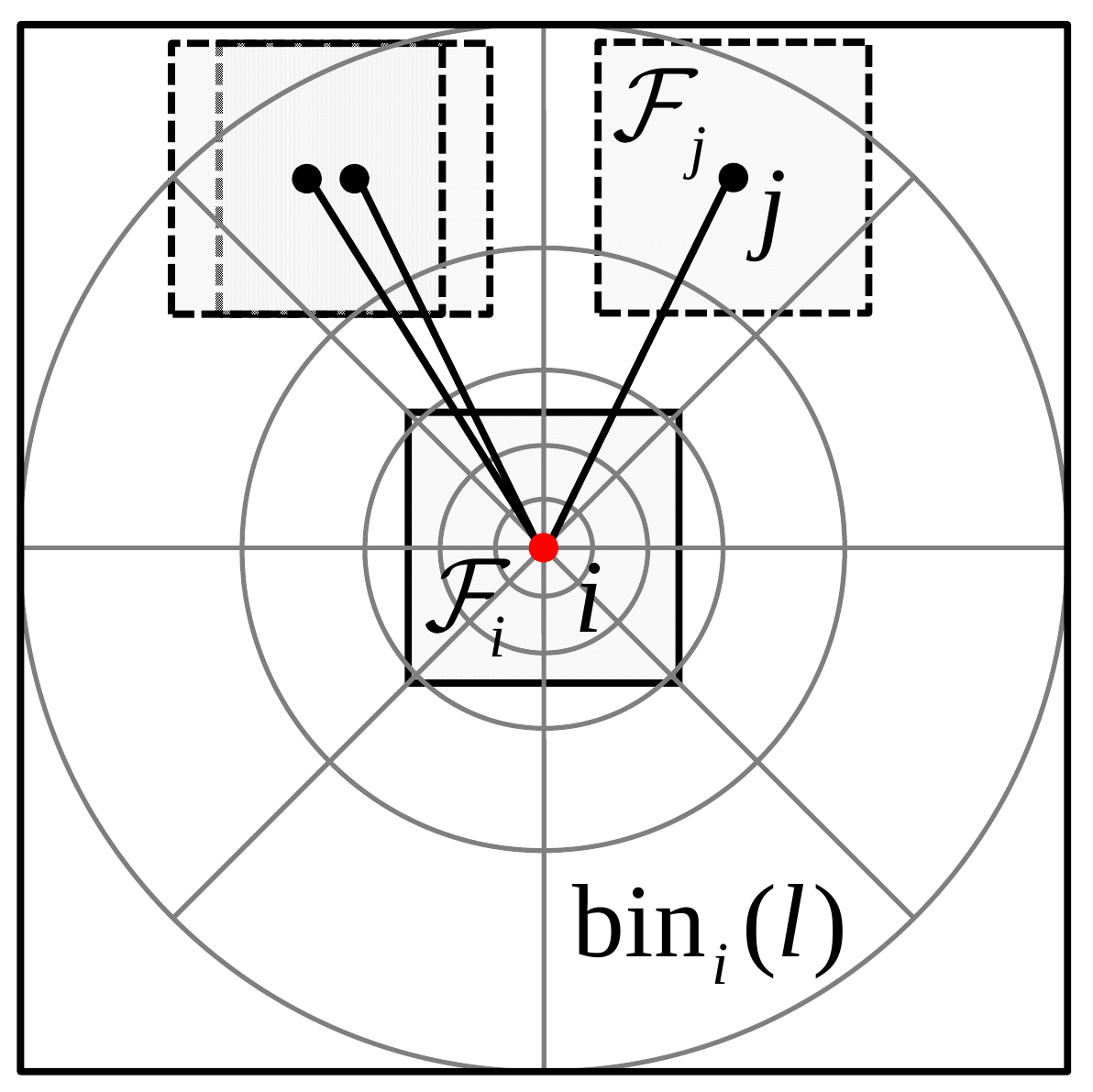}}
	\subfigure[(b) DASC descriptor]
	{\includegraphics[width=0.45\linewidth]{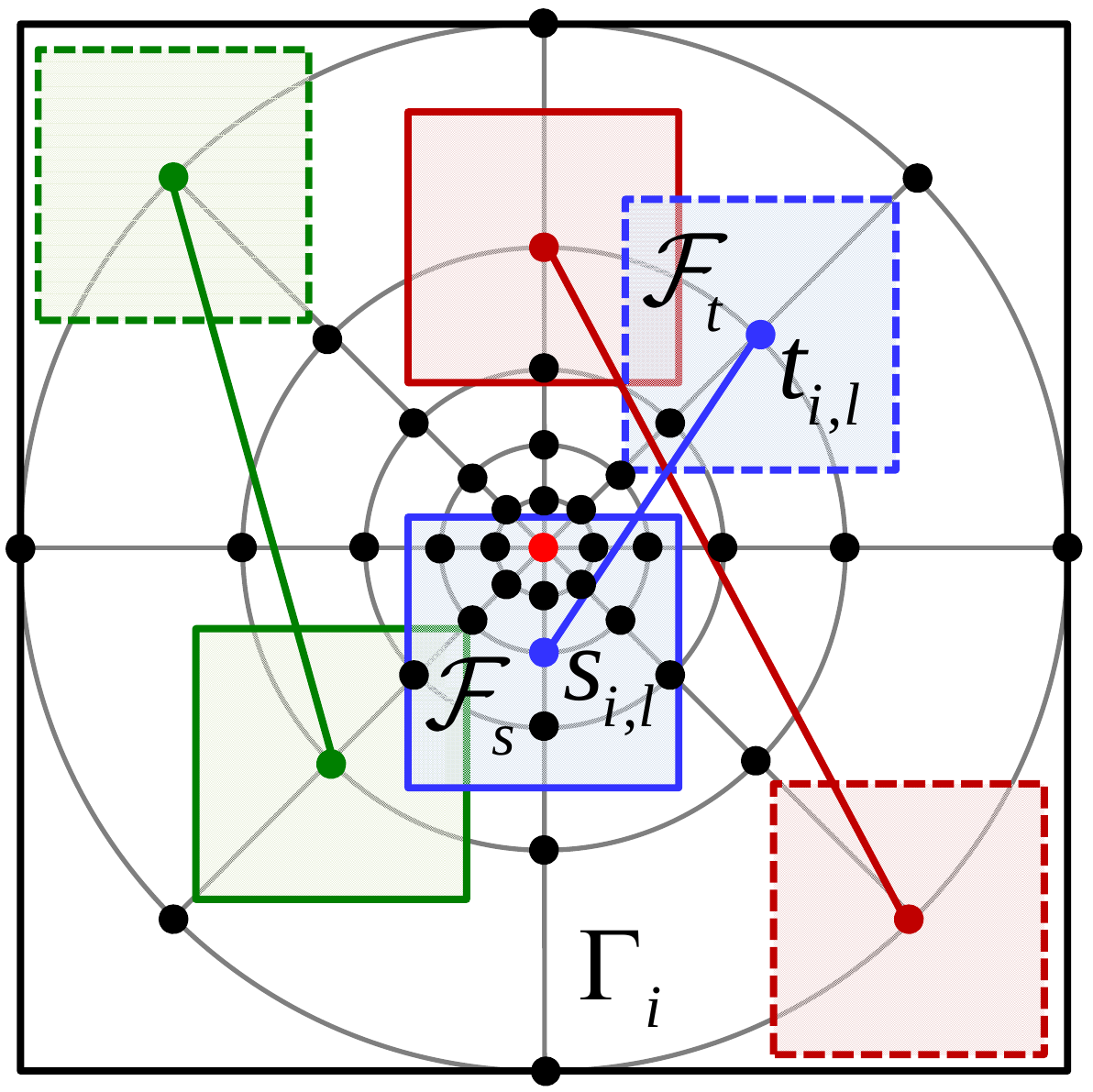}}\\
	\vspace{-8pt}
	\caption{Demonstration of the LSS \cite{Schechtman07} and the DASC descriptor.
		Within the support window, solid and dotted line box depict source and target patch, respectively.
		Unlike a center-biased dense max pooling on each $\textrm{bin}_{i}(l)$ in the LSS descriptor,
		the DASC descriptor incorporates a randomized receptive field pooling
		using sampling pattern $(s_{i,l},t_{i,l}) \in \Lambda_{i}^{\mathrm{dasc}}$ on ${\Gamma_{i}}$, optimized by a discriminative learning.}\label{img:3}\vspace{-10pt}
\end{figure}
\subsection{Geometry-Invariant Dense Correspondences}\label{sec:23}
Based on the SIFT flow (SF) \cite{Liu11} optimization, many methods
have been proposed to alleviate geometric variation problems, including
deformable spatial pyramid (DSP) \cite{Kim13}, scale-less SIFT flow
(SLS) \cite{Hassner12}, scale-space SIFT flow (SSF)
\cite{Qiu14}, and generalized DSP (GDSP) \cite{Hur15}.
However, they have a critical limitation as huge
computational complexity derived from dramatically large search space
in geometry-invariant dense correspondence.
A generalized PatchMatch (GPM) \cite{Barnes10} 
was proposed for efficient matching leveraging a randomized search scheme.
The DAISY Filter Flow (DFF) \cite{Yang14}, which exploits DAISY descriptor
\cite{Tola10} with PatchMatch Filter (PMF) \cite{Lu13}, was proposed
to provide geometric invariance.
However, their weak spatial smoothness often induces mismatched results.
The scale invariant descriptor (SID) \cite{Kokkinos08} was proposed
to encode geometric robustness on the descriptor itself, but it is not
tailored to multi-modal matching. Segmentation-aware approach
\cite{Trulls13} was proposed to provide geometric robustness for
descriptors, \emph{e.g.}, SIFT \cite{Lowe04} or SID
\cite{Kokkinos08}, but it may have a negative effect on the discriminative power of the descriptor. \vspace{-5pt}
\section{Background}\label{sec:3}
Let us define an image as ${f_i}:\mathcal{I} \to {\mathbb R}$ for pixel $i$, where
$\mathcal{I} \subset{{\mathbb N}^2}$ is
a discrete image domain. Given the image ${f_i}$, a dense descriptor
${\mathcal{D}_i}:\mathcal{I} \to \mathbb{R}^L$ is defined on a local
support window $\mathcal{R}_i$ centered at pixel $i$ with a feature dimension
$L$. Conventionally, descriptors were computed based on the
assumption that there is a common underlying visual pattern which is
shared by two images. However, as shown in \figref{img:2},
multi-spectral images such as a pair of RGB-NIR have a nonlinear
photometric deformation even within a small window, \emph{e.g.},
gradient reverse and intensity order variation. More seriously,
there are outliers including structure divergence caused by shadow
or highlight. In these cases, conventional descriptors using an
image gradient (SIFT \cite{Lowe04}) or an intensity comparison
(BRIEF \cite{Calonder11}) cannot capture coherent matching
evidences, resulting erroneous local minima in estimating
dense correspondences. 

Unlike these conventional descriptors, the LSS descriptor
$\mathcal{D}_i^{\textrm{lss}}$ measures a correlation between two
patches ${\mathcal{F}_i}$ and ${\mathcal{F}_{j}}$ centered at two
pixels $i$ and $j$ within a local support window $\mathcal{R}_i$
\cite{Schechtman07}. As shown in \figref{img:3}(a), it discretizes
the correlation surface on a log-polar grid, generates a set of
bins, and then stores a maximum correlation value within each bin.
Formally, $\mathcal{D}_{i}^{\textrm{lss}} = { \bigcup
_{l}}d_{i,l}^{\textrm{lss}}$ for $l=1,...,L^{\textrm{lss}}$ is a
$L^{\textrm{lss}}\times1$ feature vector,
and $d_{i,l}^{\textrm{lss}}$ can be computed as follows:
\begin{equation}\label{equ:1}
\begin{array}{l}
d_{i,l}^{\textrm{lss}} = \mathop {\max }\limits_{j \in
\textrm{bin}_{i}(l)} \{ \mathcal{C}(i,j)\},
\end{array}
\end{equation}
where $\textrm{bin}_{i}(l) = \{j|j\in\mathcal{R}_i,\rho_{r-1}<{|i -
j|}\leq\rho_{r}, \theta_{a-1}<{\angle (i - j)}\leq\theta_{a}\}$ with
a log radius $\rho_r$ for $r\in \{1,\cdots,N_\rho\}$ and a quantized
angle $\theta_a$ for $a\in \{1,\cdots,N_\theta\}$ with $\rho_{0}=0$
and $\theta_{0}=0$.
In that case, $L^{\textrm{lss}} = N_\rho \times N_\theta$.
The correlation surface $\mathcal{C}(i,j)$ is
typically computed using a simple similarity metric such as the sum
of squared difference (SSD) with a normalization factor $\sigma_{s}
$:
\begin{equation}\label{equ:2}
\mathcal{C}(i,j) = \exp \left( { -
\textrm{SSD}\left({\mathcal{F}_i},{\mathcal{F}_{j}}\right)/\sigma_{s} } \right).
\end{equation}

This LSS descriptor has been shown to be robust in cross-domain object detection \cite{Schechtman07},
but it provides unsatisfactory results in densely matching multi-modal images as
shown in \figref{img:2}. It is because the max pooling strategy performed in each
$\textrm{bin}_{i}(l)$ loses matching details, leading to a poor discriminative power.
Furthermore, the center-biased correlation measure cannot handle severe outliers effectively, which
frequently exist in multi-modal and multi-spectral images. In terms of a
computational complexity, there exists no efficient computational
scheme designed for dense matching descriptor.
\vspace{-5pt}
\section{The DASC Descriptor}\label{sec:4}
\subsection{Randomized Receptive Field Pooling}\label{sec:41}
Instead of using a center-biased max pooling of the LSS descriptor in
\figref{img:3}(a), our DASC descriptor incorporates a randomized receptive field
pooling with sampling patterns in such a way that a pair of two patches are
randomly selected within a local support window. It is motivated by
three observations; 1) In multi-spectral and multi-modal images, there
frequently exist non-informative regions which are locally degraded,
\emph{e.g.}, shadows or outliers. 2) Center-biased pooling is very
sensitive to a degradation of a center patch, and cannot deal with a
homogeneous or salient center pixel which does not contain
self-similarities \cite{Schechtman07}. 3) From the relationship
between Census transform \cite{Zabih94} and BRIEF \cite{Calonder11}
descriptor, it is shown that the randomness enables a descriptor
to encode structural information more robustly.

Our approach encodes a similarity between patch-wise receptive
fields sampled from log-polar circular point set $\Gamma_{i}$ as
shown in \figref{img:3}(b). It is defined as $\Gamma_{i} = \{j | j
\in \mathcal{R}_i, |{i} - {j}|=\rho_{r}, \angle ({i} -
{j})=\theta_{a} \}$ where the number of points is defined as $N_{c}
= N_\rho \times N_\theta +1$, and has a higher density of points
near a center pixel, similar to DAISY descriptor \cite{Tola10}.
Given ${N_c}$ points in $\Gamma_{i}$, there exist $N_{pc}={\{N_c
\times (N_c-1)\}/2}$ candidate sampling patterns, leading to a
dramatically high-dimension descriptor. However, many of the
sampling pattern pairs might not be useful in describing a local
support window. Therefore, we employ a randomized approach to
extract $L^{\mathrm{dasc}}$ sampling patterns from $N_{pc}$ pattern
candidates. Our descriptor $\mathcal{D}^{\mathrm{dasc}}_{i} = {
\bigcup _{l}}d^{\mathrm{dasc}}_{i,l}$ for
$l=1,...,L^{\mathrm{dasc}}$ is encoded with a set of patch
similarity between two patches based on sampling patterns that are
selected from $\Gamma_{i}$:
\begin{equation}\label{equ:dasc}
\begin{array}{l}
d^{\mathrm{dasc}}_{i,l} = \mathcal{C}(s_{i,l},t_{i,l}),\quad s_{i,l},t_{i,l} \in {\Gamma_{i}},\\
\end{array}
\end{equation}
where $s_{i,l}$ and $t_{i,l}$ are $l^{th}$ selected sampling
patterns at pixel $i$. Note that the sampling patterns are fixed for
all pixels in an image. Namely, all pixels share the same set
of offset vectors $t_{i,l}-s_{i,l}$ for
$l=1,...,L^{\mathrm{dasc}}$, enabling a fast computation of dense
descriptors, which will be detailed in \secref{sec:44}. Although the
DASC descriptor uses only sparse patch-wise pairs in a local support
window, many of patches are overlapped when computing patch
similarities between the sparse pairs, allowing the descriptor to
consider the majority of pixels in the support window and reflect
original image attributes effectively. \vspace{-5pt}
\begin{figure}[!t]
\centering
\renewcommand{\thesubfigure}{}
\subfigure[]
{\includegraphics[width=0.33\linewidth]{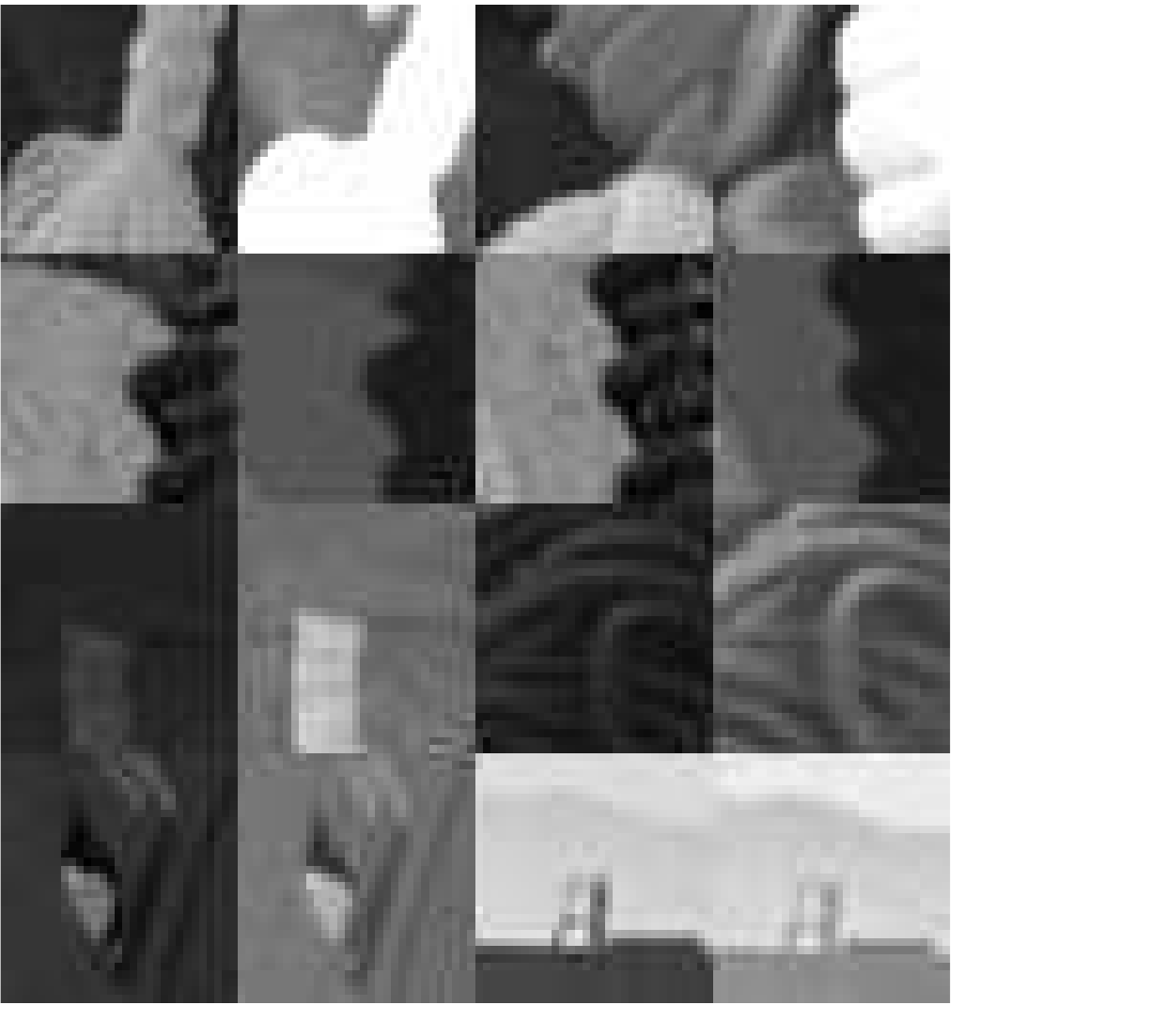}}\hfill
\subfigure[]
{\includegraphics[width=0.33\linewidth]{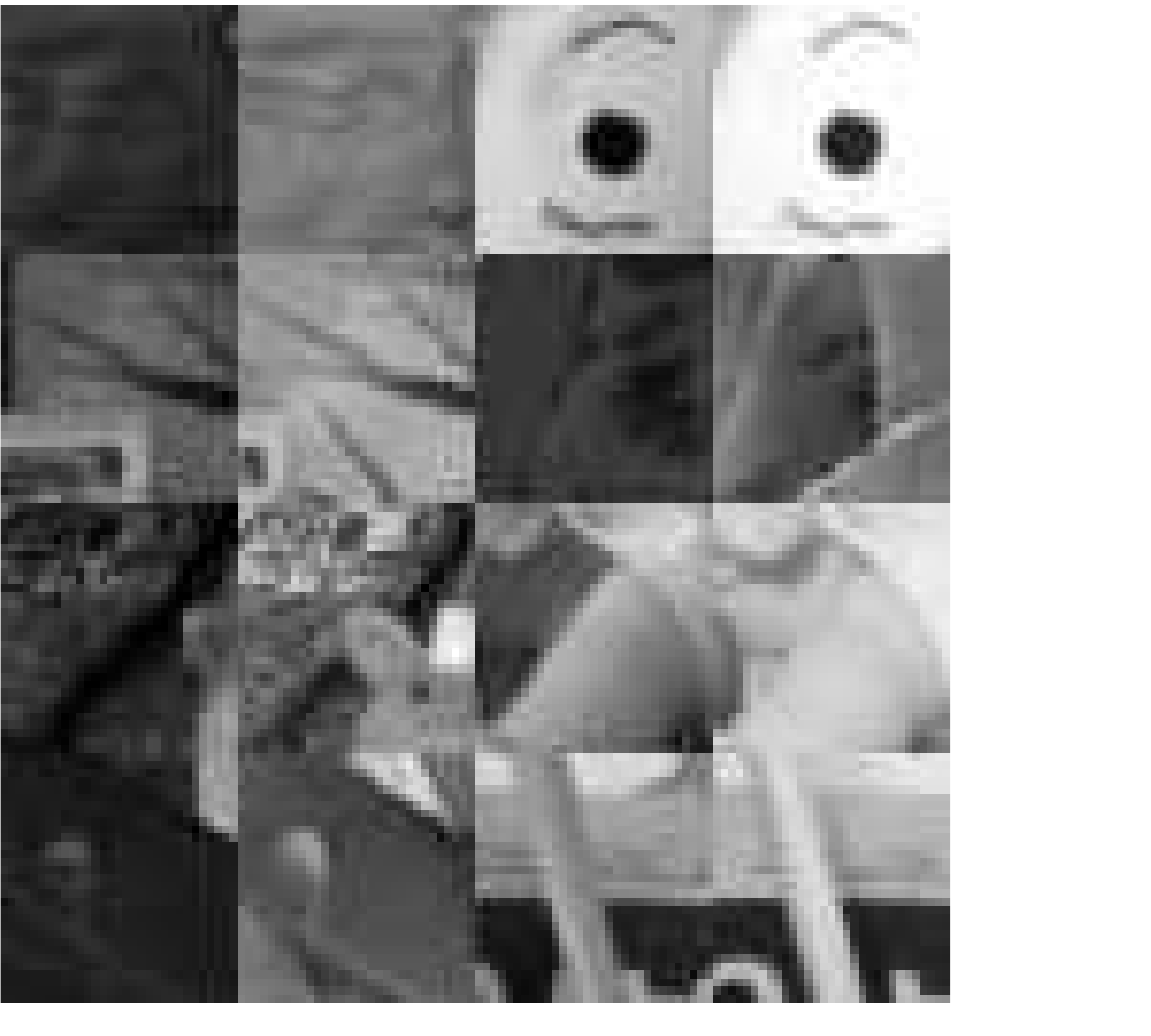}}\hfill
\subfigure[]
{\includegraphics[width=0.33\linewidth]{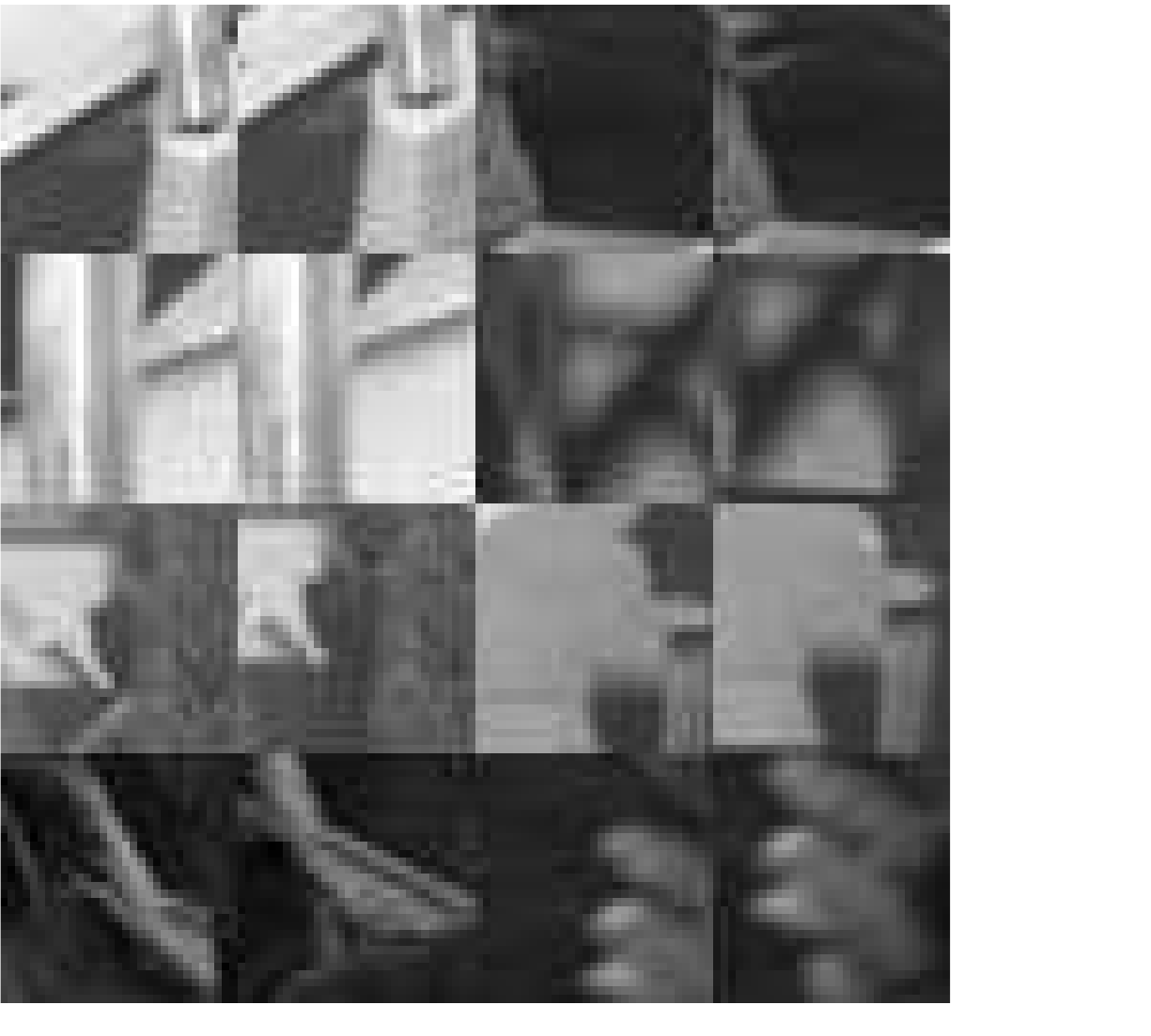}}\hfill
\vspace{-20pt} \subfigure[(a) Middlebury \cite{middlebury}]
{\includegraphics[width=0.33\linewidth]{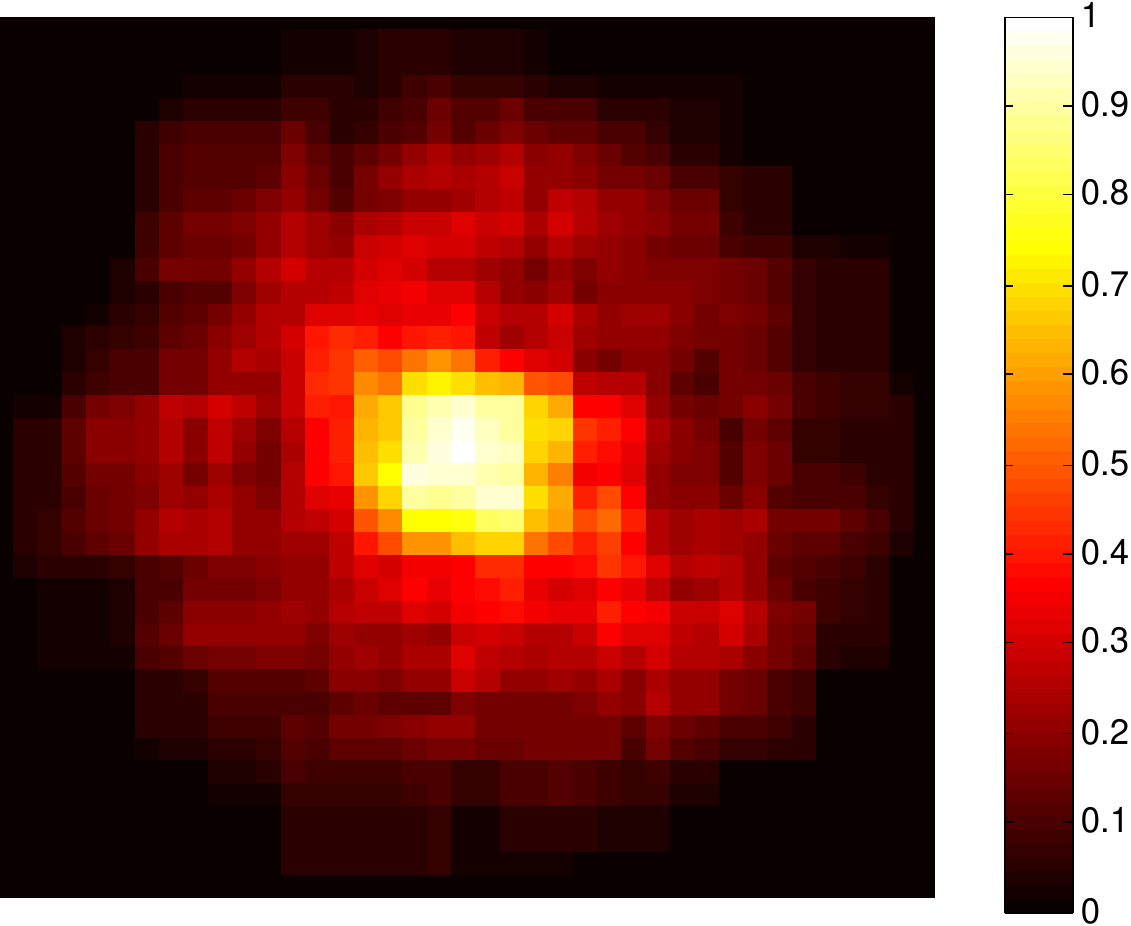}}\hfill
\subfigure[(b) Multi-modal \cite{Shen14}]
{\includegraphics[width=0.33\linewidth]{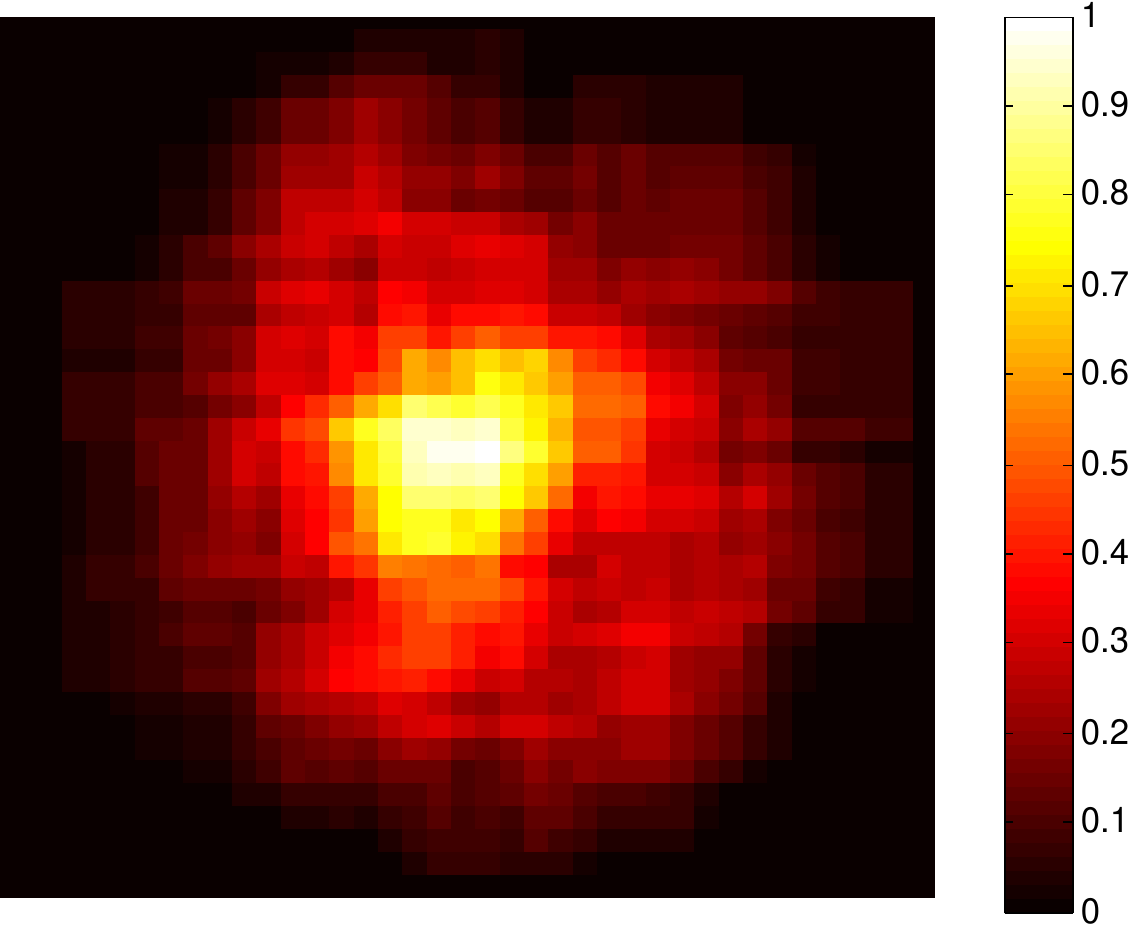}}\hfill
\subfigure[(c) MPI SINTEL \cite{Butler12}]
{\includegraphics[width=0.33\linewidth]{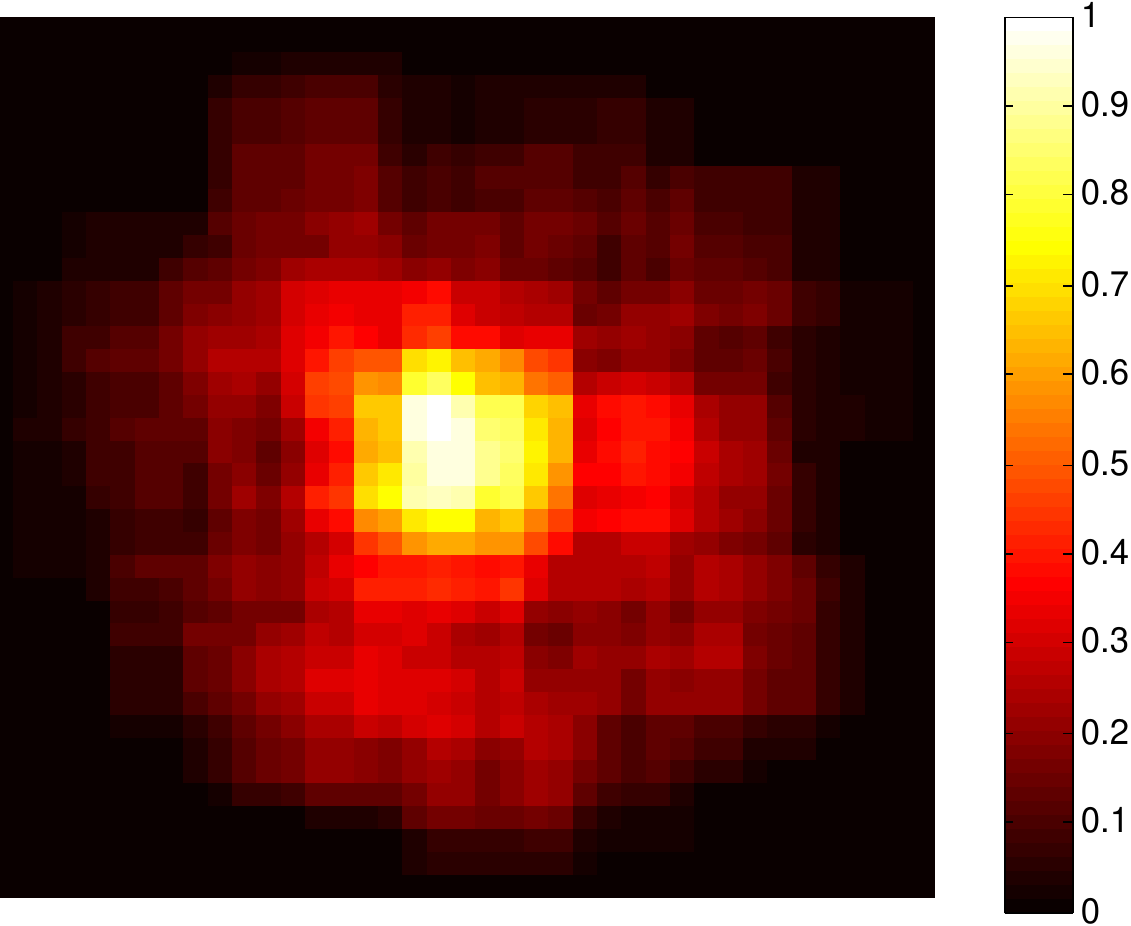}}\hfill
\vspace{-8pt} \caption{Visualization of patch-wise receptive fields
of the DASC descriptor learned from the training set $\mathcal{P}$
built with the Middlebury benchmark \cite{middlebury},
multi-modal benchmark \cite{Shen14}, and the MPI
SINTEL benchmark \cite{Butler12}. Similar to \cite{Fan14}, we
stacked all patch-wise receptive fields learned from each training
image, and normalized them with the maximal
value.}\label{img:4}\vspace{-10pt}
\end{figure}
\subsubsection{Sampling pattern learning}
Finding an optimal sampling pattern is a critical issue in
the DASC descriptor. With the assumption that there is no single
hand-craft feature that always provides the robustness to all
circumstances \cite{Fan14}, we employ a discriminative learning to
obtain optimal sampling patterns within a local support
window. Given candidate sampling patterns $ \Lambda_{i} =
\{(s_{i,l},t_{i,l})|l=1,...,N_{pc}\}$, our goal is to select the
best sampling patterns which derive an important spatial layout.

Our approach exploits support vector machines (SVMs) with a linear
kernel \cite{Change11}. For learning, we build a dataset
$\mathcal{P} = \{(\mathcal{R}_h^1,\mathcal{R}_h^2,{y_h})| h=1,...,N_{tr}\} $, where
$(\mathcal{R}^1,\mathcal{R}^2)$ are support window pairs in
multi-modal or multi-spectral images, and $N_{tr}$ is the
number of training samples. $y$ is a binary label that becomes 1 if
two patches are matched, or 0 otherwise. The training data set
$\mathcal{P}$ was built with images captured under varying illumination conditions and/or with
imaging devices \cite{middlebury,Butler12,Shen14}. In experiments, $N_{tr}=10,000$.

First, the feature $\mathbf{r}_h={\bigcup_{l}r_{h,l}}$ that describes two support window
pairs $\mathcal{R}_{h}^1$ and $\mathcal{R}_{h}^2$ is defined
\begin{equation}\label{equ:8}
{r _{h,l}} = \exp \left( { - {{\left(d_{h,l}^{\mathrm{dasc},1} - d_{h,l}^{\mathrm{dasc},2}\right)}^2}/2{\sigma_{r}^2}} \right),
\end{equation}
where $\sigma_r$ is a Gaussian parameter, and
$d^{\mathrm{dasc}}_{h,l}$ is the DASC descriptor. The decision
function $\mathcal{Q}$ to classify training dataset $\mathcal{P}$
into matching and non-matching can be represented as
\begin{equation}\label{equ:9}
\mathcal{Q}(\mathbf{r}_h ) = \mathbf{v}^T \mathbf{r}_h  + \mathbf{b},
\end{equation}
where the weight $\mathbf{v}={\bigcup_{l}v_{l}}$ indicates an amount of contribution of each
candidate sampling pattern, and $\mathbf{b}$ is a bias.
Learning $\mathbf{v}$ can be formulated as minimizing
\begin{equation}\label{equ:10}
\mathbf{E}_{\mathrm{svm}}(\mathbf{v}) = \|\mathbf{v}\|^{2} +
C_{\mathrm{svm}}\sum\nolimits_{h=1}^{N_{tr}} {l_\mathrm{hinge}\left({y_h} \cdot
\mathcal{Q}(\mathbf{r}_h)\right)},
\end{equation}
where the hinge loss function $l_\mathrm{hinge}(x) = \mathrm{max}(0,1-x)$ and
$C_{\mathrm{svm}}$ represents a regularization parameter. We use
LIBSVM \cite{Change11} to minimize this objective function. The
$|v_{l}|$ encodes the importance of corresponding sampling
pattern towards the final decision \cite{Lee14}. Therefore, we rank
top $L^{\mathrm{dasc}}$ sampling patterns based on $|v_{l}|$ value,
and use them in our descriptor, which is denoted as $
\Lambda_{i}^{\mathrm{dasc}}$. 

\figref{img:4} visualizes learned patch-wise receptive fields of the DASC. 
It looks similar to the Gaussian weighting, 
which has been proven to be effective in terms of a structural encoding of 
descriptor in many literatures \cite{Fan14,Trzcinski15}. 
According to training set, it learns optimal receptive fields.
\vspace{-5pt}
\begin{figure}[!t]
	\centering
	\renewcommand{\thesubfigure}{}
	\subfigure[(a) Window 1]
	{\includegraphics[width=0.245\linewidth]{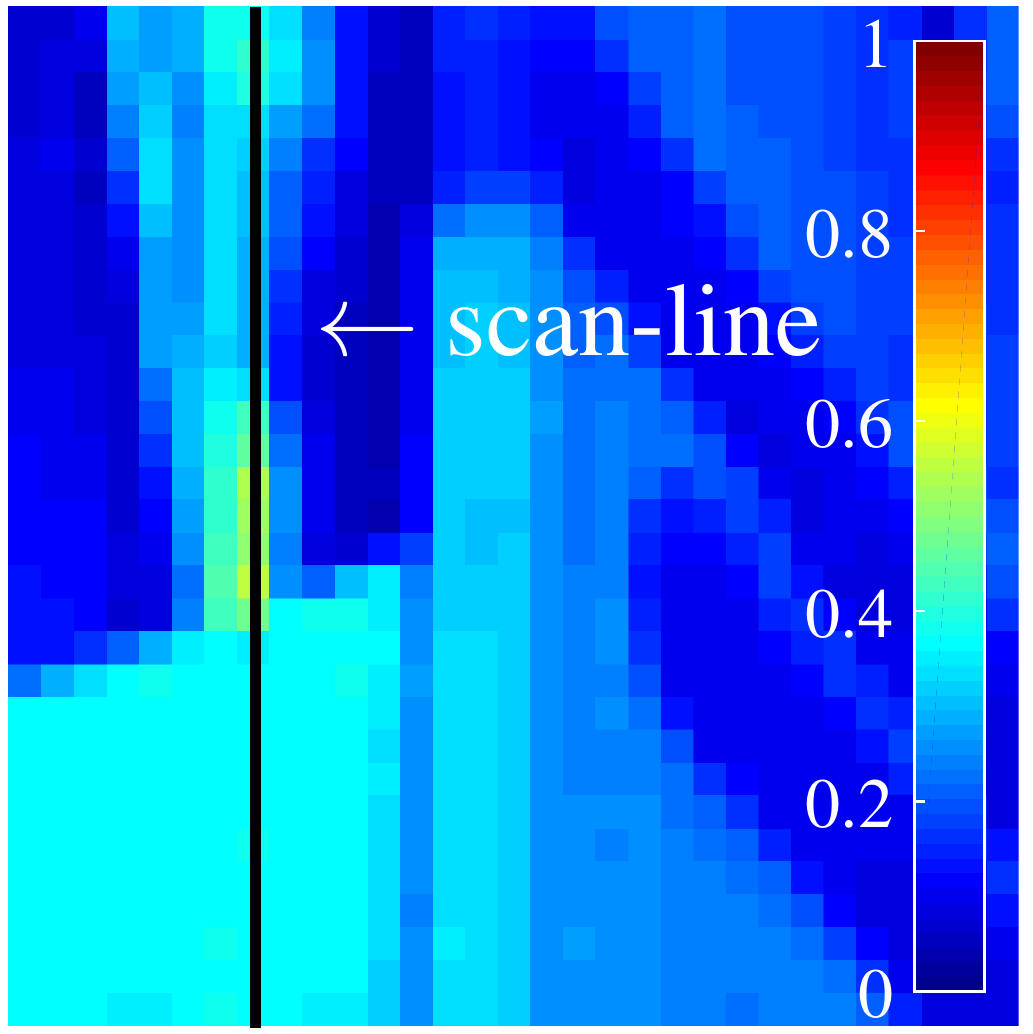}}\hfill
	\subfigure[(b) Window 2]
	{\includegraphics[width=0.245\linewidth]{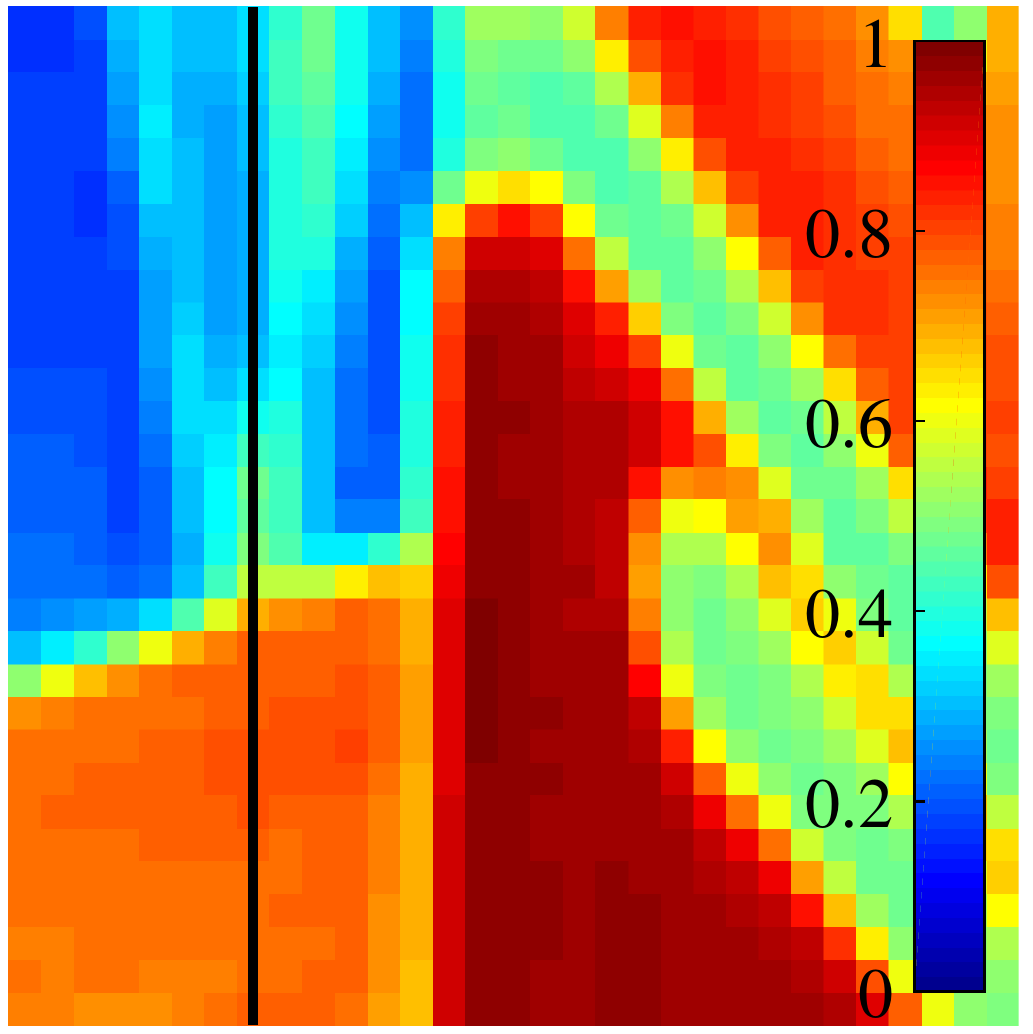}}\hfill
	\subfigure[(c) Gradient orientation]
	{\includegraphics[width=0.5\linewidth]{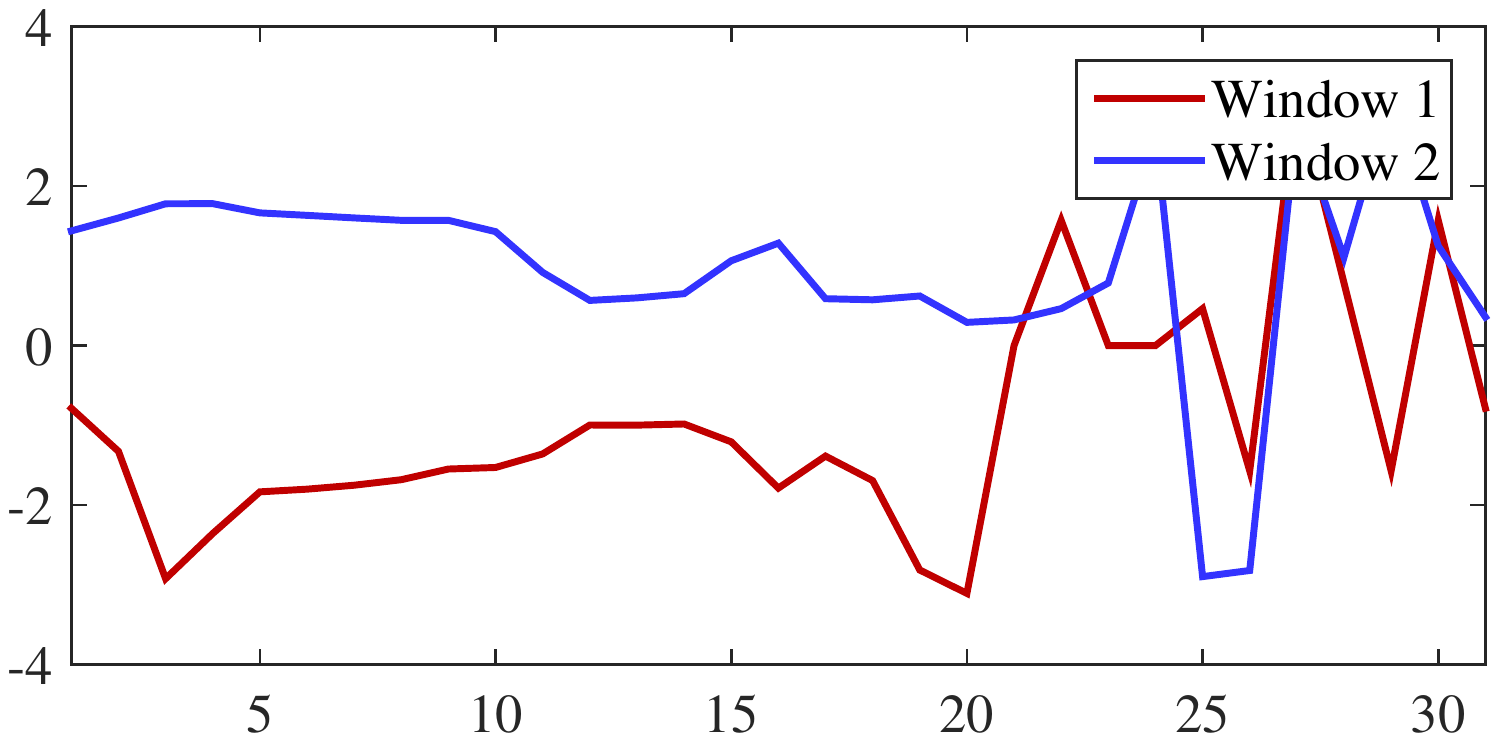}}\hfill
	\vspace{-8pt}
	\subfigure[(d) DAISY \cite{Tola10} descriptor]
	{\includegraphics[width=0.5\linewidth]{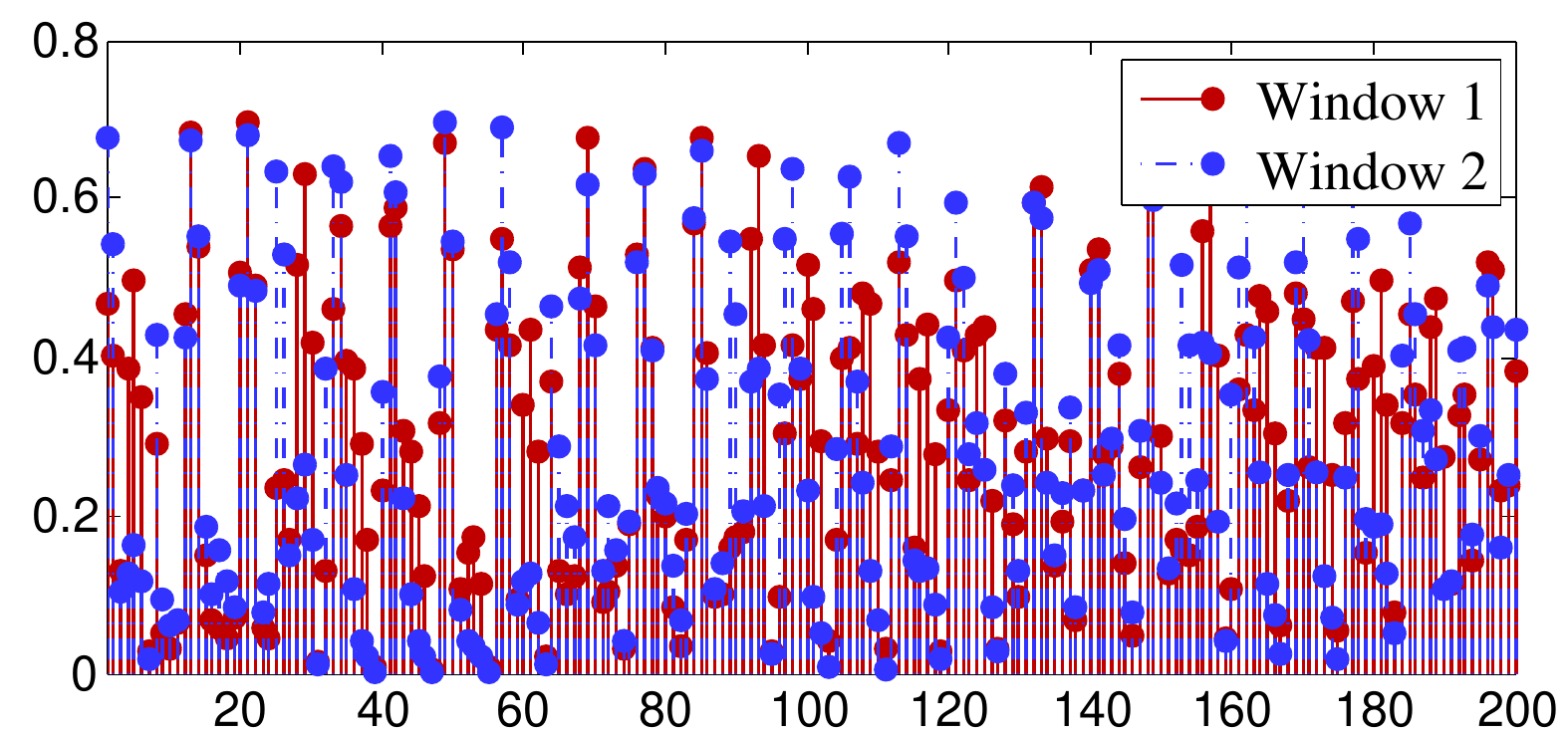}}\hfill
	\subfigure[(e) BRIEF \cite{Calonder11} descriptor]
	{\includegraphics[width=0.5\linewidth]{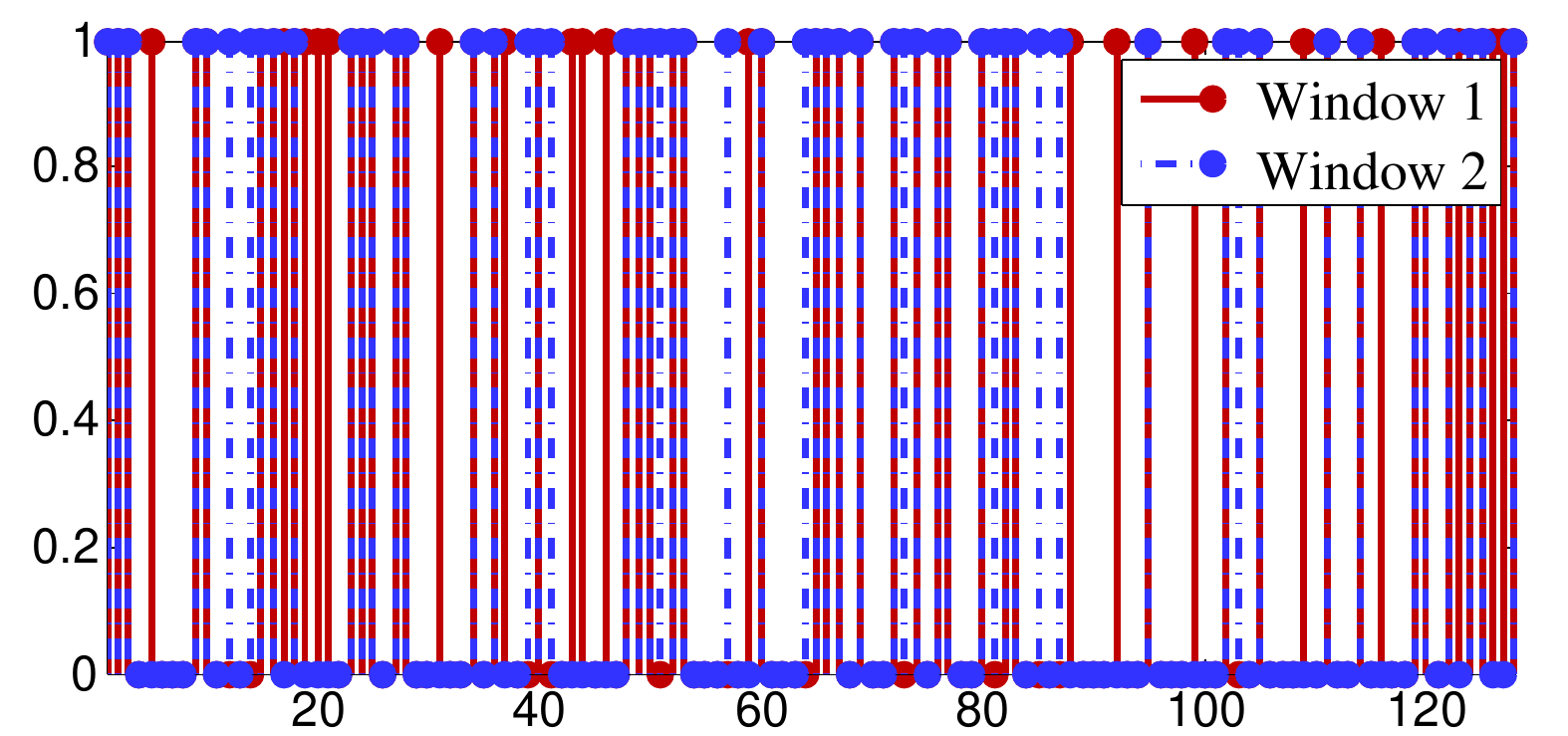}}\hfill
	\vspace{-8pt}
	\subfigure[(f) LSS \cite{Schechtman07} descriptor]
	{\includegraphics[width=0.5\linewidth]{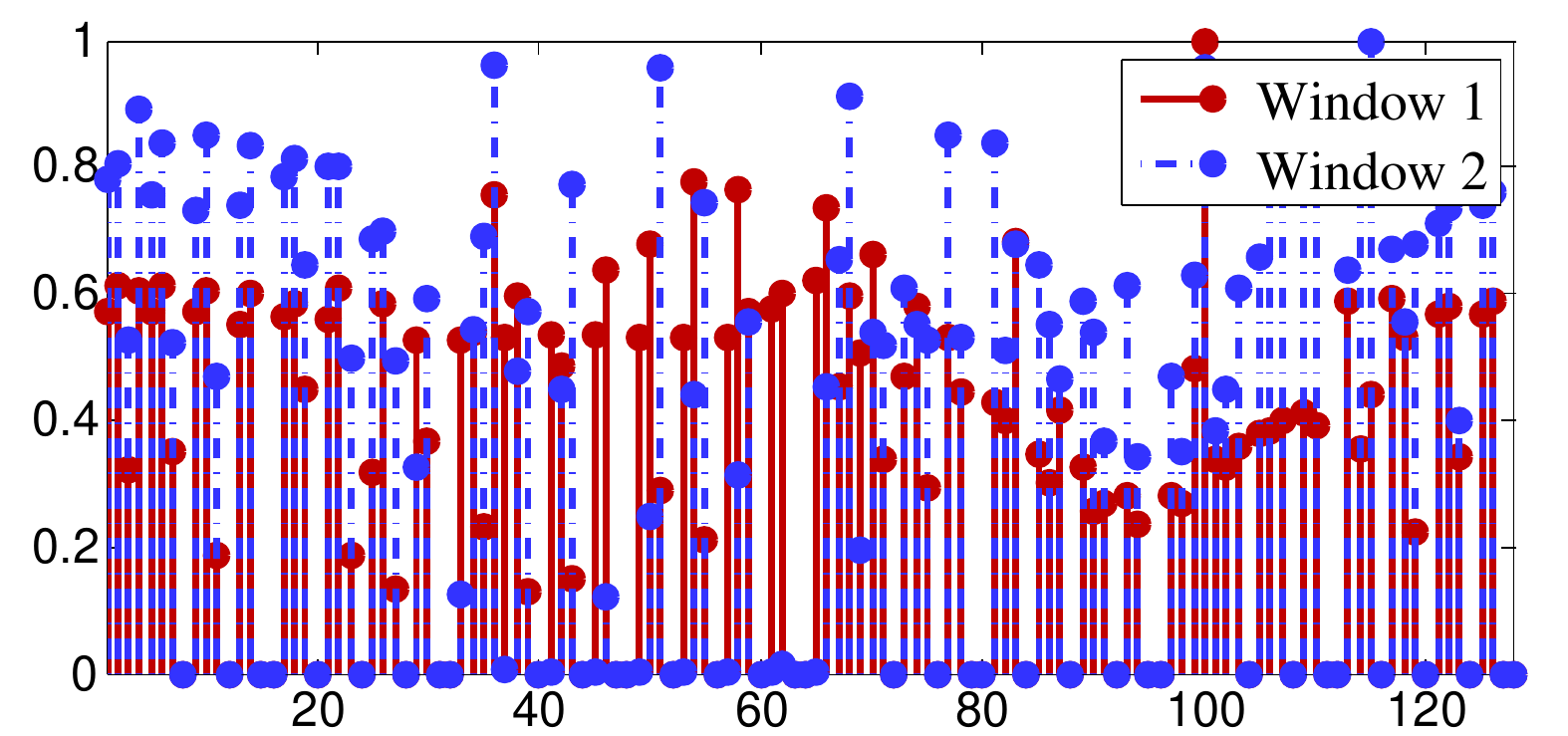}}\hfill
	\subfigure[(g) DASC descriptor]
	{\includegraphics[width=0.5\linewidth]{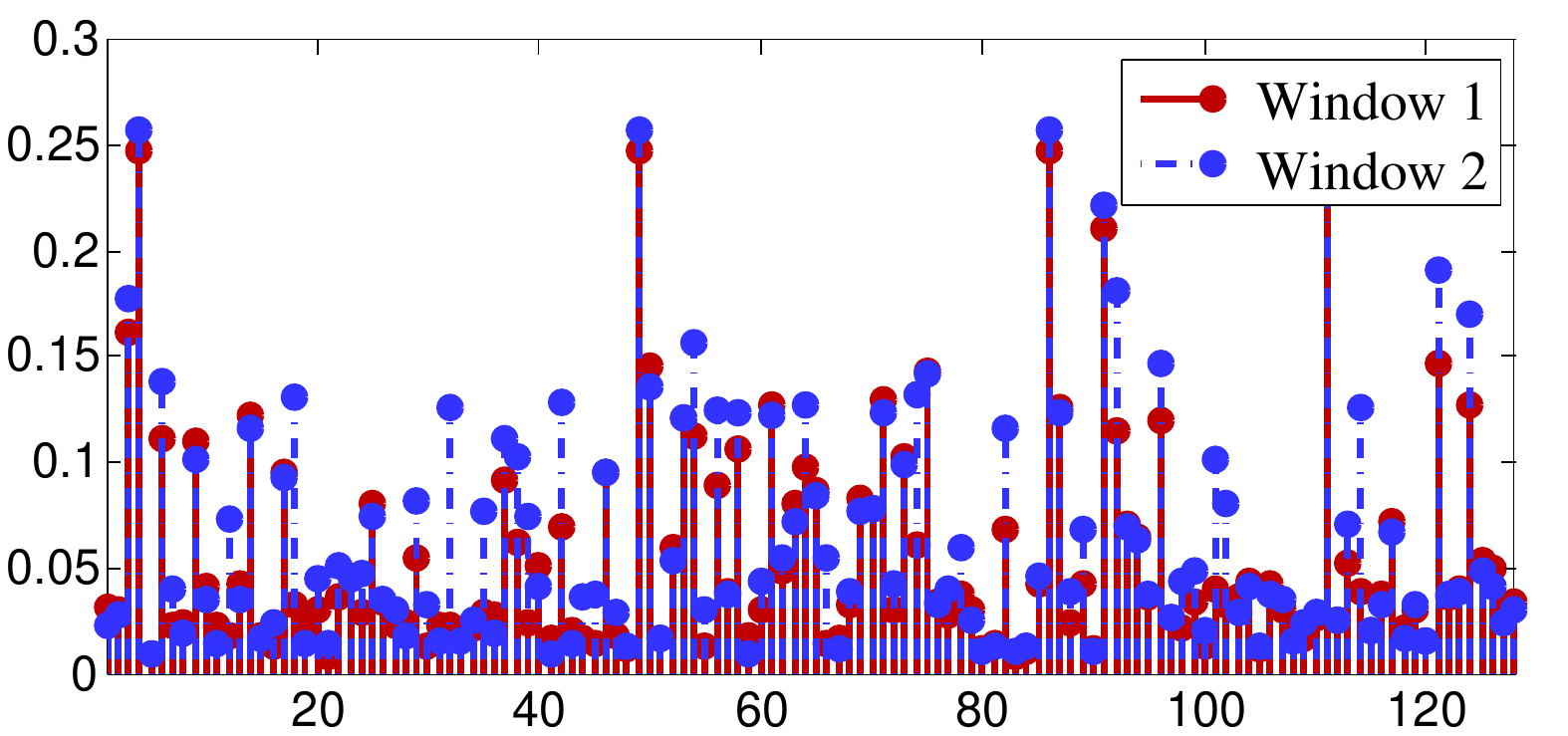}}\hfill
	\vspace{-10pt}
	\caption{Visualization of support window pairs on multi-spectral RGB and NIR images 
		denoted as `A' in \figref{img:2} having gradient 
			orientation variations, and descriptors for these window pairs. 
			Conventional descriptors such as DAISY \cite{Tola10},
			BRIEF \cite{Calonder11}, and LSS \cite{Schechtman07} vary across modality
			variations.
		Unlike those methods, our DASC descriptor remains unchanged to modality variations.}\label{img:6}\vspace{-10pt}
\end{figure}
\subsection{Adaptive Self-Correlation Measure}\label{sec:42}
With estimated sampling patterns $(s_{i,l},t_{i,l})$, the DASC
descriptor measures a patch similarity using an adaptive
self-correlation (ASC) measure in order to robustly encode a local
internal layout of self-similarities. For the sake of simplicity, we
omit $(i,l)$ in the correlation metric from here on, as it is
repeatedly computed for all $(i,l)$. For $(s,t) \in
\Lambda^{\mathrm{dasc}}$, the adaptive self-correlation
${\Psi(s,t)}$ between two patches $\mathcal{F}_{s}$ and
$\mathcal{F}_{t}$ centered at pixels $s$ and $t$ is computed as
follows:
\begin{equation}\label{equ:asc}
\Psi (s,t) = \frac{{\sum\limits_{s',t'} {{\omega _{s,s'}}{\omega
_{t,t'}}({f_{s'}} - {\mathcal{G}_{s}})({f_{t'}} - {\mathcal{G}_{t}})} }}{{\sqrt
{\sum\limits_{s'} {{{\{{\omega _{s,s'}}({f_{s'}} - {\mathcal{G}_s})\}}^2}} }
\sqrt {\sum\limits_{t'} {{{\{{\omega _{t,t'}}({f_{t'}} - {\mathcal{G}_t})\}}^2}}
} }},
\end{equation}
where $s' \in \mathcal{F}_{s}$ and $t' \in\mathcal{F}_{t}$ and
weighted averages on $\mathcal{F}_{s}$ and $\mathcal{F}_{t}$ are defined as ${\mathcal{G}_s}=\sum\nolimits_{s'} {{\omega_{s,s'}}{f_{s'}}}$
and ${\mathcal{G}_t}=\sum\nolimits_{t'} {{\omega_{t,t'}}{f_{t'}}}$.

The weight ${\omega _{s,s'}}$ represents how similar two pixels $s$
and $s'$ are, and is normalized, \emph{i.e.}, $\sum\nolimits_{s'}
{{\omega _{s,s'}}}=1$. It can be defined with any kind of edge-aware
weights \cite{Yang09,He13,Gastal11}. This weighted sum better
handles outliers and local variations in patches compared to other
patch-wise similarity metrics. It is worth noting that the adaptive
self-correlation used here is conceptually similar to the ANCC
\cite{Heo}, but our descriptor employs the correlation metric for
measuring self-similarity within a single image which is
used for matching two or more images later, while the ANCC is used
to directly measure inter-similarity between different
images.

Finally, our patch-wise similarity between ${\mathcal{F}_s}$ and
${\mathcal{F}_t}$ is computed with a truncated exponential function,
which has been widely used in the robust estimator \cite{Black98}:
\begin{equation}\label{equ:6}
\mathcal{C}(s,t) = \max (\exp ( - (1 - \left| {\Psi (s,t)} \right|)/\sigma_c ),\tau_c ),
\end{equation}
where $\sigma_c$ is a bandwidth of Gaussian kernel and $\tau_c$ is a
truncation parameter. Here, a absolute value of ${\Psi (s,t)}$ is
used to mitigate the effect of intensity reverses. The correlation
$\mathcal{C}(s_{i,l},t_{i,l})$ for $i$ is normalized with an unit
norm for all $l$. 

\figref{img:6} represents examples of visualizing the results of various descriptors. 
The conventional descriptors show the sensitivity to modality variations, 
however the DASC shows the robustness against multi-modal variations.
\vspace{-5pt}
\subsection{Efficient Computation for Dense Descriptor}\label{sec:44}
\begin{figure}[!t]
	\centering
	\renewcommand{\thesubfigure}{}
	\subfigure[]
	{\includegraphics[width=1\linewidth]{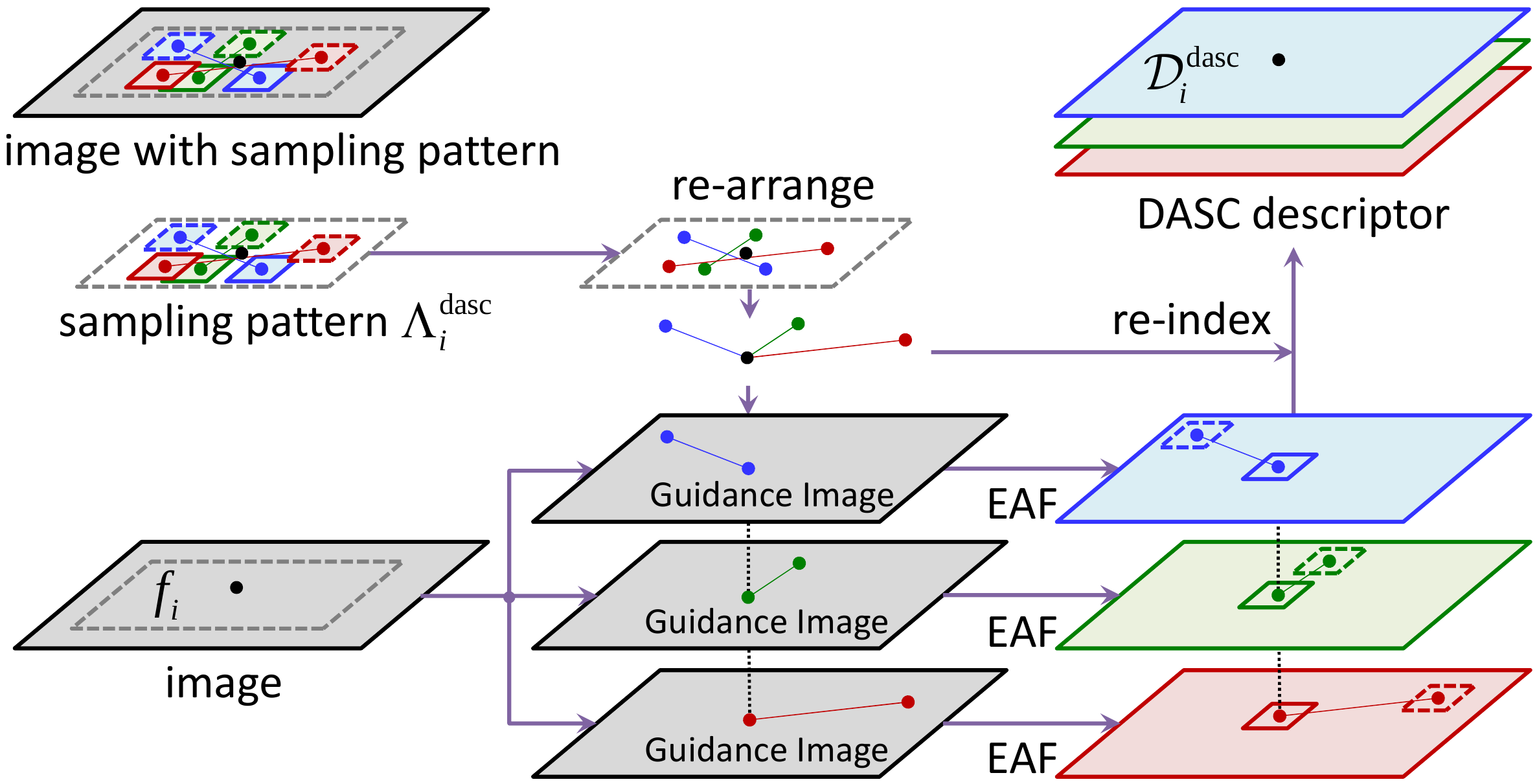}}
	\vspace{-25pt}
	\caption{Efficient computation framework of the DASC descriptor.
		In order to reduce a computational load in computing the adaptive self-correlation,
		it re-arranges the sampling pattern and employs fast EAF scheme.
		The DASC descriptor is then computed with re-indexing.}\label{img:7}\vspace{-10pt}
\end{figure}
For densely constructing our descriptor on an entire image, we should
compute $\mathcal{C}(s_{i,l},t_{i,l})$ for all patch pairs belonging to $(s_{i,l},t_{i,l}) \in \Lambda_{i}^{\mathrm{dasc}}$ for each pixel $i$. Thus, a
straightforward computation can be extremely time-consuming.
In this section, we present an efficient method for computing the DASC descriptor. To compute all weighted sums in \equref{equ:asc}
for $(s_{i,l},t_{i,l})$ efficiently, we employ a constant-time
edge-aware filter (EAF), \emph{e.g.}, the guided filter (GF) \cite{He13}. However, the symmetric
weight $w_{s,s'}w_{t,t'}$ varies for each $l$, and thus computing
the numerator in \equref{equ:asc} is still very time-consuming.

To alleviate these limitations, we simplify \equref{equ:asc} by considering only the weight $w_{s,s'}$
from the source patch $\mathcal{F}_s$ so that a fast computation of
\equref{equ:asc} using fast edge-aware filter is feasible. It should
be noted that such an asymmetric weight approximation also has been
used in cost aggregation for stereo matching \cite{Rhemann11}. We also
found that in our descriptor, a performance gap between using the
asymmetric weight $w_{s,s'}$ and the symmetric weight
$w_{s,s'}w_{t,t'}$ is negligible, which will be shown in \secref{sec:441}.
For efficient description, we also re-arrange the sampling pattern $(s_{i,l},t_{i,l})$ to
referenced-biased pairs $(i,j)=(i,i+t_{i,l}-s_{i,l})$.
\equref{equ:asc} is then approximated as follows:
\begin{equation}\label{equ:asasc}
\tilde{\Psi} (i,j) = \frac{{\sum\limits_{i',j'} {{\omega _{i,i'}}({f_{i'}} - {\mathcal{G}_{i}})({f_{j'}} - {\mathcal{G}_{i,j}})} }}{{\sqrt
{\sum\limits_{i'} {{{\omega _{i,i'}}{({f_{i'}} - {\mathcal{G}_i})}^2}} }
\sqrt {\sum\limits_{i',j'} {{{\omega _{i,i'}}{({f_{j'}} - {\mathcal{G}_{i,j}})}^2}} } }},
\end{equation}
where ${\mathcal{G}_{i}}=\sum\nolimits_{i'} {{\omega_{i,i'}}{f_{i'}}}$.
Furthermore, ${\mathcal{G}_{i,j}}=\sum\nolimits_{i',j'}{{\omega _{i,i'}}{f_{j'}}}$
which means weighted average of $f_{j'} \in \mathcal{F}_j$ with a guidance image $f_{i'} \in \mathcal{F}_i$.
It is worth noting that the robustness of $\Psi (s,t)$ can be still applied to 
$\tilde{\Psi} (i,j)$ since their difference is just weight factors.
\begin{table}[!t]
	\begin{center}
		\begin{tabularx}{\linewidth}{p{1mm} p{0.02mm}| p{80mm}}
			\hlinewd{0.8pt}
			\multicolumn{3}{ p{85mm} }{{\bf Algorithm 1}: Dense Adaptive Self-Correlation (DASC)}\\
			\hlinewd{0.8pt}
			\multicolumn{3}{ p{85mm} }{{\bf Input} : image ${f_i}$, candidate sampling patterns $ \Lambda_{i} $, training patch pairs dataset $\mathcal{P}$.}\\
			\multicolumn{3}{ p{85mm} }{{\bf Output} : the DASC descriptor volume ${\mathcal{D}^{\mathrm{dasc}}_{i}}$.}\\
			\\
			~&\multicolumn{2}{ p{80mm} }{\emph{$/*$ Offline Procedure $*/$}} \\
			$\mathbf{1:}$&\multicolumn{2}{ p{80mm} }{Compute ${\mathbf{r}_h}$ using \equref{equ:8} for possible candidate sampling patterns $ \Lambda_{i} $ on training support window pairs $\mathcal{P}$.} \\
			$\mathbf{2:}$&\multicolumn{2}{ p{80mm} }{Learn a weight ${\mathbf{v}}$ by optimizing \equref{equ:10}.} \\
			$\mathbf{3:}$&\multicolumn{2}{ p{80mm} }{Select the maximal $L^{\mathrm{dasc}}$ sampling patterns $(s_{i,l},t_{i,l})$ in terms of $\left| {v_l} \right|$, denoted as $\Lambda_{i}^{\mathrm{dasc}}$.} \\
			\\
			~&\multicolumn{2}{ p{80mm} }{\emph{$/*$ Online Procedure $*/$}} \\
			$\mathbf{4:}$&\multicolumn{2}{ p{80mm} }{Compute ${\mathcal{G}_{i}} = \sum\nolimits_{i'} {{\omega _{i,i'}}{f_{i'}}} $ for all pixel $i$.}\\
			$\mathbf{5:}$&\multicolumn{2}{ p{80mm} }{Compute ${\mathcal{G}_{i^{2}}} = \sum\nolimits_{i'} {{\omega _{i,i'}}f_{i'}^2} $.}\\
			~&\multicolumn{2}{ p{80mm} }{{\bf for } $l = 1:L^{\mathrm{dasc}}$ {\bf do }}\\
			$\mathbf{6:}$&~&  Re-arrange $(s_{i,l},t_{i,l}) \in \Lambda_{i}^{\mathrm{dasc}}$ as $(i,j) = (i,i+t_{i,l}-s_{i,l})$.\\
			$\mathbf{7:}$&~&  Compute ${\mathcal{G}_{i,ij}} = \sum\nolimits_{i',j'} {{\omega _{i,i'}}{f_{i'}}{f_{j'}}} $.\\
			$\mathbf{8:}$&~&  Compute ${\mathcal{G}_{i,j}} = \sum\nolimits_{i',j'} {{\omega _{i,i'}}{f_{j'}}} $.\\
			$\mathbf{9:}$&~&  Compute ${\mathcal{G}_{i,j^{2}}} = \sum\nolimits_{i',j'} {{\omega _{i,i'}}f_{j'}^2} $.\\
			$\mathbf{10:}$&~&  Estimate ${\tilde{\Psi} (i,i')}$ and $\mathcal{C}(i,i')$ using \equref{equ:asasc} and \equref{equ:6}.\\
			$\mathbf{11:}$&~&  Compute the DASC descriptor ${\mathcal{D}^{\mathrm{dasc}}_{i}}$ by re-indexing sampling patterns such that ${d^{\mathrm{dasc}}_{i,l}} = \mathcal{C}(s_{i,l},t_{i,l})$.\\
			~&\multicolumn{2}{ p{80mm} }{{\bf end for }}\\
			\hlinewd{0.8pt}
		\end{tabularx}
	\end{center}\label{alg:1}\vspace{-10pt}
\end{table}

We then decompose numerator and denominator in \equref{equ:asasc} after some
arithmetic derivations such that
\begin{equation}\label{equ:asasc_app}
\frac{{{\mathcal{G}_{i,ij}}  - {\mathcal{G}_{i}}  \cdot {\mathcal{G}_{i,j}} }}
{{\sqrt {{\mathcal{G}_{i^{2}}} - {{{\mathcal{G}_{i}^2}}}} \cdot \sqrt {{\mathcal{G}_{i,j^{2}}}  - {{{\mathcal{G}_{i,j}^2}}}} }},
\end{equation}
where ${\mathcal{G}_{i^{2}}} = \sum\nolimits_{i'} {{\omega
_{i,i'}}f_{i'}^2} $, ${\mathcal{G}_{i,ij}}=\sum\nolimits_{i',j'}
{{\omega _{i,i'}}{f_{i'}}{f_{j'}}}$, and
${\mathcal{G}_{i,j^{2}}}=\sum\nolimits_{i',j'} {{\omega
_{i,i'}}{f_{j'}^{2}}}$.
While the ${\mathcal{G}_{i}}$ and ${\mathcal{G}_{i^{2}}}$ can be
computed on image domain once, ${\mathcal{G}_{i,ij}}$,
${\mathcal{G}_{i,j}}$, and ${\mathcal{G}_{i,j^{2}}}$ should be
computed on each offset. However, the weight $\omega_{i,i'}$ is
fixed for all offsets, thus it can be shared in all offsets.
All these components can be efficiently computed using a
constant-time edge-aware filter (EAF) \cite{He13}. Finally, the dense
descriptor $\mathcal{D}^{\mathrm{dasc}}_{i}$ is computed with
re-indexing as ${d^{\mathrm{dasc}}_{i,l}} =
\mathcal{C}(s_{i,l},t_{i,l})$ though the robust function in
\equref{equ:6}. \figref{img:7} describes our efficient method for
computing the DASC descriptor. Algorithm 1 summarizes
the efficient computation of the DASC descriptor. \vspace{-5pt}
\subsubsection{Comparison of symmetric and asymmetric version of adaptive self-correlation measure}\label{sec:441}
This section analyzes the performance of the DASC descriptor when
using the symmetric weight $\omega_{s,s'}\omega_{t,t'}$ of $\Psi (s,t)$ in
\equref{equ:asc} and with the asymmetric weight $\omega_{i,{i'}}$ of $\tilde{\Psi} (i,j)$ in
\equref{equ:asasc}.
The symmetric weight case in the DASC can also be computed similar to \secref{sec:44}.
After re-arranging the sampling pattern as $(i,j)=(i,i+t_{i,l}-s_{i,l})$,
the \equref{equ:asc} can be then
decomposed as similar in \equref{equ:asasc_app}
\begin{equation}
\Psi(i,j) = \frac{{{\mathcal{G}_{ij,ij}} - {\mathcal{G}_{ij,i}} - {\mathcal{G}_{ij,j}}  + {\mathcal{G}_{i}}  \cdot {\mathcal{G}_{j}} }}
{{\sqrt {{\mathcal{G}_{i^{2}}} - {{{\mathcal{G}_{i}^2}}}} \cdot \sqrt {{\mathcal{G}_{j^{2}}}  - {{{\mathcal{G}_{j}^2}}}} }},
\end{equation}
where ${\mathcal{G}_{ij,ij}} = \sum\nolimits_{i'} {{\omega_{i,i'}}{\omega_{j,j'}}f_{i'}^2 f_{j'}^2} $,
${\mathcal{G}_{ij,i}}=\sum\nolimits_{i',j'}{{\omega _{i,i'}}{\omega _{j,j'}}{f_{i'}}}$, and
${\mathcal{G}_{ij,j}}=\sum\nolimits_{i',j'}{{\omega _{i,i'}}{\omega _{j,j'}}{f_{j'}}}$.
The denominator can be easily computed on overall image once.
However, compared to the asymmetric measure in \equref{equ:asasc},
$\omega_{i,i'}\omega_{j,j'}$ in ${\mathcal{G}_{ij,ij}}$, ${\mathcal{G}_{ij,i}}$, and ${\mathcal{G}_{ij,j}}$ varies for each $l$.
Furthermore, it should be computed with a range distance using 6-D vector (or 2-D vector), when an input is a color image
(or an intensity image). It significantly increases a computational burden
needed for employing constant-time EAFs \cite{He13,Paris09}. A performance gap between using the symmetric measure $\Psi (s,t)$ and the asymmetric measure $\tilde{\Psi} (i,j)$ in the DASC descriptor is negligible, which will be shown in \secref{sec:441}. \vspace{-5pt}
\begin{figure}[!t]
\centering
\renewcommand{\thesubfigure}{}
\subfigure[]
{\includegraphics[width=1\linewidth]{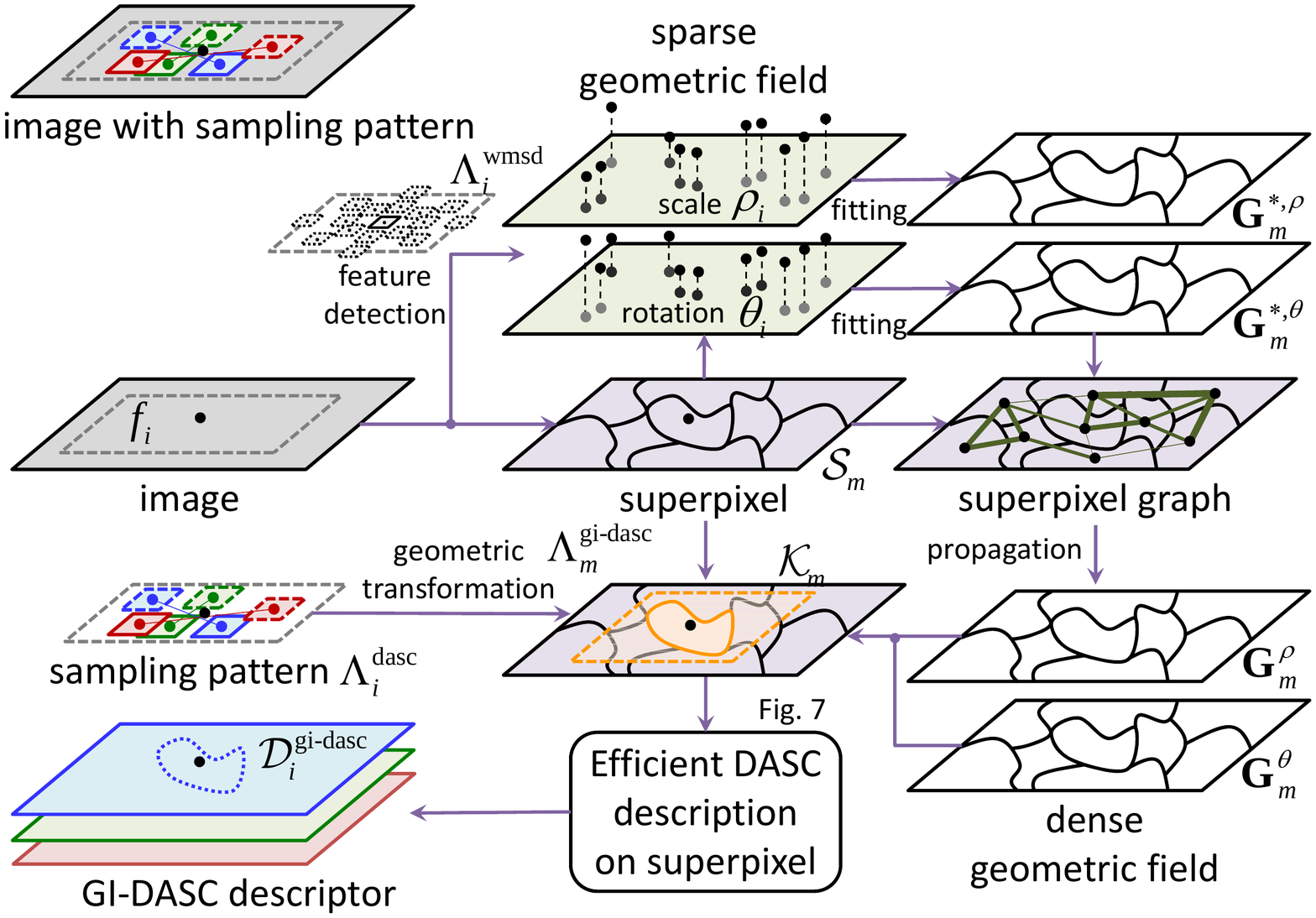}}
\vspace{-25pt}
\caption{Efficient computation framework of the geometry-invariant DASC (GI-DASC) descriptor.
To leverage the efficient computation scheme of the DASC, we employ a superpixel-based description
with inferred geometric fields on each superpixel using the WMSD detection.}\label{img:9}\vspace{-10pt}
\end{figure}

\subsection{Computational Complexity Analysis}\label{sec:45}
The computational complexity of the DASC descriptor on the
brute-force implementation becomes $O(INL)$, where $I$, $N$, and $L$
represent an image size, a patch size, and a descriptor dimension, respectively.
With our efficient computation model, our approach removes the complexity dependency on the patch size $N$,
\emph{i.e.}, $O(IL)$ due to fast constant-time EAF. Furthermore,
since there exist repeated offsets, the complexity is further
reduced as $O(I\tilde{L})$ for $\tilde{L} < L$. \vspace{-5pt}

\section{Geometry-Invariant DASC Descriptor}\label{sec:5}
Similar to the DAISY \cite{Tola10}, the DASC descriptor is not
appropriate to deal with geometric variations. In this section, we
propose the geometry-invariant DASC descriptor, called GI-DASC, that
addresses severe geometric variations as well as image modality
variations. A key idea is to geometrically transform sampling
patterns used to measure the patch similarity according to scale and
rotation fields when computing the DASC descriptor. To estimate the
scale and rotation fields, we first infer initial geometric fields
only for sparse points. These initial fields are then fitted and propagated through a
superpixel graph. Finally, the GI-DASC descriptor is efficiently computed with
geometrically transformed sampling patterns in a manner similar to
computing the DASC descriptor, except the fact that the descriptor
computation is done for each superpixel independently.
\begin{figure}[t]
\centering
\renewcommand{\thesubfigure}{}
\subfigure[(a) Sampling patterns $\Lambda^{\mathrm{wmsd}}_i$]
{\includegraphics[width=0.5\linewidth]{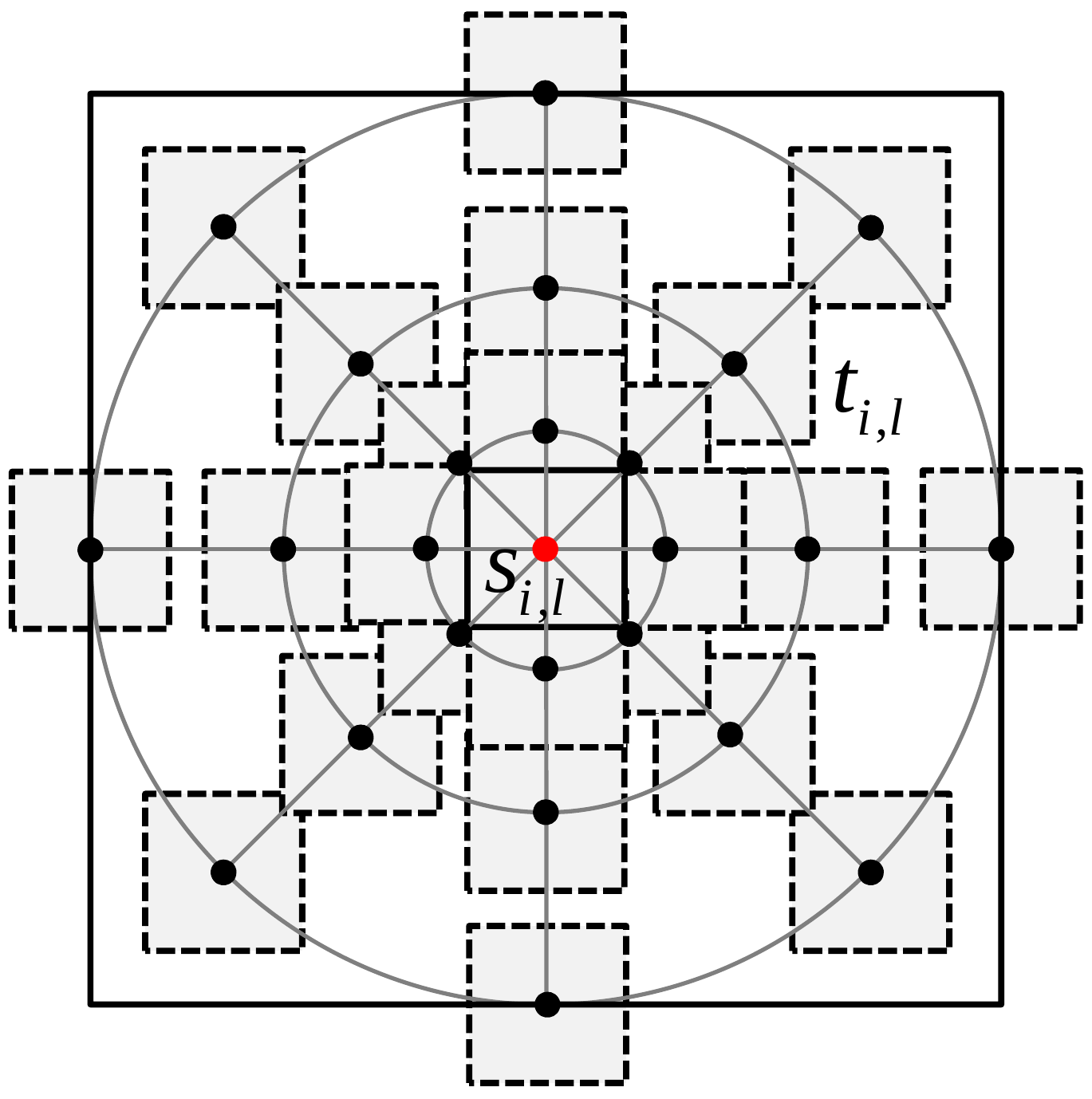}}\hfill
\subfigure[(b) Index set ${\Pi^o_i}$ ]
{\includegraphics[width=0.5\linewidth]{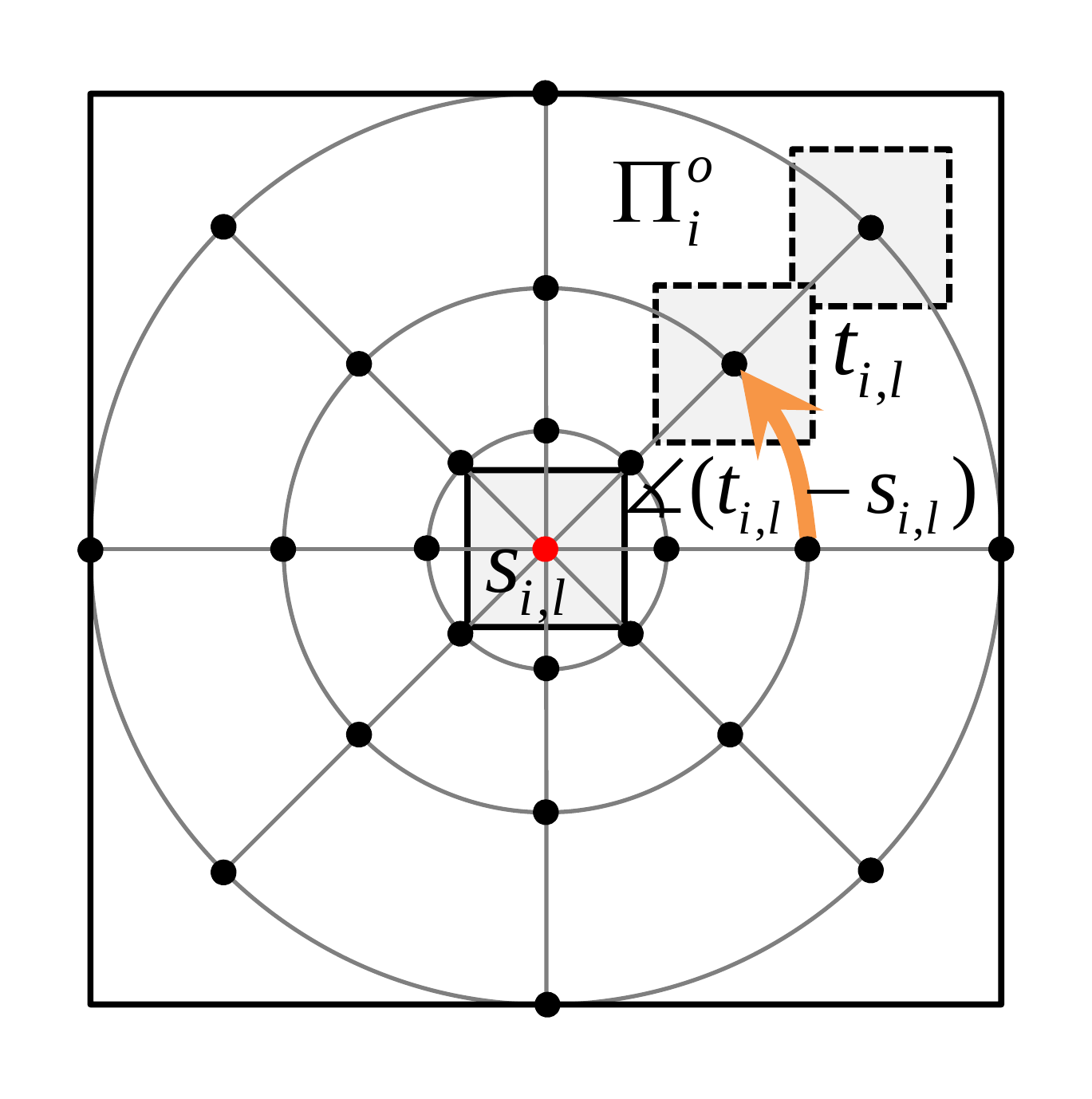}}\hfill
\vspace{-10pt}
\caption{Demonstration of sampling patterns $(s_{i,l},t_{t,l}) \in \Lambda^{\mathrm{wmsd}}_i$ for the WMSD detector
and the index set for the $o$ most smallest value ${\Pi^o_i}$.
It enables us to extract reliable feature points $i \in \mathcal{I'}$ with corresponding geometric fields (scale $\rho_i$ and rotation $\theta_i$).}\label{img:10}\vspace{-5pt}
\end{figure}

Adopting the superpixel-based geometry field inference has the
following three reasons. First, the reliable geometry field
can be estimated reliably only at distinctive pixels.
Second, the geometric fields tend to vary smoothly, except object
boundaries. Third, the transformed sampling
patterns should be fixed for each superpixel so that the
computational scheme based on the fast EAF \cite{He13} can be
used for efficiently obtaining the GI-DASC for each superpixel. 
\figref{img:9} represents the overview of the GI-DASC. \vspace{-5pt}

\subsection{Initial Sparse Geometric Field Inference}\label{sec:51}
Conventional feature detectors, \emph{e.g.}, SIFT \cite{Lowe04}, 
are very sensitive to multi-modal and
multi-spectral deformation. In order to extract sparse features with
distinctive geometric information available, we employ maximal
self-dissimilarity (MSD) thanks to its robustness for modality
deformation \cite{Tombari14}. We propose weighted MSD (WMSD) that
improves the performance of the MSD in terms of both complexity and
robustness by employing an weighted similarity measure and an
efficient computation scheme similar to the DASC.

Similar to $\Gamma_i$ used in the DASC, the log-polar circular
point set $\Gamma^{\mathrm{wmsd}}_i$ is defined for feature detector.
The sampling pattern $\Lambda^{\mathrm{wmsd}}_i$ is then defined
in such a way that the source patch is always located at center pixel and the target
patches are located at other neighboring points as shown in
\figref{img:10}(a). In order to consider the scale deformation, we build the Gaussian
image pyramid $u^k_i = f_i * \varrho_k$ for $k = 1,...,N_k$, where
$\varrho_k$ is the $k$-th Gaussian kernel with a sigma $\rho_k$ and $N_k$ is the number of
pyramids. After re-arranging the sampling pattern as
$(i,j)=(i,i+t_{i,l}-s_{i,l})$, The
self-dissimilarity measure ${\Phi^{k}(i,l)}$
for $l = 1,...,L^{\mathrm{wmsd}} (=N^{\mathrm{wmsd}}_\rho \times N^{\mathrm{wmsd}}_\theta)$ is
computed using weighted sum of squared difference (SSD) with a guidance image $u^k_{i'}$ such that
\begin{equation}\label{equ:wmsd}
\begin{split}
{\Phi^{k}(i,l)} &= \sum\nolimits_{i',j'} {{\omega _{i,{i'}}}{{({u^k_{i'}} - {u^k_{j'}})}^2}}  \\
&= \mathcal{U}^k_{i^2} + \mathcal{U}^k_{i,j^2} - 2\mathcal{U}^k_{i,{ij}},
\end{split}
\end{equation}
where ${\mathcal{U}^k_{i^{2}}} = \sum\nolimits_{i'} {{\omega
_{i,i'}}(u^{k}_{i'})^2}$, ${\mathcal{U}^k_{i,j^2}} =
\sum\nolimits_{i',j'} {{\omega _{i,i'}}(u^{k}_{j'})^2}$, and
${\mathcal{U}^k_{i,ij}} = \sum\nolimits_{i',j'} {{\omega _{i,i'}}
u^k_{i'} u^k_{j'}}$. Similar to the DASC, \eqref{equ:wmsd} can be
computed efficiently using constant time EAF \cite{He13,Paris09}.

We extract the index set ${\Pi^o_i}$ for the $o$ most smallest value
${\Psi^{\mathrm{wmsd},k}_{i,l}}$ for all $l$, \emph{i.e.,} $o$
nearest neighbors for center patch in \figref{img:10}(b). 
It should be noted that parameter $o$ trades distinctiveness and 
computational efficiency \cite{Tombari14}. 
We then compute feature response map $\Omega^k_i$ by estimating the
summation of $\Phi^{k}(i,l)$ for ${l\in{\Pi^o_i}}$
such that
\begin{equation}
\Omega^k_i = \sum\nolimits_{l\in{\Pi^o_i}}{\Phi^{k}(i,l)}.
\end{equation}

For feature response maps $\mathbf{\Omega}_i = \{\Omega^k_i\}$,
the local maxima are obtained by the non maximal suppression, which compares $\Omega^k_i$ to its $8$ neighbors
on the current scale and $18$ neighbors on the $(k+1)^{th}$ and
$(k-1)^{th}$ scales. Similar to SIFT \cite{Lowe04}, a feature point
$i \in \mathcal{I'}$ is detected only if $\{\Omega^k_i\}$ has an
extreme value compared to all of these neighbors,
and its scale $\rho_i$ is defined with $\rho_k$, 
where $\mathcal{I'} \subset \mathcal{I}$ is a sparse discrete image domain.
 
A canonical orientation is further associated to $i \in \mathcal{I'}$ by
constructing a histogram with angles
$\measuredangle\left(t_{i,l}-s_{i,l}\right)$ for ${l\in{\Pi^o_i}}$
weighted by ${\Phi^{k}(i,l)}$ as
\begin{equation}\label{equ:histang}
l_\mathrm{hist}(i,\theta) = \sum\nolimits_{l\in{\Pi^o_i}}{{\Phi^{k}(i,l)} \cdot \delta(\measuredangle\left(t_{i,l}-s_{i,l}\right)-\theta)},
\end{equation}
where $\delta$ is the Kronecker delta function.
Then, we simply choose the direction
corresponding to the highest bin in the histogram,
\emph{i.e.,} $\theta_i = \mathrm{argmax}_\theta l_\mathrm{hist}(i,\theta)$.
The WMSD detector is summarized in Algorithm 2. \vspace{-5pt}
\begin{table}[t]
	\begin{center}
		\begin{tabularx}{\linewidth}{p{1mm} p{0.02mm}| p{0.02mm}| p{80mm}}
			\hlinewd{0.8pt}
			\multicolumn{4}{ p{85mm} }{{\bf Algorithm 2}: Weighted Maximal Self-Dissimilarity (WMSD)} \\
			\hlinewd{0.8pt}
			\multicolumn{4}{ p{85mm} }{{\bf Input} : image ${f_i}$, feature detection sampling patterns $\Lambda_{i}^{\mathrm{det}}$.}\\
			\multicolumn{4}{ p{85mm} }{{\bf Output} : feature points $i\in \mathcal{I'}$ with scale $\rho_i$, rotation $\theta_i$.}\\
			\\
			~&\multicolumn{3}{ b{80mm} }{{\bf for } $k = 1:N_k$ {\bf do }}\\
			$\mathbf{1:}$&~&  \multicolumn{2}{ p{80mm} }{Compute $u^k_i = f_i * \varrho_k$ with the Gaussian kernel $\varrho_k$.} \\
			$\mathbf{2:}$&~&  \multicolumn{2}{ p{80mm} }{Compute ${\mathcal{U}^k_{i^{2}}} = \sum\nolimits_{i'} {{\omega _{i,i'}}(u^{k}_{i'})^2}$ for all pixel $i$.} \\
			~&~&  \multicolumn{2}{ b{80mm} }{{\bf for } $l = 1:L_{\mathrm{wmsd}}$ {\bf do }}\\
			$\mathbf{3:}$&~&  ~&  Compute ${\mathcal{U}^k_{i,j^2}} = \sum\nolimits_{i',j'} {{\omega _{i,i'}}(u^{k}_{j'})^2}$ for $j=i+t_{i,l}-s_{i,l}$. \\
			$\mathbf{4:}$&~&  ~&  Compute ${\mathcal{U}^k_{i,ij}} = \sum\nolimits_{i',j'} {{\omega _{i,i'}} u^k_{i'} u^k_{j'}}$. \\
			$\mathbf{5:}$&~&  ~&  Estimate ${\Phi^{k}(i,l)} 
			= \mathcal{U}^k_{i^2} + \mathcal{U}^k_{i,j^2} - 2\mathcal{U}^k_{i,ij}$. \\
			~&~&  \multicolumn{2}{ b{80mm} }{{\bf end for }}\\
			$\mathbf{6:}$&~&  \multicolumn{2}{ p{80mm} }{Extract the index set ${\Pi^o_i}$ among ${{\Phi^{k}(i,l)}}$ for all $l$.} \\
			$\mathbf{7:}$&~&  \multicolumn{2}{ p{80mm} }{Build response map as $\Omega^k_i = \sum\nolimits_{l\in{\Pi^o_i}}{{\Phi^{k}(i,l)}}$.} \\
			~&\multicolumn{3}{ b{80mm} }{{\bf end for }}\\
			$\mathbf{8:}$&\multicolumn{3}{ p{80mm} }{Detect feature points $i \in \mathcal{I'}$ from $\mathbf{\Omega} = \{\Omega^k\}$ with scale factor $\rho_i$.} \\
			$\mathbf{9:}$&\multicolumn{3}{ p{80mm} }{Compute the orientation $\theta_i$ for $i$ from $l_\mathrm{hist}(i,\theta)$.} \\
			\hlinewd{0.8pt}
		\end{tabularx}
	\end{center}\label{alg:2}\vspace{-10pt}
\end{table}
\subsection{Superpixel Graph-Based Propagation}\label{sec:52}
In order to infer dense geometric fields from sparse geometric
fields ($\rho_i$ and $\theta_i$ for $i \in \mathcal{I'}$), we
decompose the image $f$ as superpixel $\mathcal{S}=\{\mathcal{S}_m |
\bigcup_m \mathcal{S}_m = \mathcal{I}\,\mathrm{and}\,\forall m\neq
n,\,\mathcal{S}_m \bigcap \mathcal{S}_n \neq\varnothing,\,m \in
1,...,N_m \}$, where $N_m$ is the number of superpixels. The
geometric field $\mathrm{G}^{*,\rho}_m$ and
$\mathrm{G}^{*,\theta}_m$ are fitted on each superpixel
$\mathcal{S}_m$ as the average of sparse geometric fields
$\rho_i$ and $\theta_i$ for $i\in \{\mathcal{I'} \bigcap
\mathcal{S}_m\}$. Note that this fitting operation is performed only
when $\{\mathcal{I'} \bigcap \mathcal{S}_m\}$ exists, i.e., the
superpixel includes sparse feature points (at least, 1). Finally,
the $\mathbf{G}^{*,\rho} = {\bigcup_{m}}{\mathrm{G}^{*,\rho}_m} \in
\mathbb{R}^{N_m}$ and $\mathbf{G}^{*,\theta} =
{\bigcup_{m}}{\mathrm{G}^{*,\theta}_m}$ are constructed for all
superpixels.

Similar to \cite{Kim14a}, our approach then formulates an inference
of dense geometric fields $\mathbf{G}^{\rho}$ and
$\mathbf{G}^{\theta}$ as a constrained optimization problem where
surface-fitted sparse geometric fields $\mathbf{G}^{*,\rho}$ and
$\mathbf{G}^{*,\theta}$ are interpreted as soft constraints. For the
sake of simplicity, we omit ${\rho}$ and ${\theta}$ since they can
be computed using the same method. The energy function of our
superpixel-based propagation is defined as follows:
\begin{equation}\label{equ:spenergy}
\sum\limits_{m} \left\{ {p^{\mathrm{sp}}_m}{{{({\mathrm{G}_m} - \mathrm{G}_m^*)}^2}}
+ \mu\sum\limits_{n \in {\mathcal{N}_m}} {{\omega^{\mathrm{sp}}_{mn}}({\mathrm{G}_m} - {\mathrm{G}_n})^2} \right\},
\end{equation}
where $\mu$ is a regularization parameter. Here, the first term
encodes the dissimilarity between final geometric fields ${\mathrm{G}_m}$
and initial sparse geometric fields ${\mathrm{G}^*_m}$. ${p^{\mathrm{sp}}_m}$ is
an index function, which is 1 for valid (constraint) superpixel, and
0 otherwise. The second term imposes the constraint that two
adjacent superpixels ${m}$ and ${n \in {\mathcal{N}_m}}$ may have
similar geometric fields according to surperpixel feature affinity
${\omega^{\mathrm{sp}}_{mn}}$, which will be described in the
following section. \vspace{-5pt}
\begin{figure}[t]
	\centering
	\renewcommand{\thesubfigure}{}
	\subfigure[(a) Image 1]
	{\includegraphics[width=0.25\linewidth]{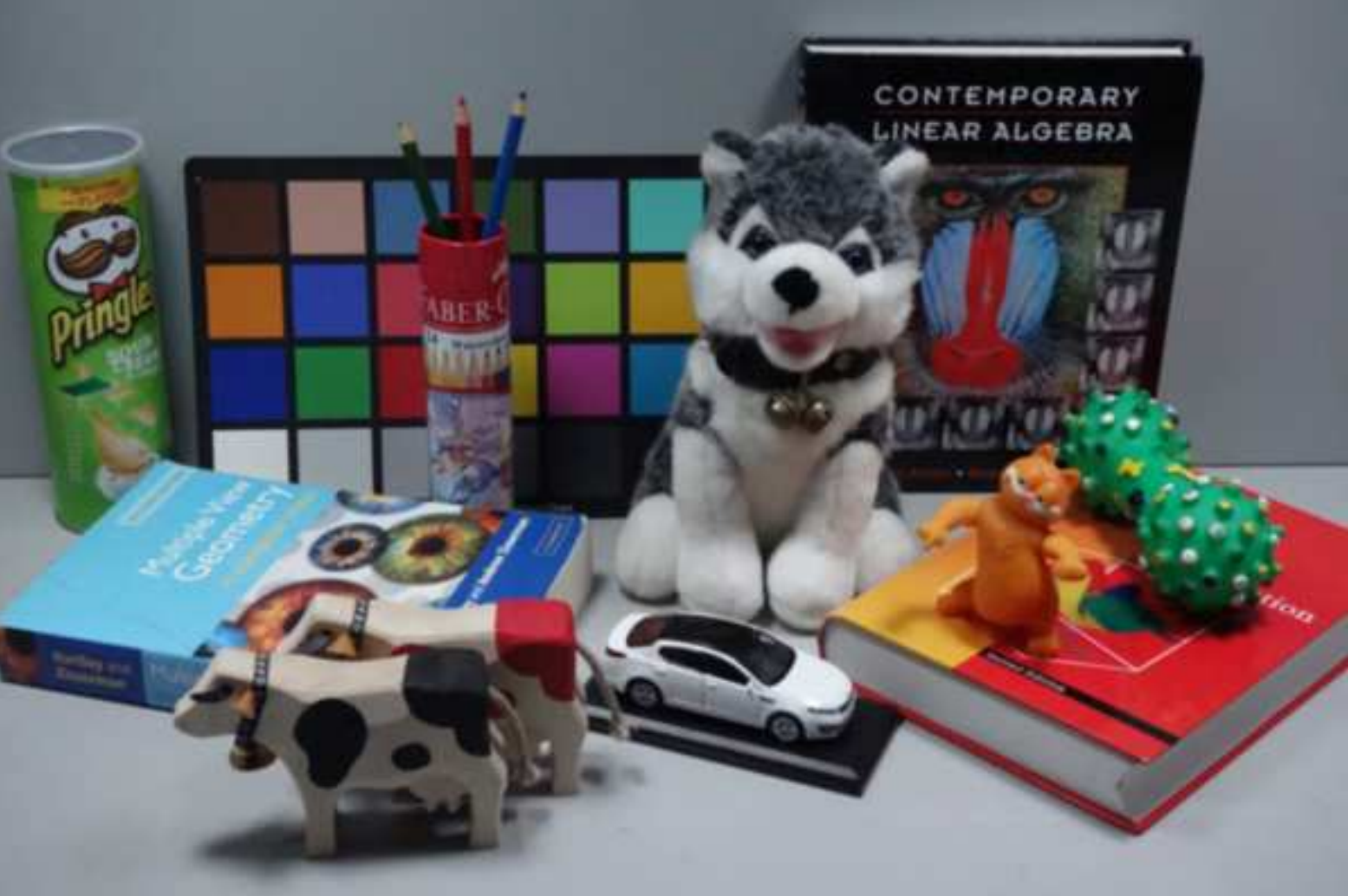}}\hfill
	\subfigure[(b) Image 2]
	{\includegraphics[width=0.25\linewidth]{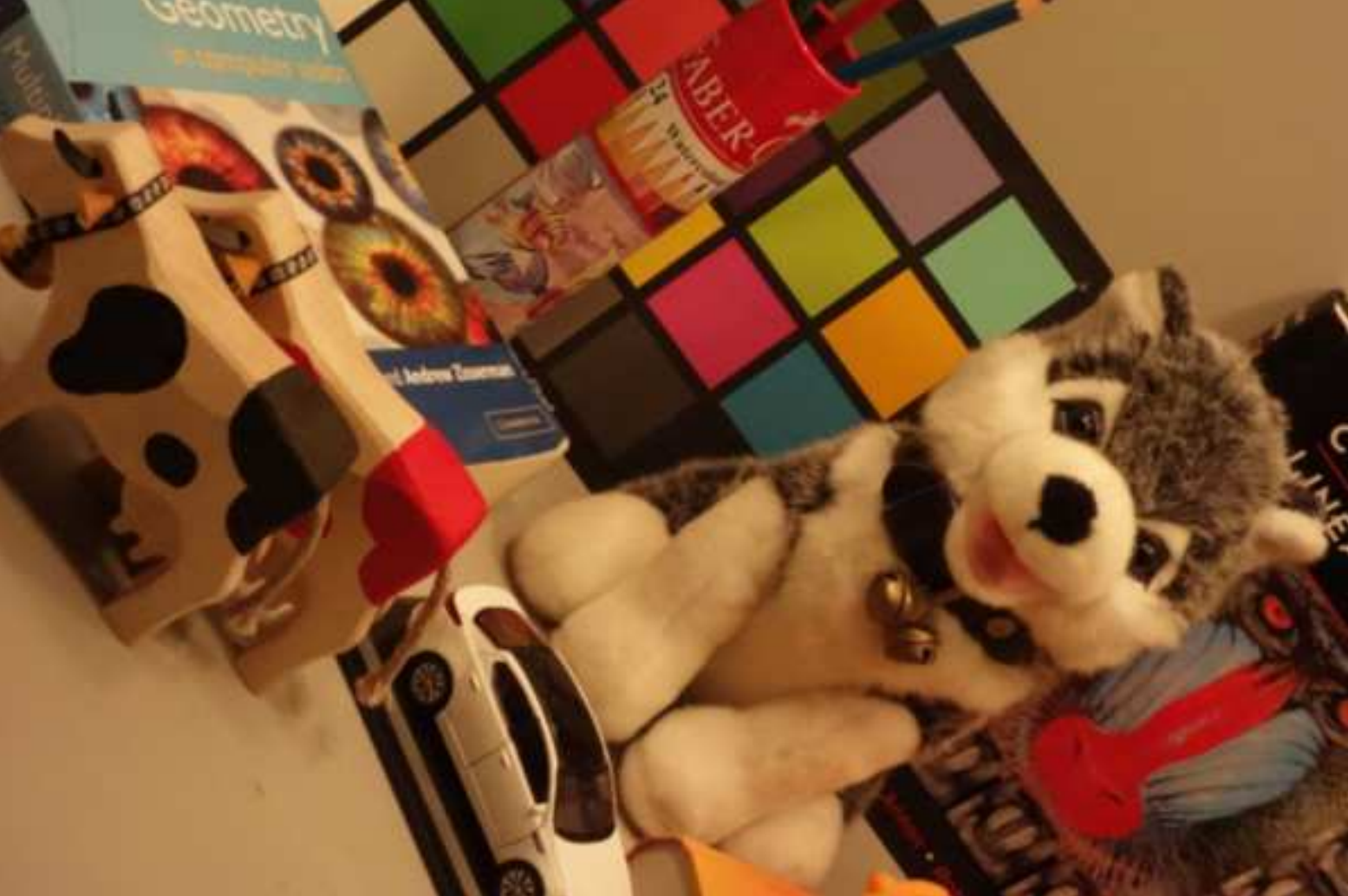}}\hfill
	\subfigure[(c) Superpixel 1]
	{\includegraphics[width=0.25\linewidth]{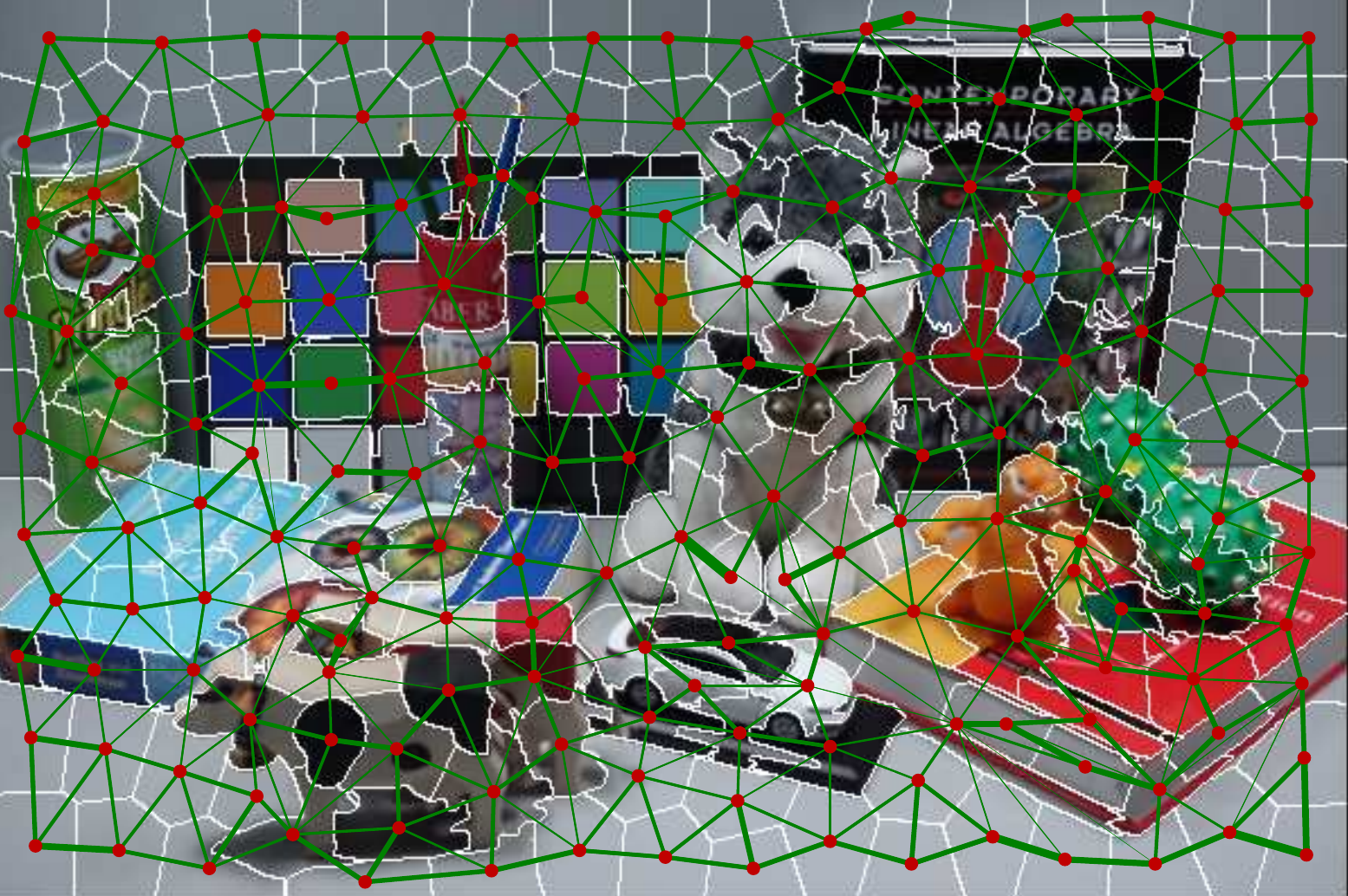}}\hfill
	\subfigure[(d) Superpixel 2]
	{\includegraphics[width=0.25\linewidth]{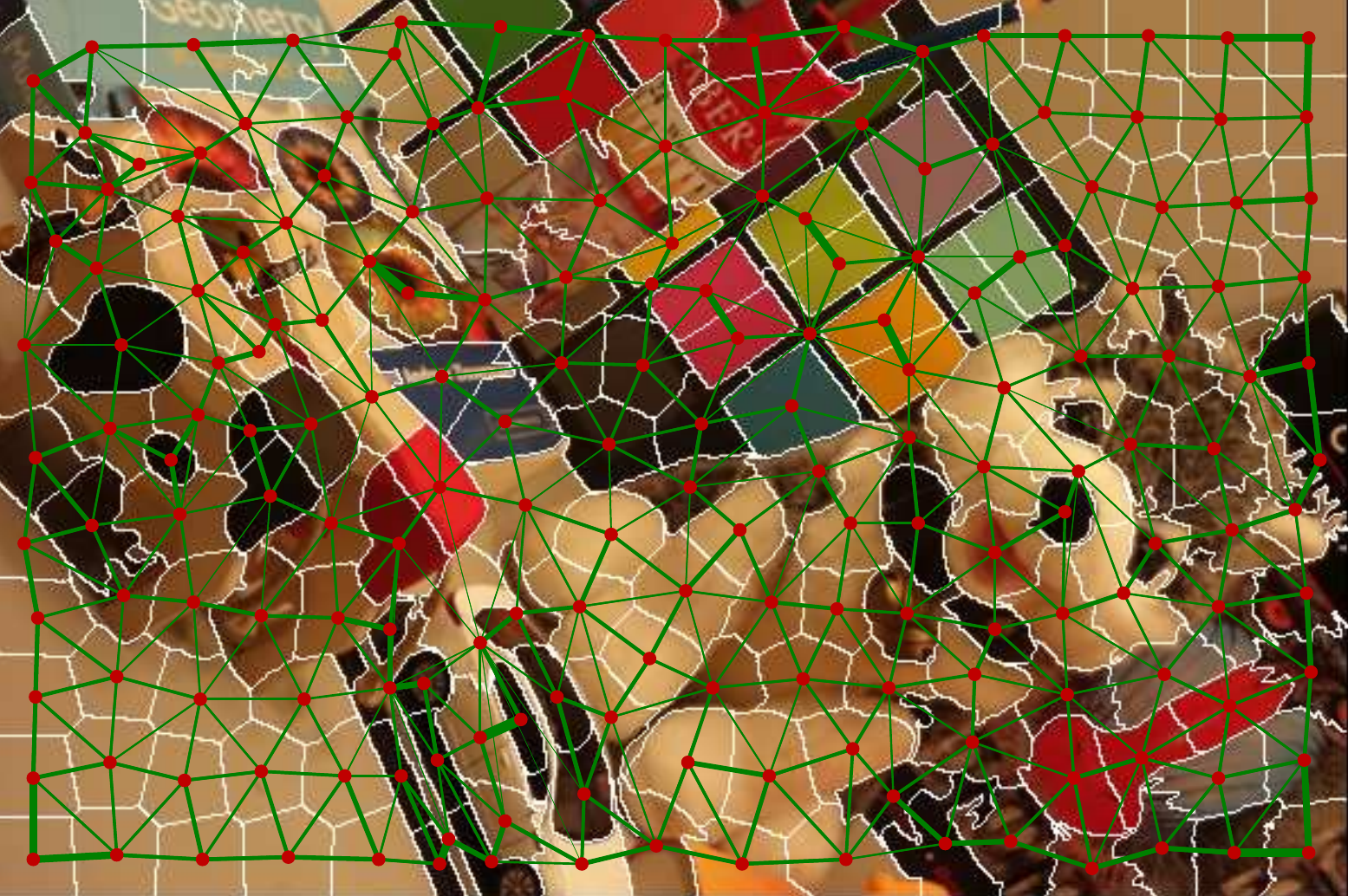}}\hfill
	\vspace{-8pt}
	\subfigure[(e) $\mathbf{G}^{*,\rho}_1$]
	{\includegraphics[width=0.25\linewidth]{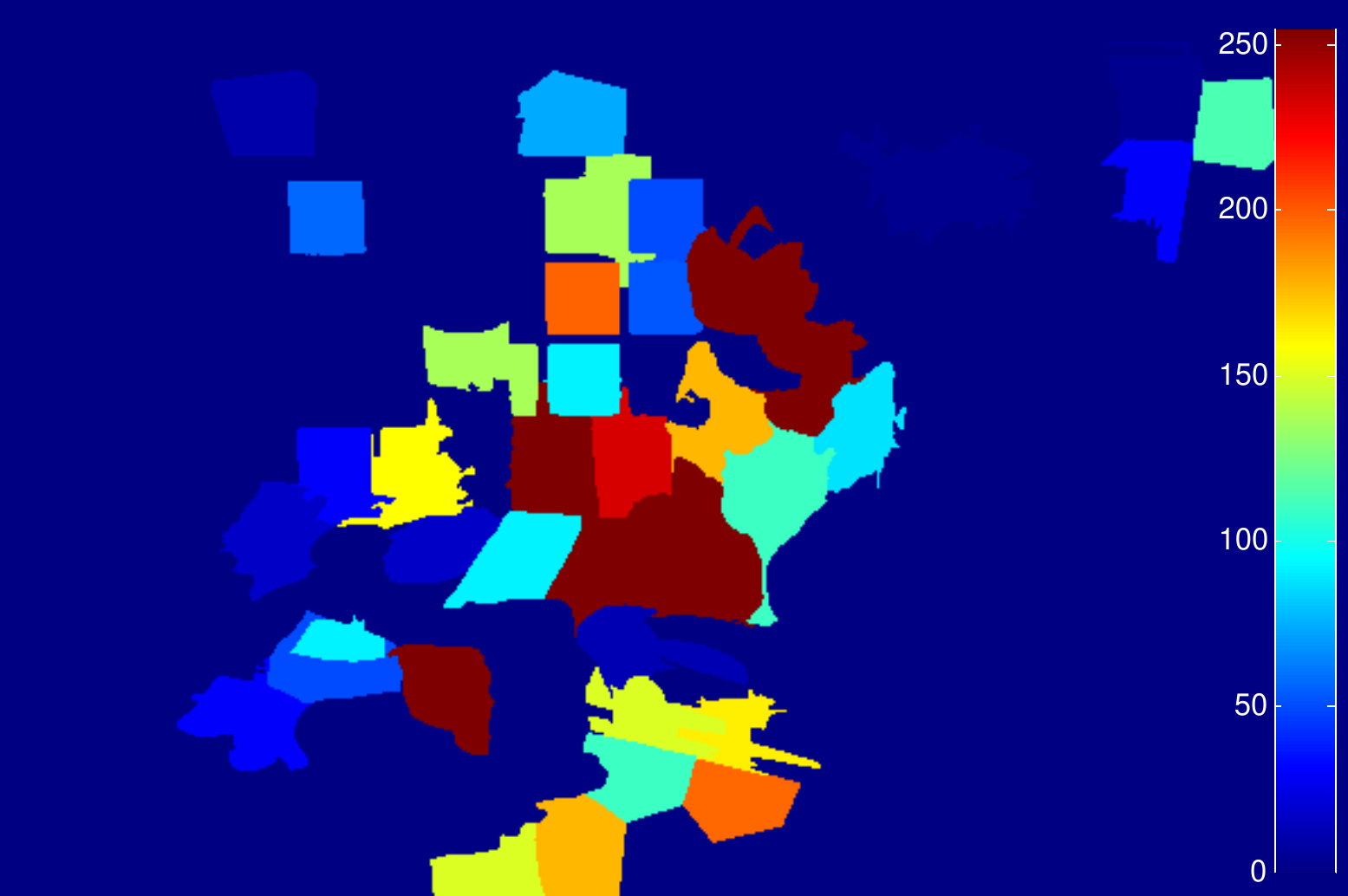}}\hfill
	\subfigure[(f) $\mathbf{G}^{*,\rho}_2$]
	{\includegraphics[width=0.25\linewidth]{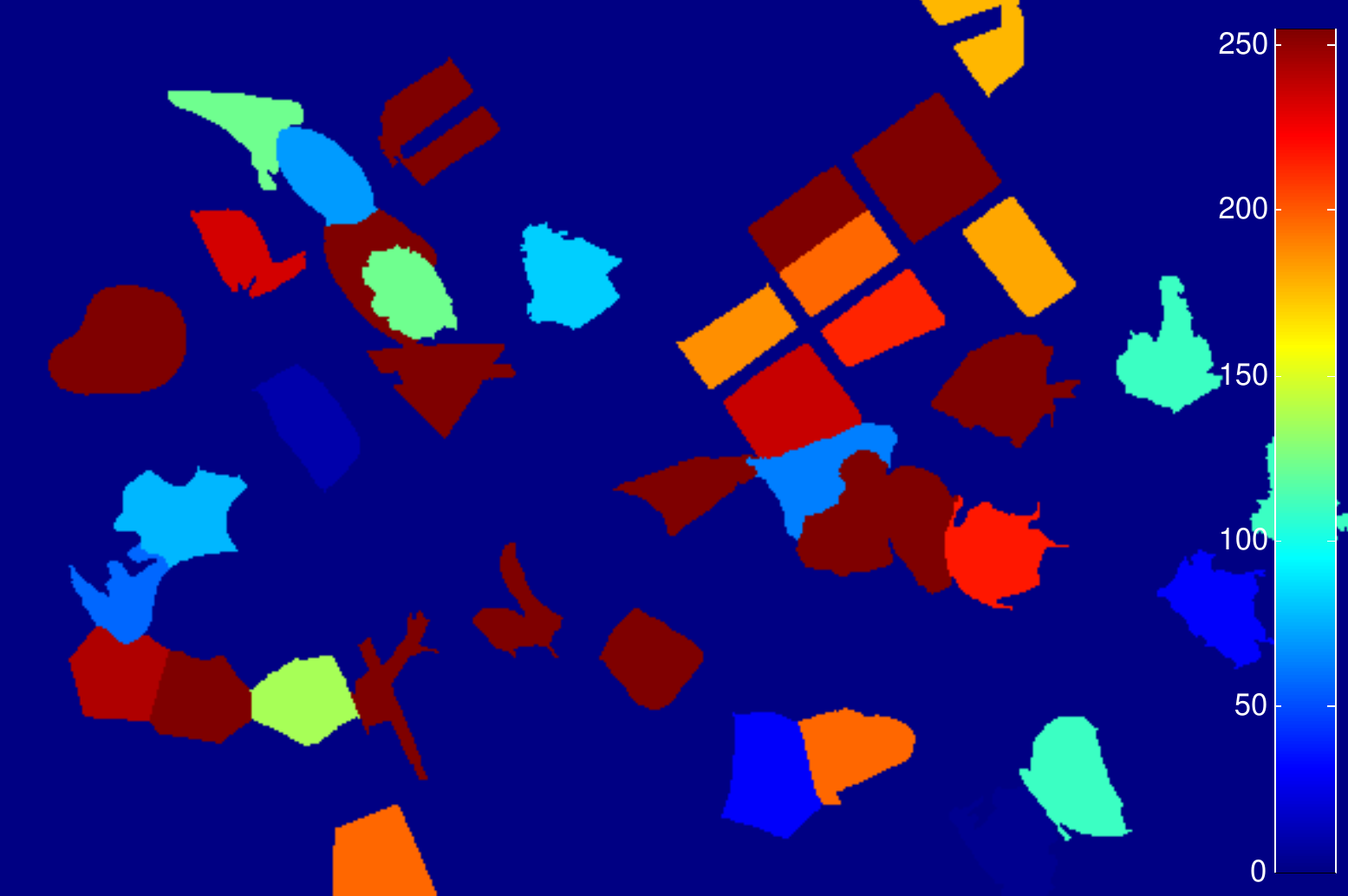}}\hfill
	\subfigure[(g) $\mathbf{G}^{*,\theta}_1$]
	{\includegraphics[width=0.25\linewidth]{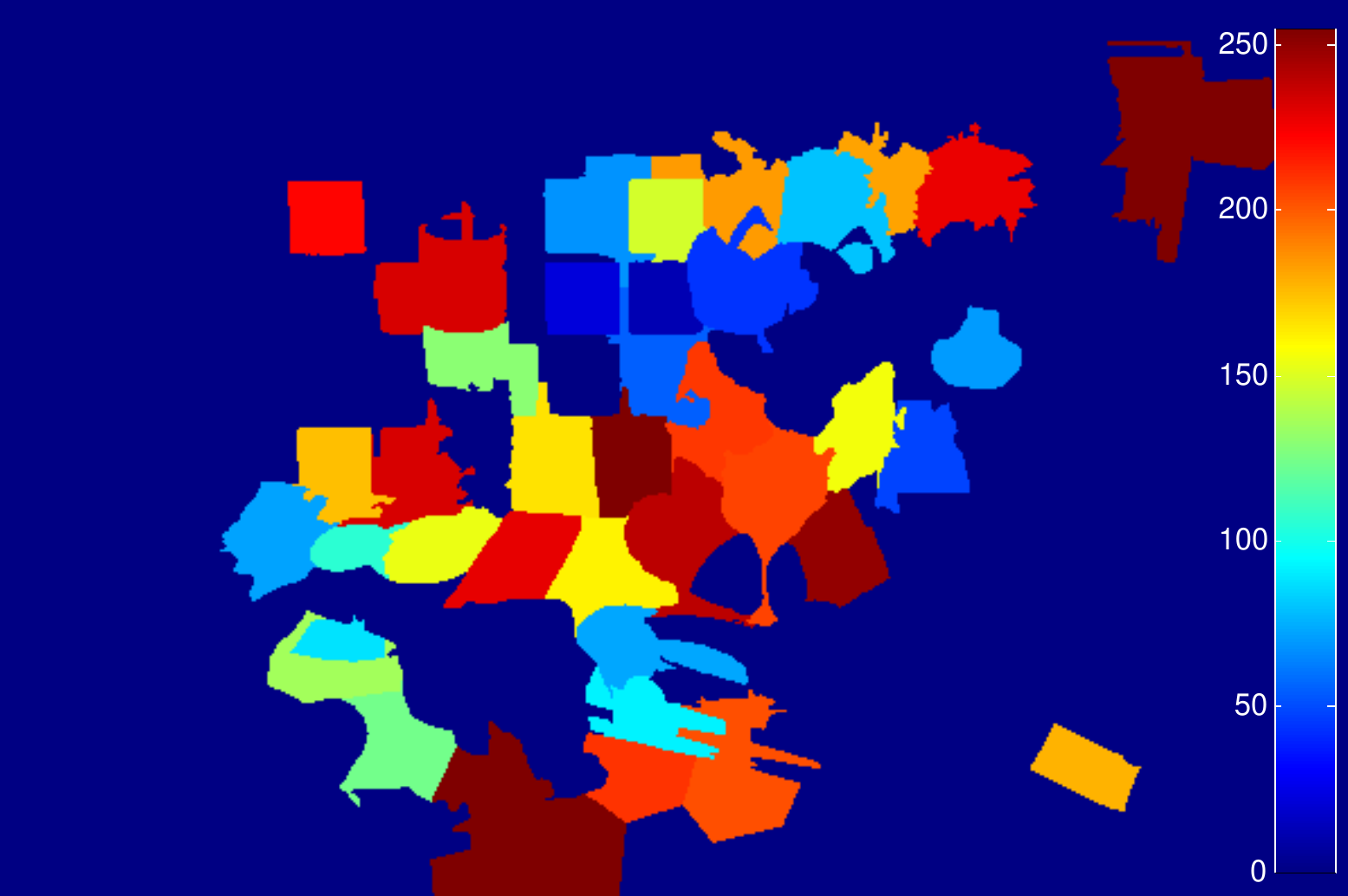}}\hfill
	\subfigure[(h) $\mathbf{G}^{*,\theta}_2$]
	{\includegraphics[width=0.25\linewidth]{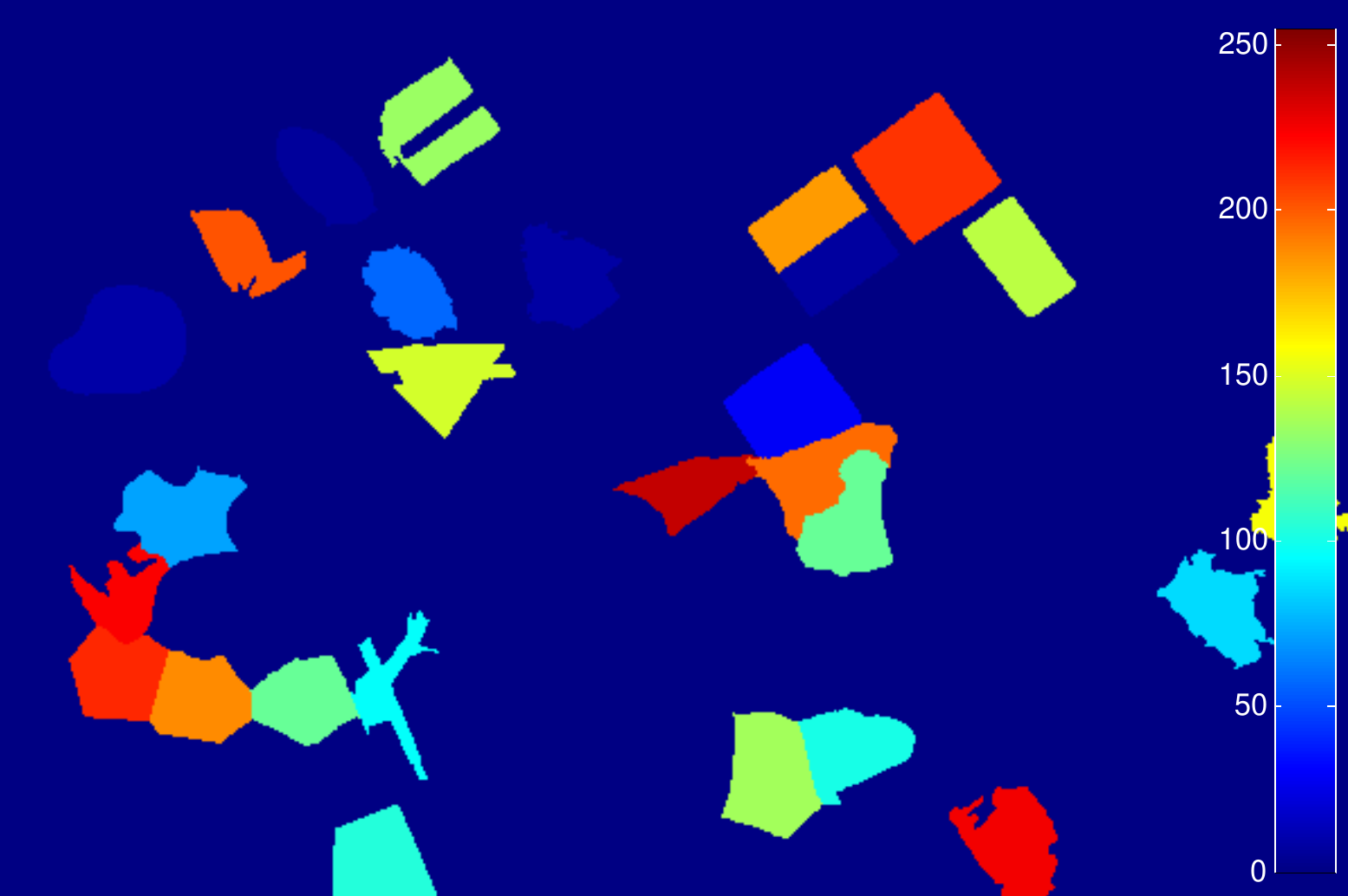}}\hfill
	\vspace{-8pt}
	\subfigure[(i) $\mathbf{G}^{\rho}_1$]
	{\includegraphics[width=0.25\linewidth]{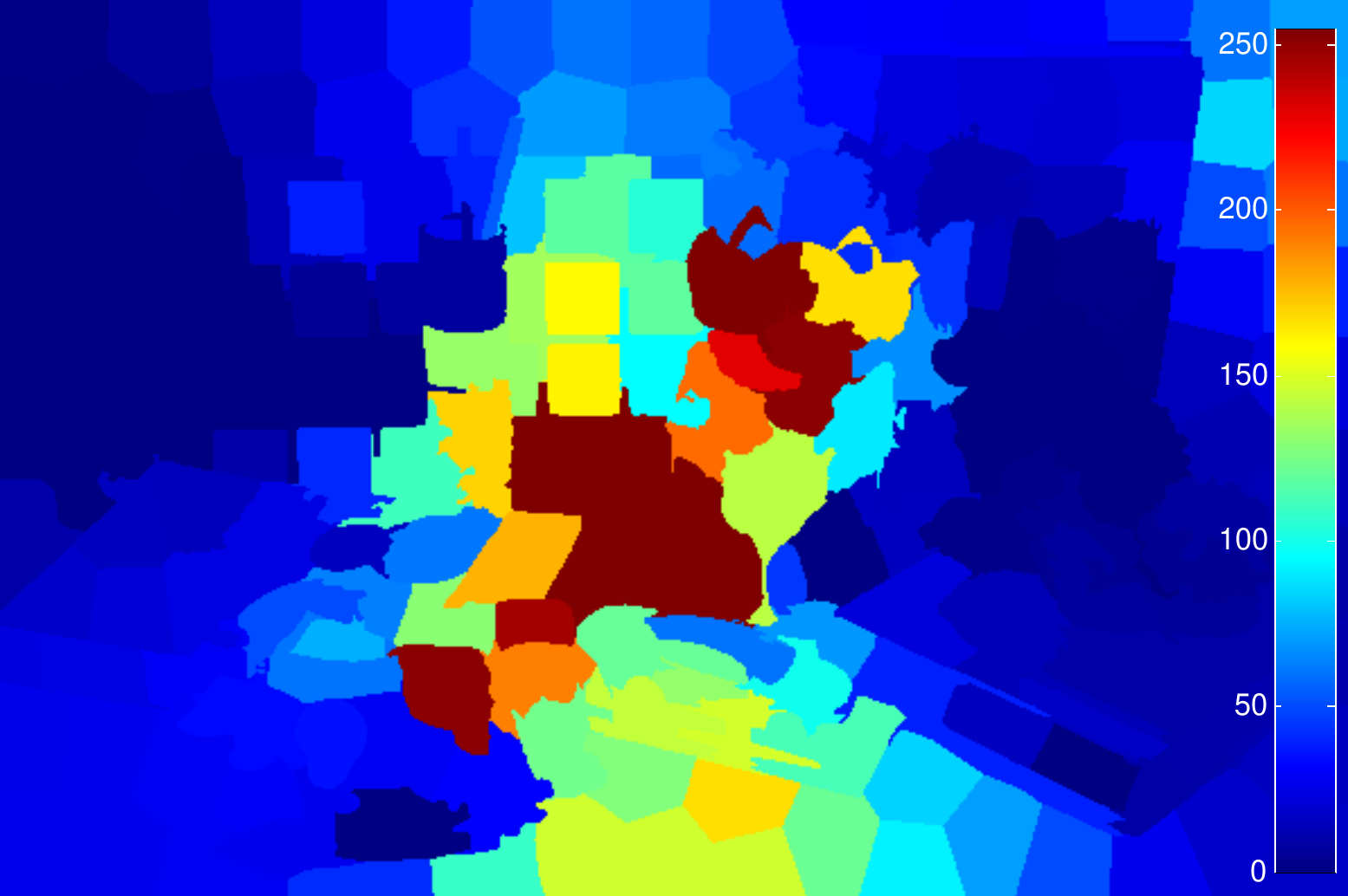}}\hfill
	\subfigure[(j) $\mathbf{G}^{\rho}_2$]
	{\includegraphics[width=0.25\linewidth]{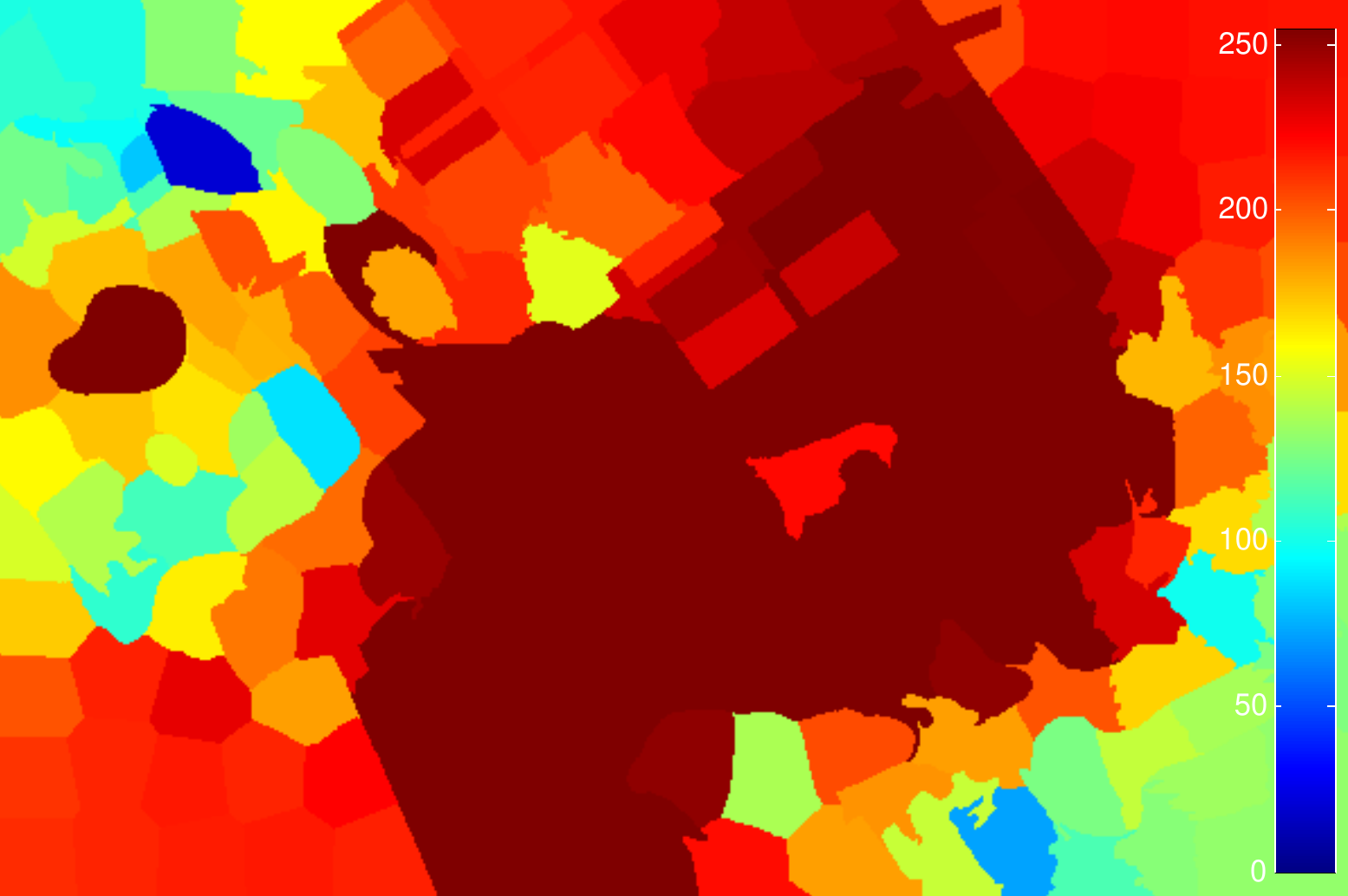}}\hfill
	\subfigure[(k) $\mathbf{G}^{\theta}_1$]
	{\includegraphics[width=0.25\linewidth]{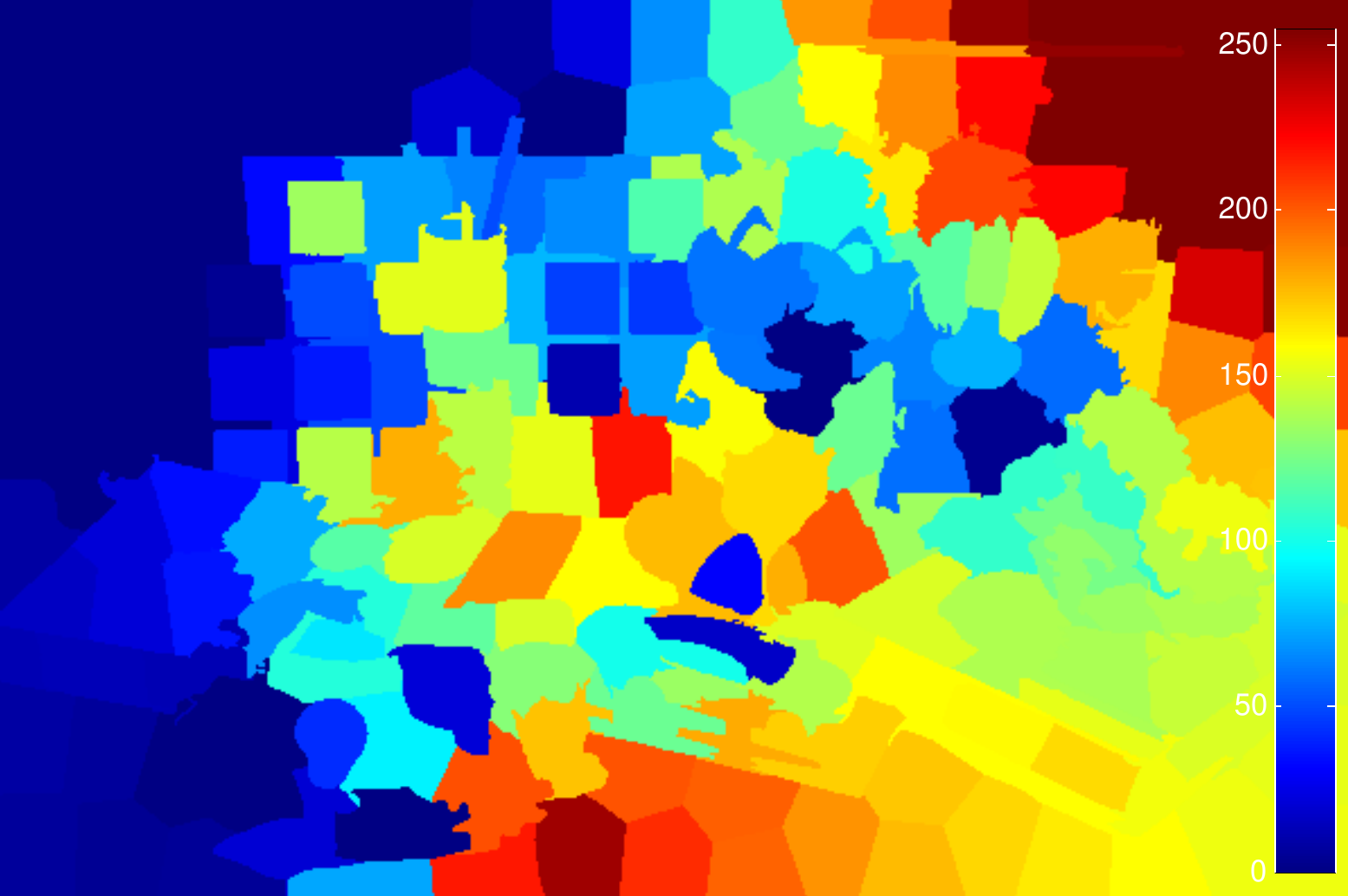}}\hfill
	\subfigure[(l) $\mathbf{G}^{\theta}_2$]
	{\includegraphics[width=0.25\linewidth]{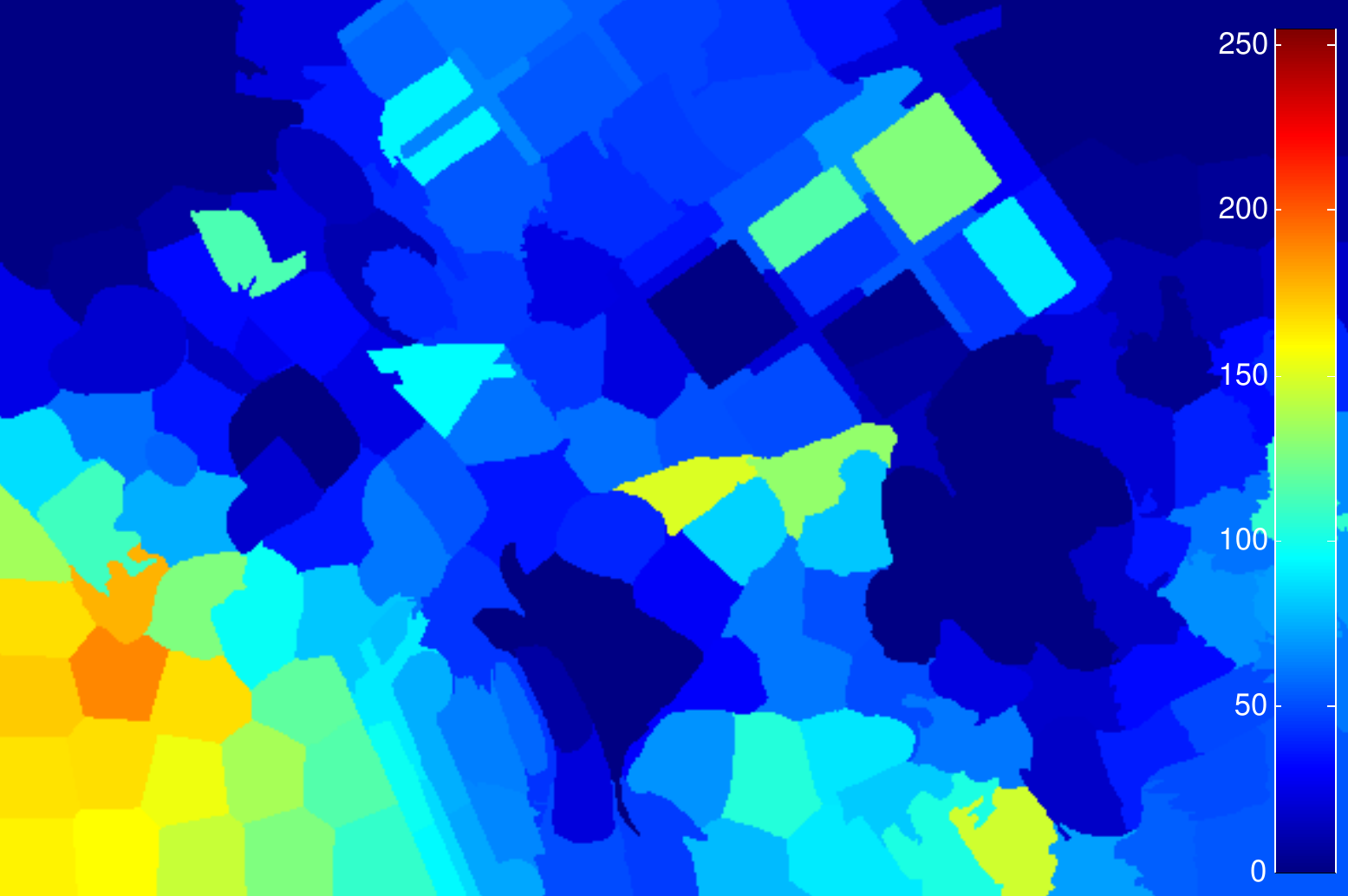}}\hfill
	\vspace{-10pt}
	\caption{Examples of a superpixel graph-based propagation.
		With each superpixel graph in (c), (d) for input images in (a), (b), sparse geometric fields (scale $\mathbf{G}^{*,\rho}$, rotation $\mathbf{G}^{*,\theta}$) in (e)-(h) are propagated
		into dense geometric fields (scale $\mathbf{G}^{\rho}$, rotation $\mathbf{G}^{\theta}$) in (i)-(l).}\label{img:14}\vspace{-10pt}
\end{figure}
\subsubsection{Superpixel feature affinity}\label{sec:521}
Our approach employs a superpixel feature
composed of an appearance and a spatial feature. 
First, appearance feature $\upsilon _m^c$
is defined as the average and standard deviation for intensities of pixels within
superpixels. 
In experiments, we used RGB, Lab, and YCbCr space for a color image, thus $\upsilon _m^c \in \mathbb{R}^{18}$.
For an NIR image, appearance feature is
defined on 1-channel intensity domain such that $\upsilon _m^c \in
\mathbb{R}^2$. 
Note that directly constructing an affinity matrix with intensity values may lead to inaccurate results due to intensity variations.
However, the effect on such variations can be greatly reduced, since the
appearance feature is defined as an aggregated form within a superpixel and the affinity value is measured within the same image domain. 
Second, spatial feature $\upsilon _m^p \in \mathbb{R}^2$ is defined as a spatial
centroid coordinate within superpixels. Based on these superpixel
features, a superpixel feature affinity
${\omega^{\mathrm{sp}}_{mn}}$ between two adjacent superpixel $m$ and $n
\in \mathcal{N}_m$ is computed as
\begin{equation}
{\omega^{\mathrm{sp}}_{mn}} = \exp ( - {\left\| {\upsilon _m^c - \upsilon _n^c} \right\|^2}/{\lambda _c} - {\left\| {\upsilon _m^p - \upsilon _n^p} \right\|^2}/{\lambda _p}) ,
\end{equation}
where ${\lambda _c}$ and ${\lambda _p}$ denote coefficients for controlling the spatial coherence of neighboring superpixels. \vspace{-5pt}
\subsubsection{Solver}\label{sec:522}
The minimum of the energy function \equref{equ:spenergy} can be obtained with the following linear system
\begin{equation}
(\mathbf{P}+\mu\mathbf{U}-\mu\mathbf{W})\mathbf{G} = \mathbf{P}\mathbf{G}^{\mathbf{*}},
\end{equation}
where $\mathbf{P}_{mm} = \mathrm{diag}[ p^{\mathrm{sp}}_1,...,p^{\mathrm{sp}}_{N_m} ]$,
$\mathbf{U}_{mm} = \mathrm{diag}[ u^{\mathrm{sp}}_1,...,u^{\mathrm{sp}}_{N_m} ]$ where $u^{\mathrm{sp}}_m = \sum\nolimits_{n \in {\mathcal{N}_m}}{\omega^{\mathrm{sp}}_{mn}}$,
and ${\mathbf{W}} = [\omega^{\mathrm{sp}}_{mn}]_{m,n=1,...,N_m}$.

This linear system with a Laplacian matrix can be easily solved
with conventional linear solvers \cite{Lang12}. \figref{img:14}
shows examples of our superpixel graph-based propagation.
\vspace{-5pt}
\subsection{Efficient Dense Descriptor on Superpixels}\label{sec:54}
\begin{figure}[!t]
	\centering
	\renewcommand{\thesubfigure}{}
	\subfigure[(a) Superpixel extended subimage]
	{\includegraphics[width=0.5\linewidth]{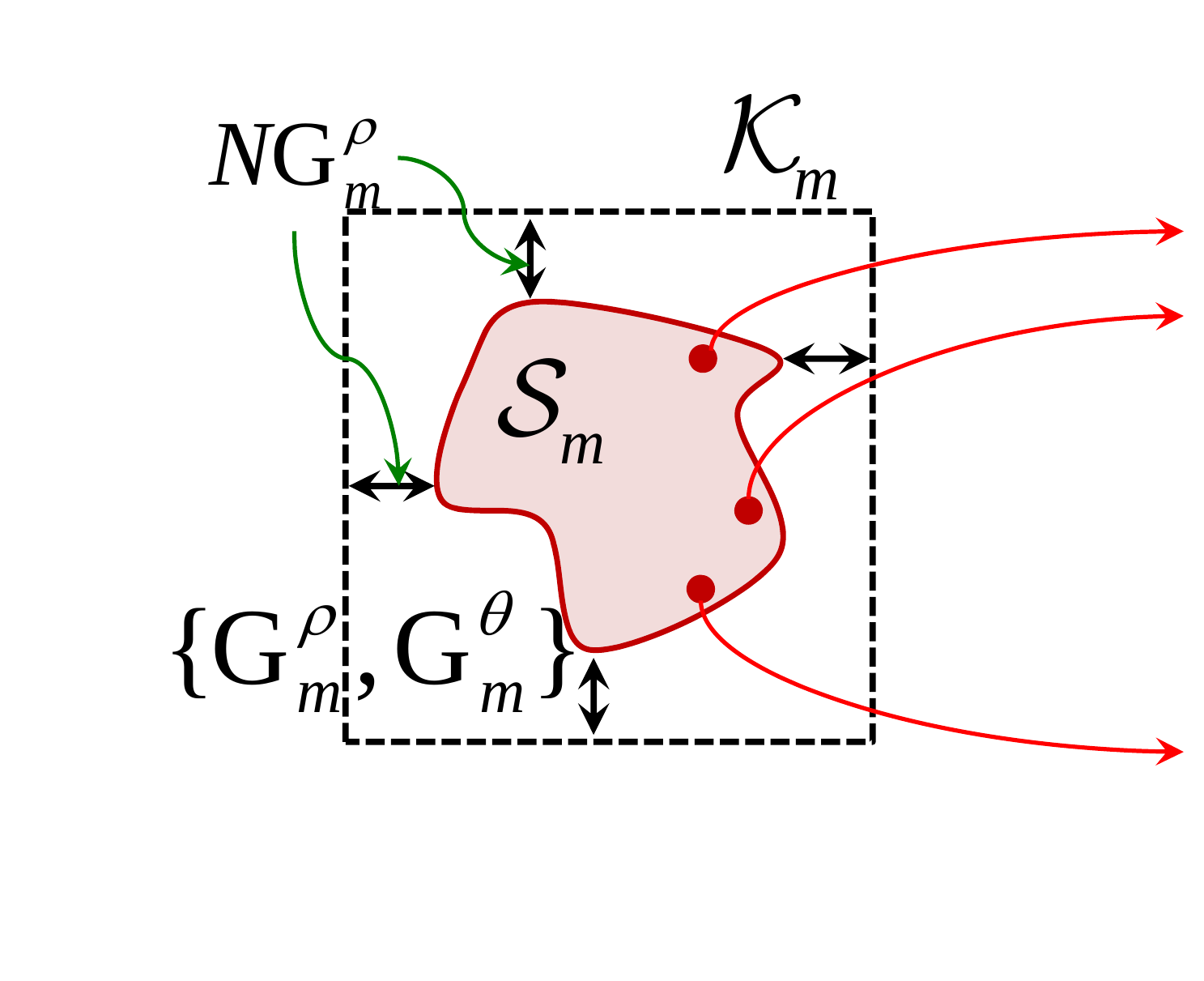}}\hfill
	\subfigure[(b) Sampling pattern $\Lambda_{m}^{\mathrm{gi-dasc}}$]
	{\includegraphics[width=0.5\linewidth]{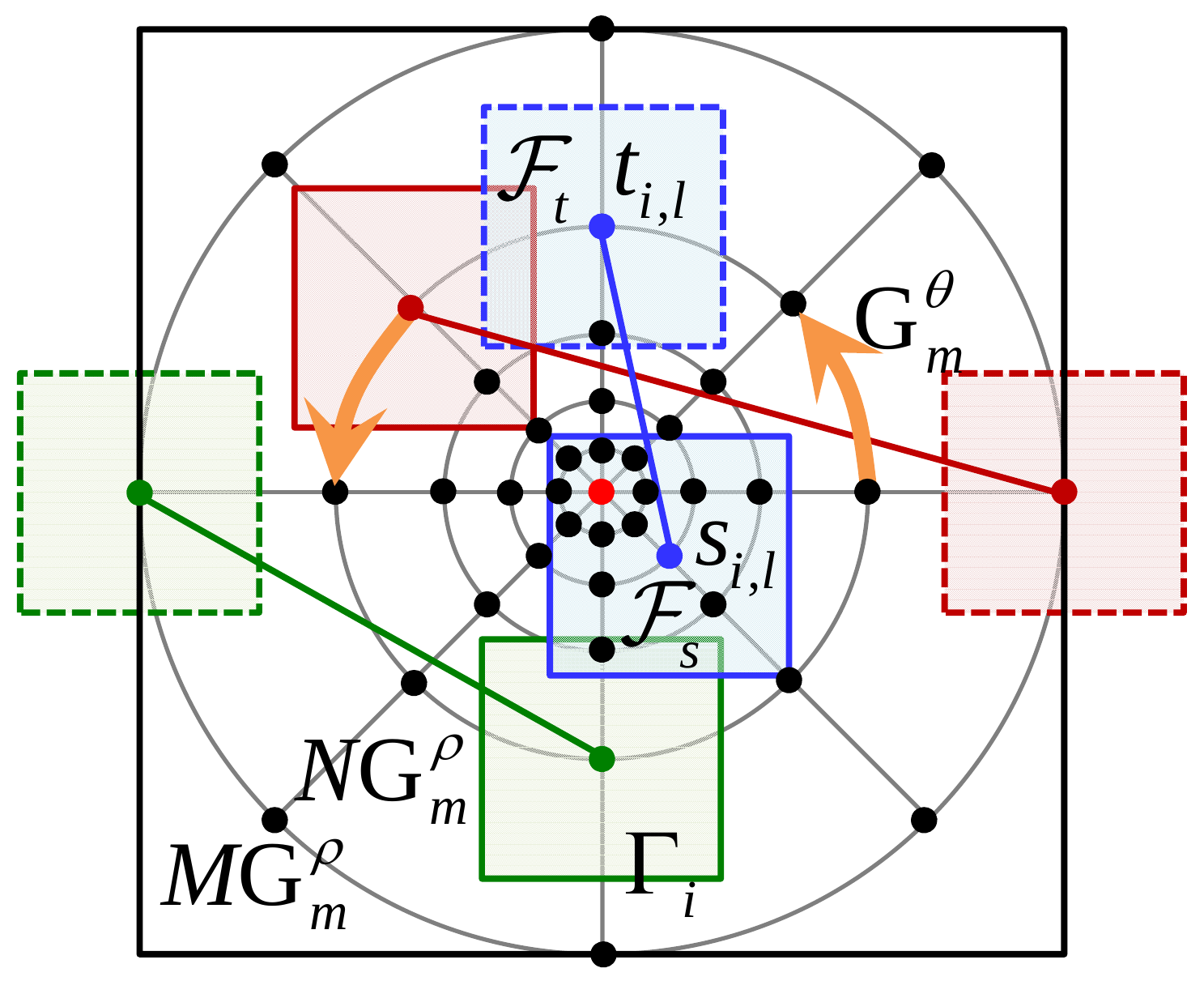}}\hfill
	\vspace{-10pt}
	\caption{Sampling pattern transformation in the GI-DASC descriptor.
		The sampling patterns $(s_{i,l},t_{i,l}) \in \Lambda_{i}^{\mathrm{dasc}}$ is transformed as $(s_{m,l},t_{m,l}) \in \Lambda_{m}^{\mathrm{gi-dasc}}$
		with $\mathrm{G}^{\rho}_m$ and $\mathrm{G}^{\theta}_m$ on superpixel $\mathcal{S}_m$, which is applied equally for all $i \in \mathcal{S}_m$.
		It provides the geometric robustness on each superpixel.}\label{img:15}\vspace{-10pt}
\end{figure}

The sampling patterns are transformed with corresponding
geometric fields $\mathbf{G}^{\rho}$ and $\mathbf{G}^{\theta}$ as
shown in \figref{img:15}. Specifically, for the $m$-th superpixel
$\mathcal{S}_m$, the sampling pattern $(s_{m,l},t_{m,l}) \in
\Lambda^{\mathrm{gi-dasc}}_m$ is transformed from $(s_{l}, t_{l})
\in \Lambda^{\mathrm{dasc}}$ with a scale factor
$\mathrm{G}^{\rho}_m$ and a rotation factor
$\mathrm{G}^{\theta}_m$,
\begin{equation}
s_{m,l} = \mathbf{S}_m \mathbf{R}_m s_{l},
\end{equation}
where the scale matrix $\mathbf{S}_m = \mathrm{diag}[ \mathrm{G}^{\rho}_m ]$ and the rotation matrix
$\mathbf{R}_m$ is defined with rotation $\mathrm{G}^{\theta}_m$. In
a similar way, $t_{m,l}$ is also estimated from $t_{l}$.
Finally, $\Lambda^{\mathrm{gi-dasc}}_m$ is
estimated. Furthermore, the patch size $N$ is enlarged as
$N\mathrm{G}^{\rho}_m$.
\begin{table}[!t]
	\begin{center}
		\begin{tabularx}{\linewidth}{p{1mm} p{0.02mm}| p{80mm}}
			\hlinewd{0.8pt}
			\multicolumn{3}{ p{85mm} }{{\bf Algorithm 3}: Geometric-Invariant DASC (GI-DASC)} \\
			\hlinewd{0.8pt}
			\multicolumn{3}{ p{85mm} }{{\bf Input} : image ${f_i}$, feature detection sampling patterns $\Lambda_{i}^{\mathrm{det}}$,
				$L^{\mathrm{dasc}}$ sampling patterns $(s_{i,l},t_{i,l}) \in \Lambda_{i}^{\mathrm{dasc}}$.} \\
			\multicolumn{3}{ p{85mm} }{{\bf Output} : the GI-DASC descriptor volume ${\mathcal{D}^{\mathrm{gi-dasc}}_{i}}$.}\\
			\\
			$\mathbf{1:}$&\multicolumn{2}{ p{80mm} }{Extract feature points $i\in \mathcal{I'}$ with scale $\rho_i$ and rotation $\theta_i$ using Algorithm 2.} \\
			$\mathbf{2:}$&\multicolumn{2}{ p{80mm} }{Decompose the image ${f_i}$ into superpixels $\mathcal{S}$.} \\
			$\mathbf{3:}$&\multicolumn{2}{ p{80mm} }{Compute a surface fitting for geometric field $\mathbf{G}^{*,\rho}_m$ and $\mathbf{G}^{*,\theta}_m$ on superpixels $\mathcal{S}_m$.} \\
			$\mathbf{4:}$&\multicolumn{2}{ p{80mm} }{Compute a Laplacian matrix ${\mathbf{P} + \mu\mathbf{U} - \mu\mathbf{}W}$ with confidences ${p^{\mathrm{sp}}_m}$ and weights ${\omega^{\mathrm{sp}}_{mn}}$.} \\
			$\mathbf{5:}$&\multicolumn{2}{ p{80mm} }{Compute dense geometric fields $\mathbf{G}^{\rho}_m$ and $\mathbf{G}^{\theta}_m$.} \\
			~&\multicolumn{2}{ b{80mm} }{{\bf for } $m = 1:N_m$ {\bf do }}\\
			$\mathbf{6:}$&~&  Transform the sampling pattern $\Lambda_{i}^{\mathrm{dasc}}$ into $\Lambda_{m}^{\mathrm{gi-dasc}}$. \\
			$\mathbf{7:}$&~&  Compute the GI-DASC descriptor ${d^{\mathrm{gi-dasc}}_{i,l}} = \mathcal{C}(s_{i,l},t_{i,l})$ for $i\in \mathcal{S}_m$ and $(s_{m,l},t_{m,l})\in \Lambda^{\mathrm{gi-dasc}}_m$  using Algorithm 1.\\
			~&\multicolumn{2}{ b{80mm} }{{\bf end for }}\\
			\hlinewd{0.8pt}
		\end{tabularx}
	\end{center}\label{alg:3}\vspace{-10pt}
\end{table}

The $m$-th superpixel extended subimage $\mathcal{K}_m$ in \figref{img:15}(a) is filtered
by a Gaussian filtering with the sigma
$\{(\mathrm{G}^{\rho}_m)^2-0.25\}^{-1/2}$ similar to scale-space
theory used in the SIFT \cite{Lowe04}. Then, our GI-DASC descriptor
$\mathcal{D}^{\mathrm{gi-dasc}}_{i} = { \bigcup
_{l}}d^{\mathrm{gi-dasc}}_{i,l}$ for $l=1,...,L^{\mathrm{gi-dasc}}$
($=L^{\mathrm{dasc}}$) is encoded with a set of patch similarity
between two patches from a transformed sampling pattern
$\Lambda^{\mathrm{gi-dasc}}_m$ on each superpixel $\mathcal{S}_m$
such that
\begin{equation}\label{equ:dasc}
\begin{array}{l}
d^{\mathrm{gi-dasc}}_{i,l} = \mathcal{C}(s_{i,l},t_{i,l}),\quad (s_{i,l},t_{i,l})\in \Lambda^{\mathrm{gi-dasc}}_m, \\
\end{array}
\end{equation}
for $i \in {\mathcal{S}_m}$. Finally, the dense GI-DASC descriptor
is efficiently computed for all the superpixels $\mathcal{S}_m \in
\mathcal{S}$. Algorithm 3 summarizes how to compute the GI-DASC
descriptor. \vspace{-5pt}
\begin{figure}[t]
\centering
\renewcommand{\thesubfigure}{}
\subfigure[(a) Support window size]
{\includegraphics[width=0.5\linewidth]{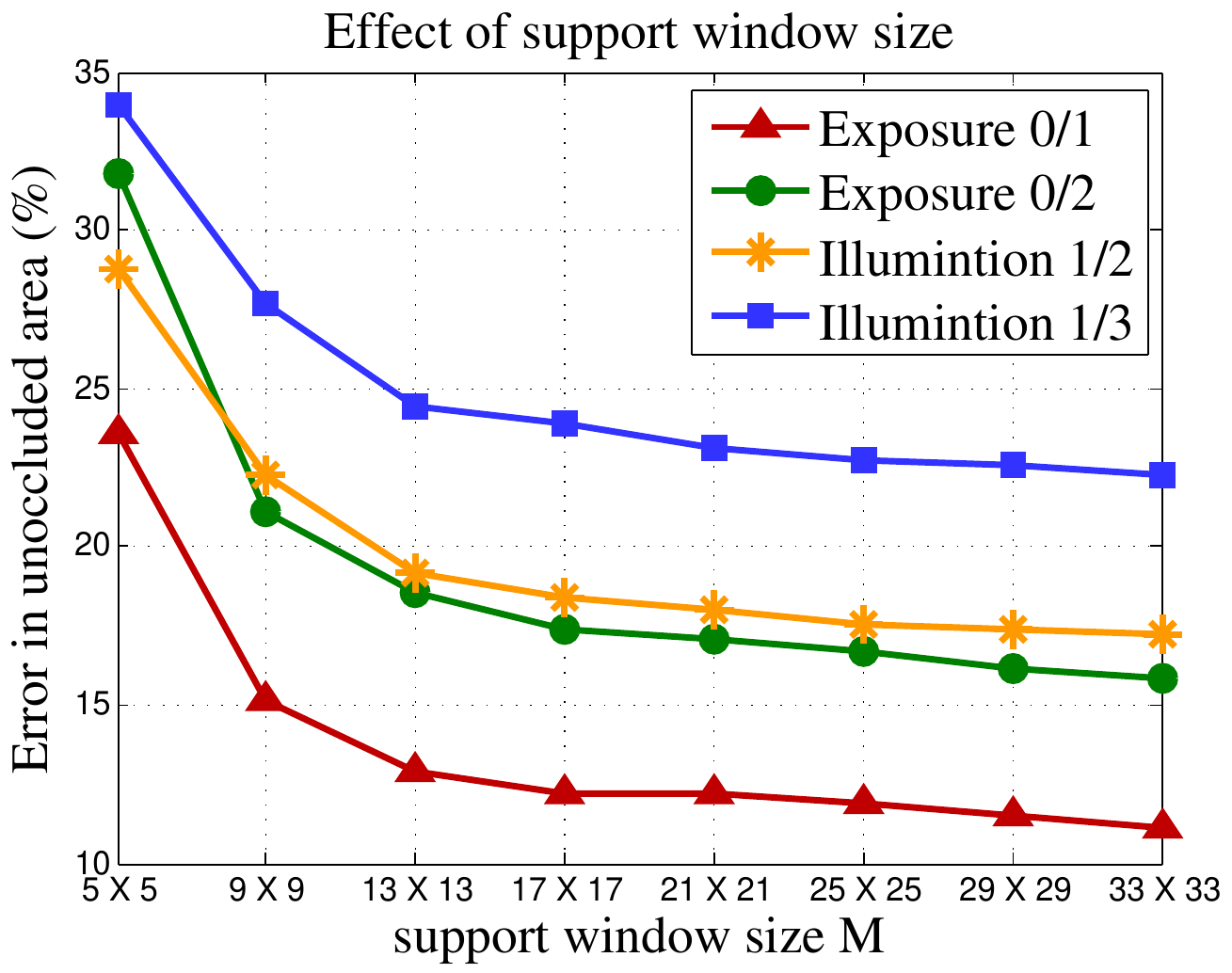}}\hfill
\subfigure[(b) Descriptor dimension]
{\includegraphics[width=0.5\linewidth]{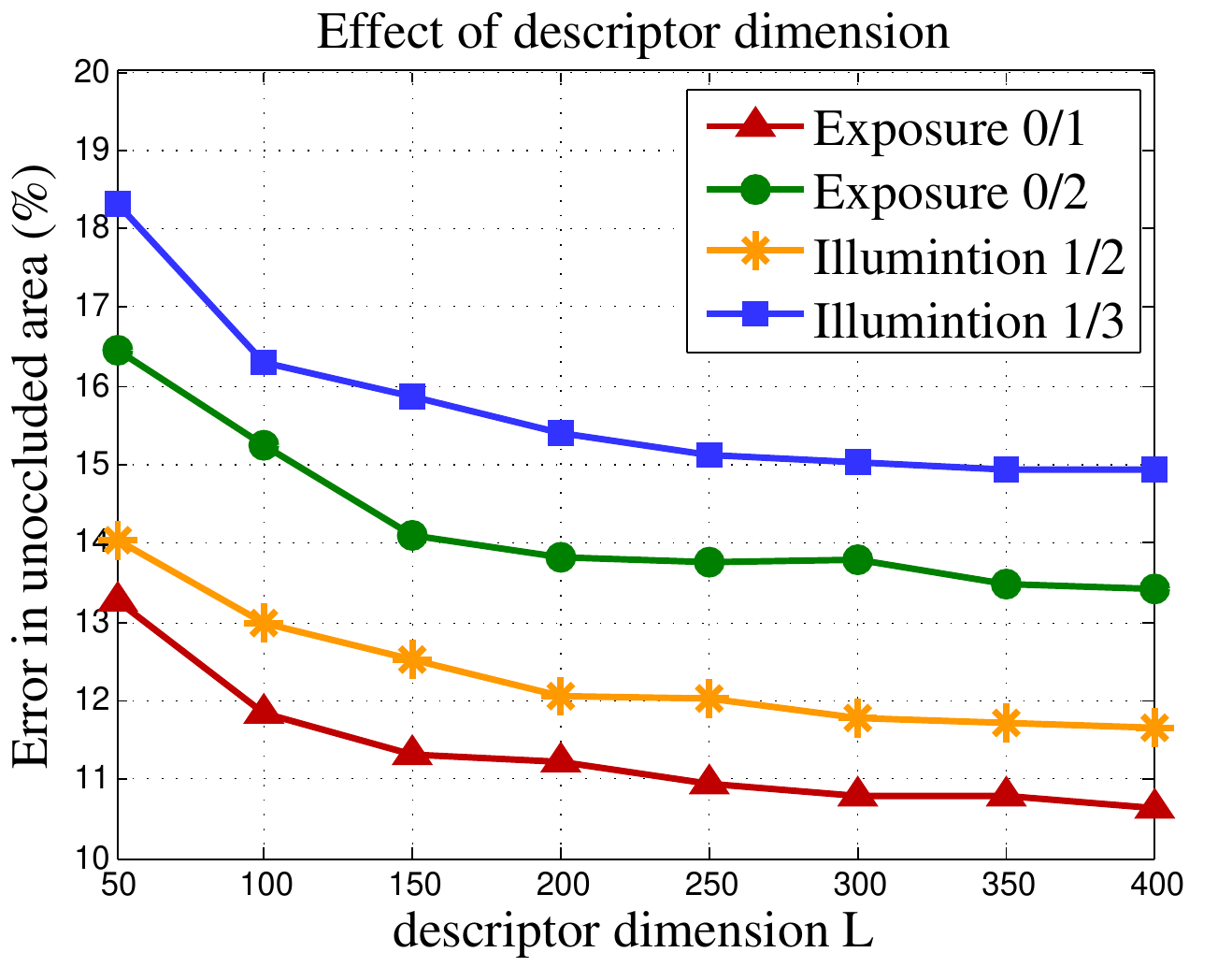}}\hfill
\subfigure[(c) Patch size]
{\includegraphics[width=0.5\linewidth]{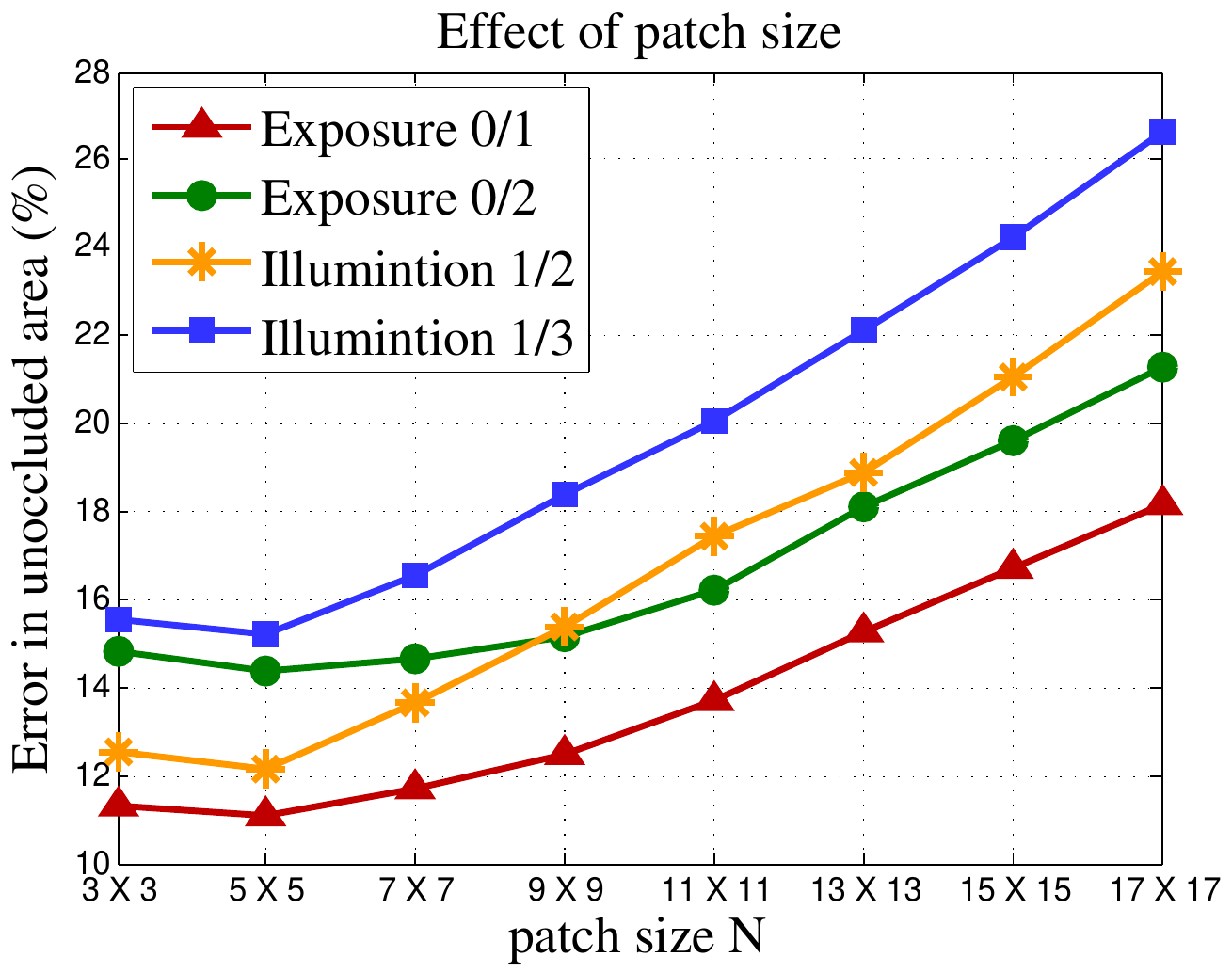}}\hfill
\subfigure[(d) Log-polar circular point]
{\includegraphics[width=0.5\linewidth]{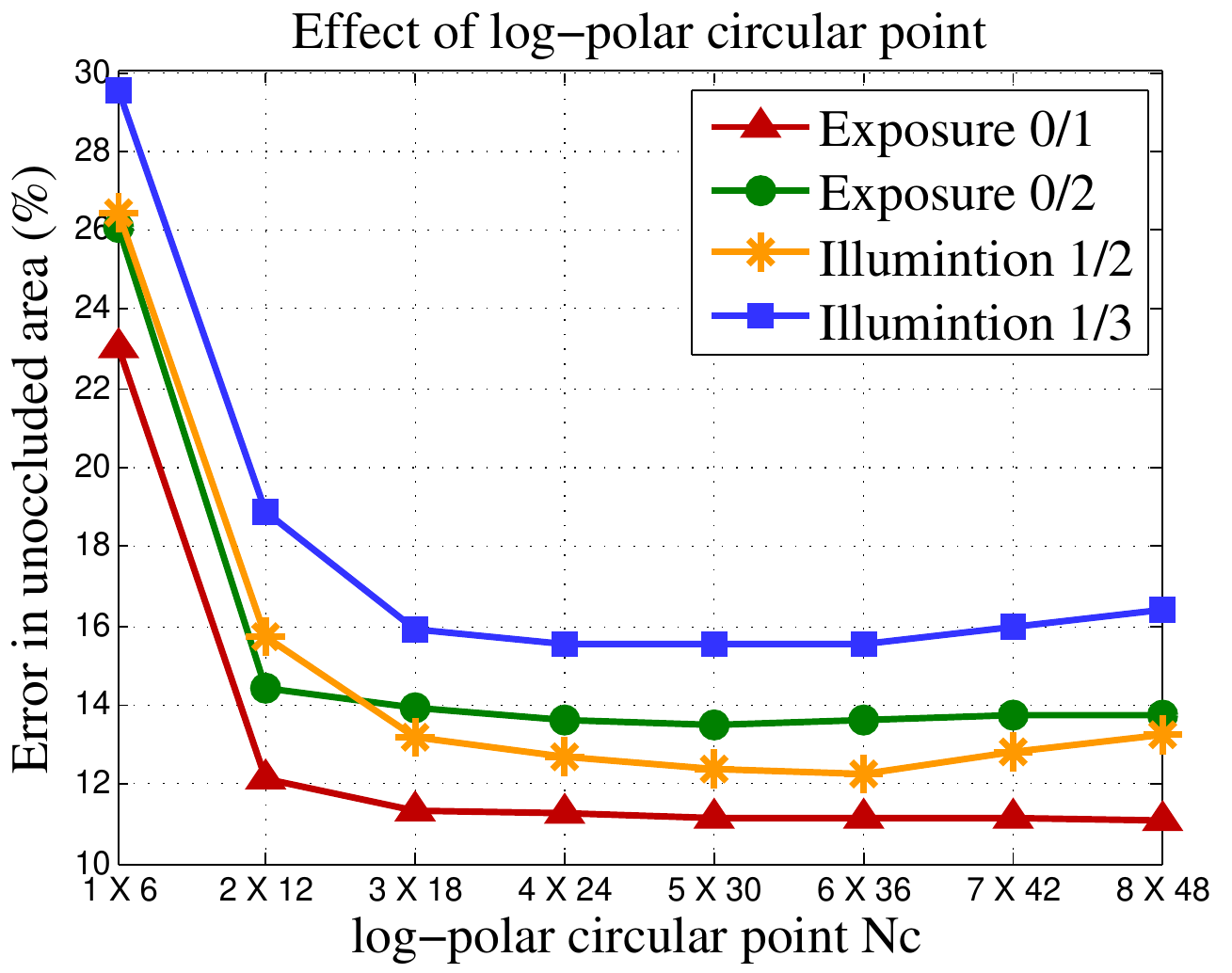}}\hfill
\vspace{-10pt}
\caption{Average bad-pixel error rate on Middlebury benchmark \cite{middlebury} of DASC+LRP descriptor with WTA optimization
as varying support window size $M$, descriptor dimension $L$, patch size $N$, and log-polar circular point $N_c$ ($\approx N_\rho \times N_\theta$).
In each experiment, all other parameters are fixed as initial values in \secref{sec:61}.}\label{img:16}\vspace{-10pt}
\end{figure}
\section{Experimental Results and Discussions}\label{sec:6}
\subsection{Experimental Environments}\label{sec:61}
In experiments, the DASC descriptor was implemented with the
following same parameter settings for all datasets: $\{
\sigma_c,\tau_c,N,M,L^{\mathrm{dasc}}\}  = \{ 0.5,0.03,5 \times 5,31
\times 31,128\}$ where $M$ is the support window size, and
$\{N_\rho,N_\theta\}  = \{4,36\}$ for candidate sampling patterns. We
set the smoothness parameter $\epsilon = 0.03^2$ in the GF
\cite{He13}. For the GI-DASC, the following parameters were used for
all datasets:
$\{N^{\mathrm{wmsd}}_\rho,N^{\mathrm{wmsd}}_\theta,N_k,o,{\lambda _c},{\lambda _p}\}  =
\{3,12,4,10,0.1,30\}$. The number of superpixels is set to about $500$.
We implemented the DASC and GI-DASC descriptor in C++ on Intel Core
i$7$-$3770$ CPU at $3.40$ GHz.

The DASC descriptor was evaluated with other state-of-the-art
descriptors, \emph{e.g.}, SIFT \cite{Lowe04}, DAISY \cite{Tola10},
BRIEF \cite{Calonder11}, and LSS \cite{Schechtman07}, and other
area-based approaches, \emph{e.g.}, ANCC \cite{Heo}, MI+SIFT\footnote{For a fair evaluation, we compared only the similarity measure in \cite{Heo13} without further techniques.}
\cite{Heo13}, and RSNCC \cite{Shen14}. We also compared the DASC
using a randomized pooling (DASC+RP) with the DASC using a learned
randomized pooling (DASC+LRP). Furthermore, the state-of-the-art
geometry robust methods such as SID \cite{Kokkinos08}, SegSID
\cite{Kokkinos08}, SegSF \cite{Trulls13}, GPM \cite{Barnes10},
DSP \cite{Kim13}, and SSF \cite{Qiu14} were also compared to the GI-DASC descriptor.
For learning the DASC, we built training sets
$\mathcal{P}$ from benchmark databases used in each experiment, and
these training sets were excluded from experiments. \vspace{-8pt}
\begin{figure}[!t]
	\centering
	\renewcommand{\thesubfigure}{}
	\subfigure[(a) Illumination variation]
	{\includegraphics[width=0.5\linewidth]{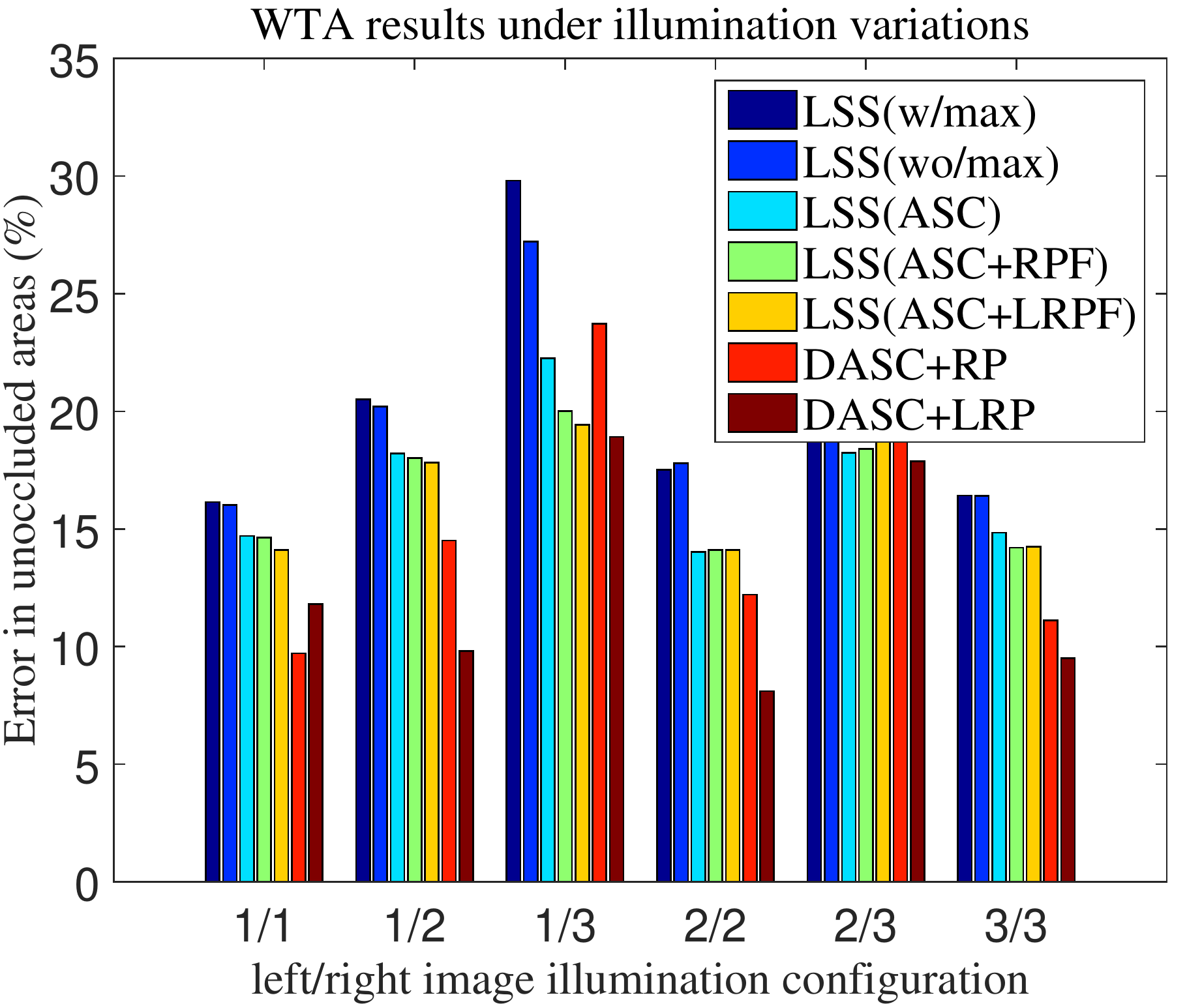}}\hfill
	\subfigure[(b) Exposure variation]
	{\includegraphics[width=0.5\linewidth]{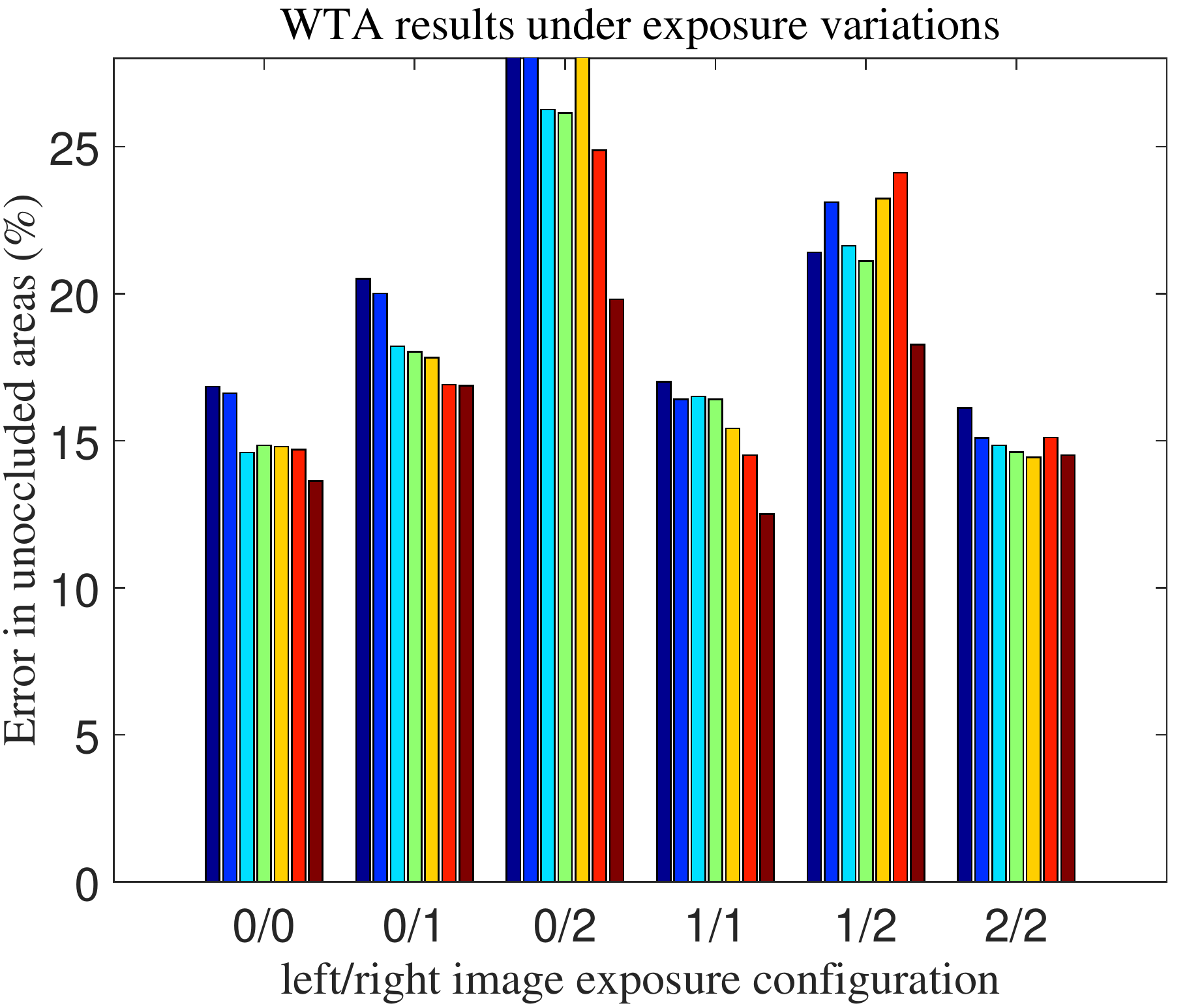}}\hfill
	\vspace{-10pt}
	\caption{Average bad-pixel error rate for original LSS \cite{Schechtman07}, LSS without max-pooling, LSS with ASC, LSS using randomized-pooling with fixed center pixel,
		and the DASC descriptor on Middlebury benchmark \cite{middlebury}.}\label{img:17}\vspace{-10pt}
\end{figure}
\begin{figure}[t]
\centering
\renewcommand{\thesubfigure}{}
\subfigure[(a) Illumination variation]
{\includegraphics[width=0.5\linewidth]{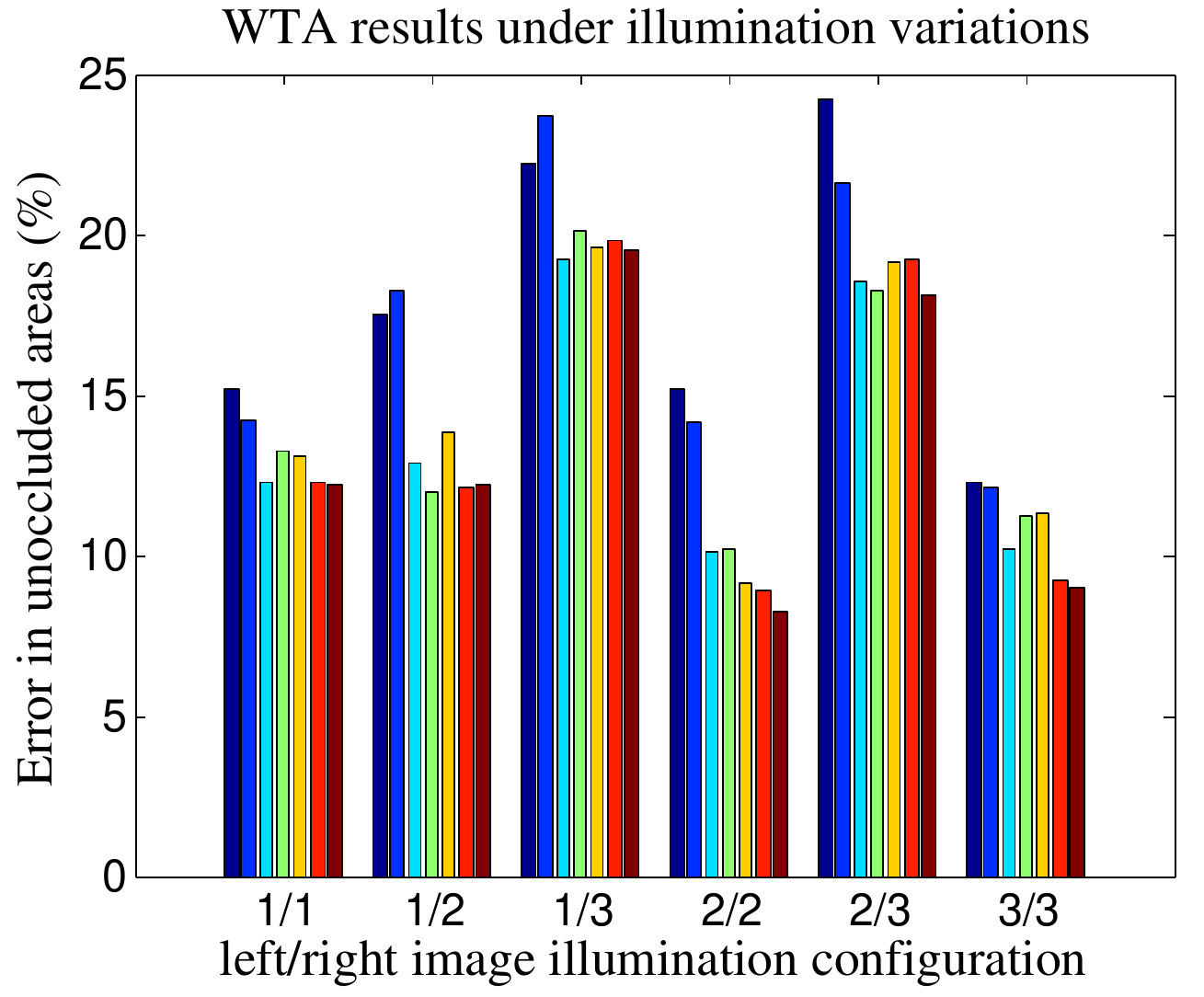}}\hfill
\subfigure[(b) Exposure variation]
{\includegraphics[width=0.5\linewidth]{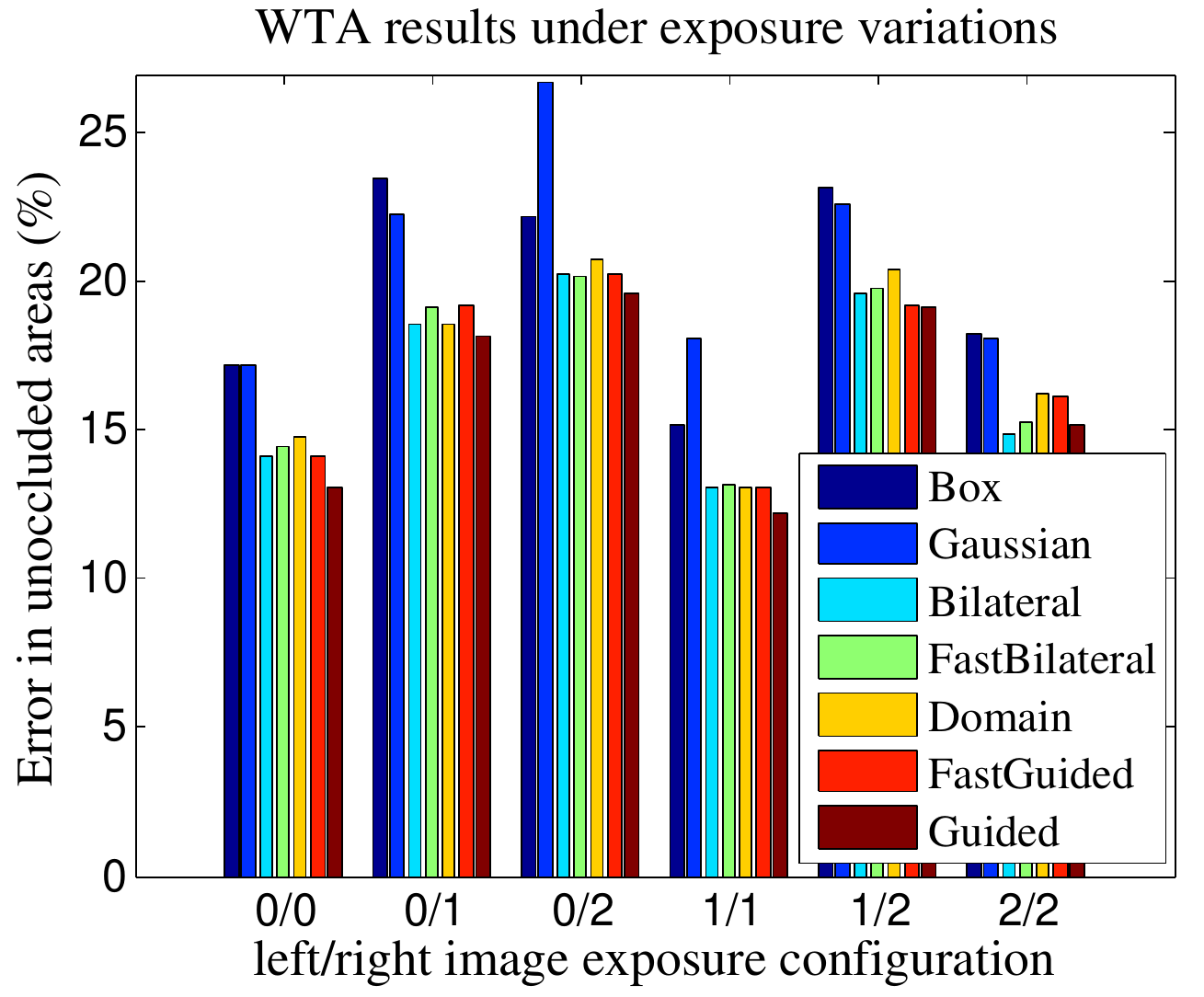}}\hfill
\vspace{-10pt}
\caption{Average bad-pixel error rate for the DASC descriptor as varying EAF including Box,
Gaussian, Bilateral \cite{Tomasi98}, FastBilateral \cite{Yang09}, Domain Transform \cite{Gastal11},
FastGF \cite{He15}, and GF \cite{He13} on Middlebury benchmark \cite{middlebury}.}\label{img:18}\vspace{-10pt}
\end{figure}
\subsection{Parameter and Component Analysis}\label{sec:62}
\subsubsection{Parameter sensitivity analysis}\label{sec:621}
\begin{figure}[t]
\centering
\renewcommand{\thesubfigure}{}
\subfigure[(a) Graffiti]
{\includegraphics[width=0.5\linewidth]{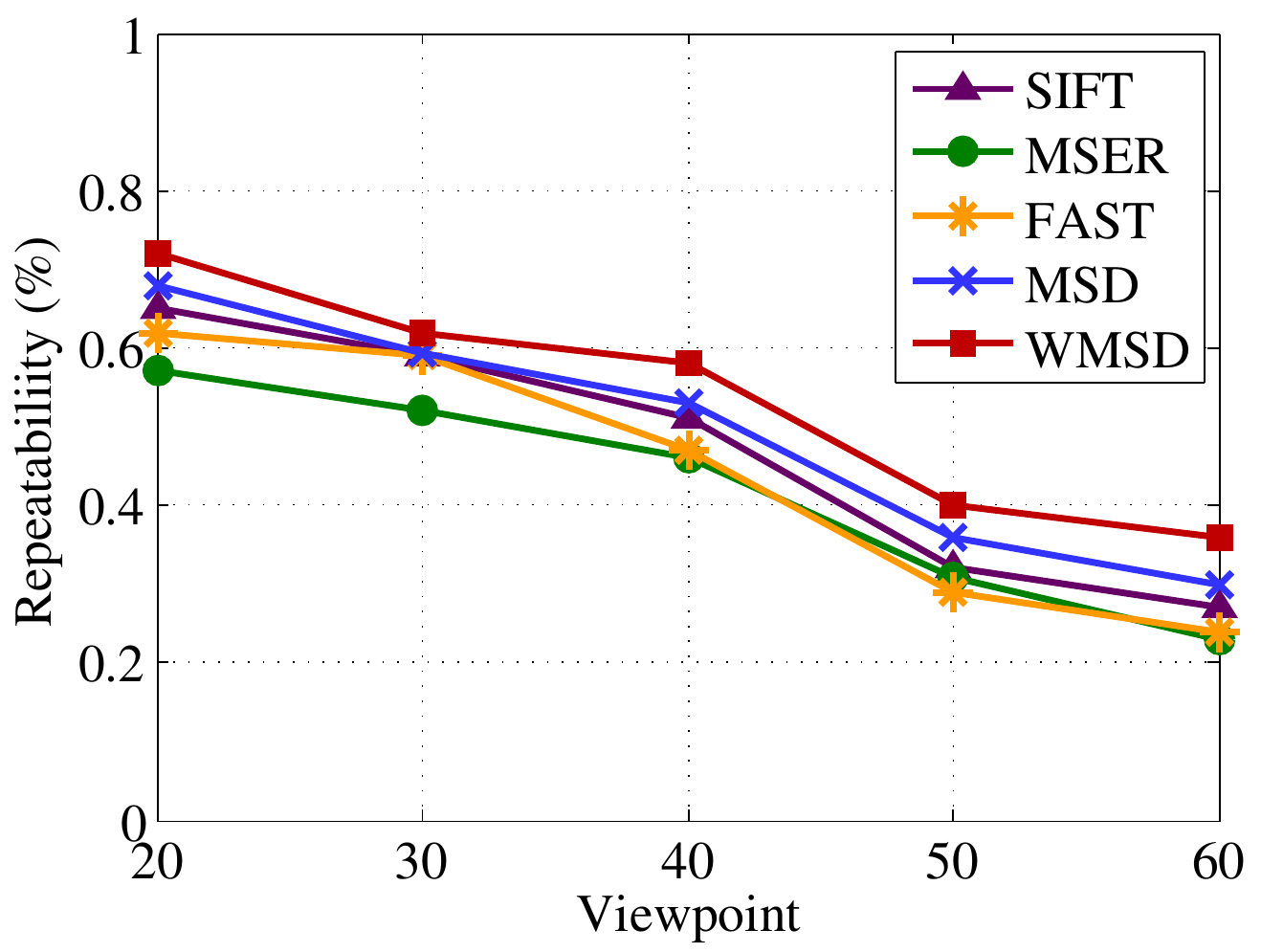}}\hfill
\subfigure[(b) Trees]
{\includegraphics[width=0.5\linewidth]{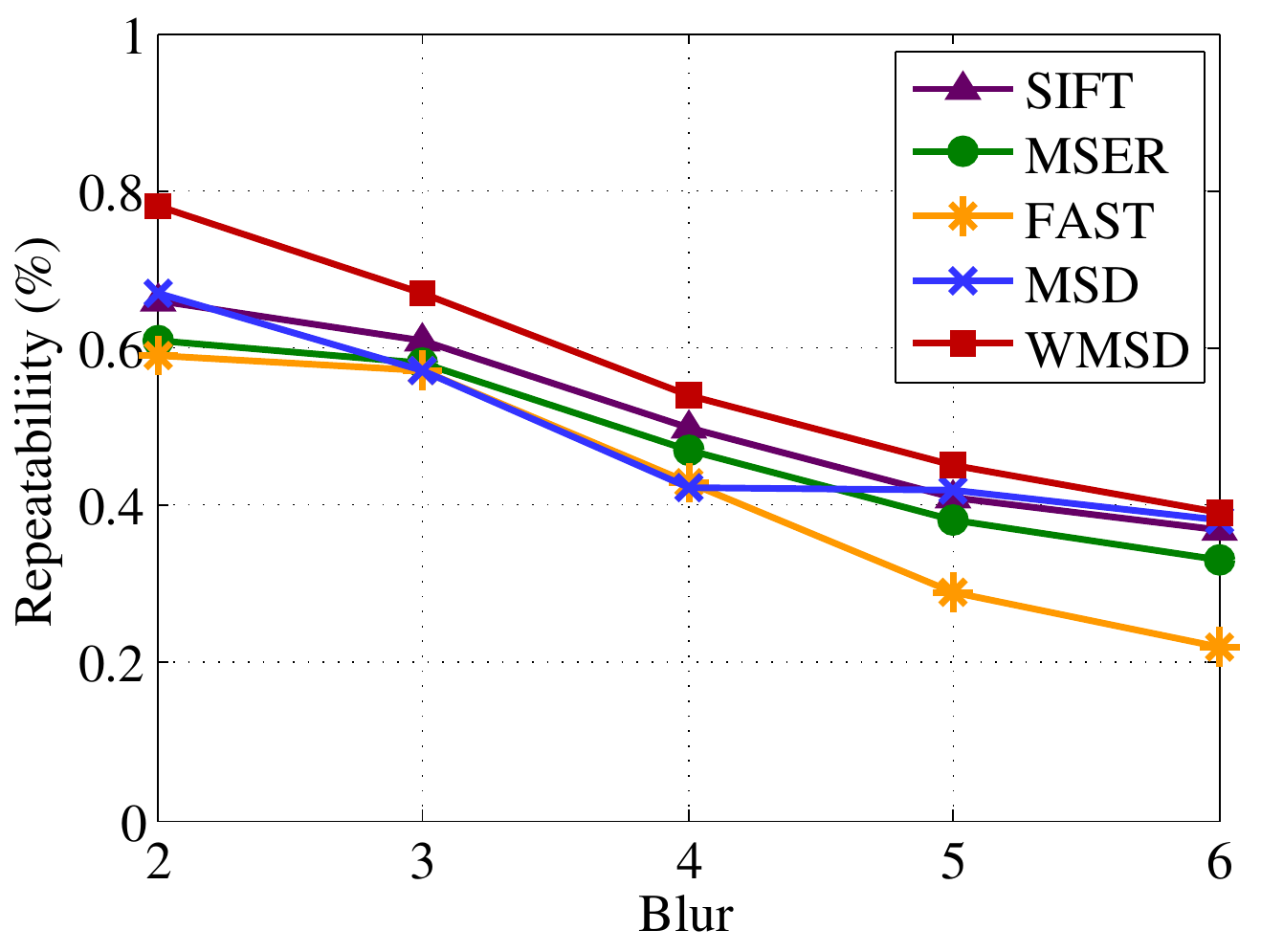}}\hfill
\vspace{-8pt}
\subfigure[(c) Bikes]
{\includegraphics[width=0.5\linewidth]{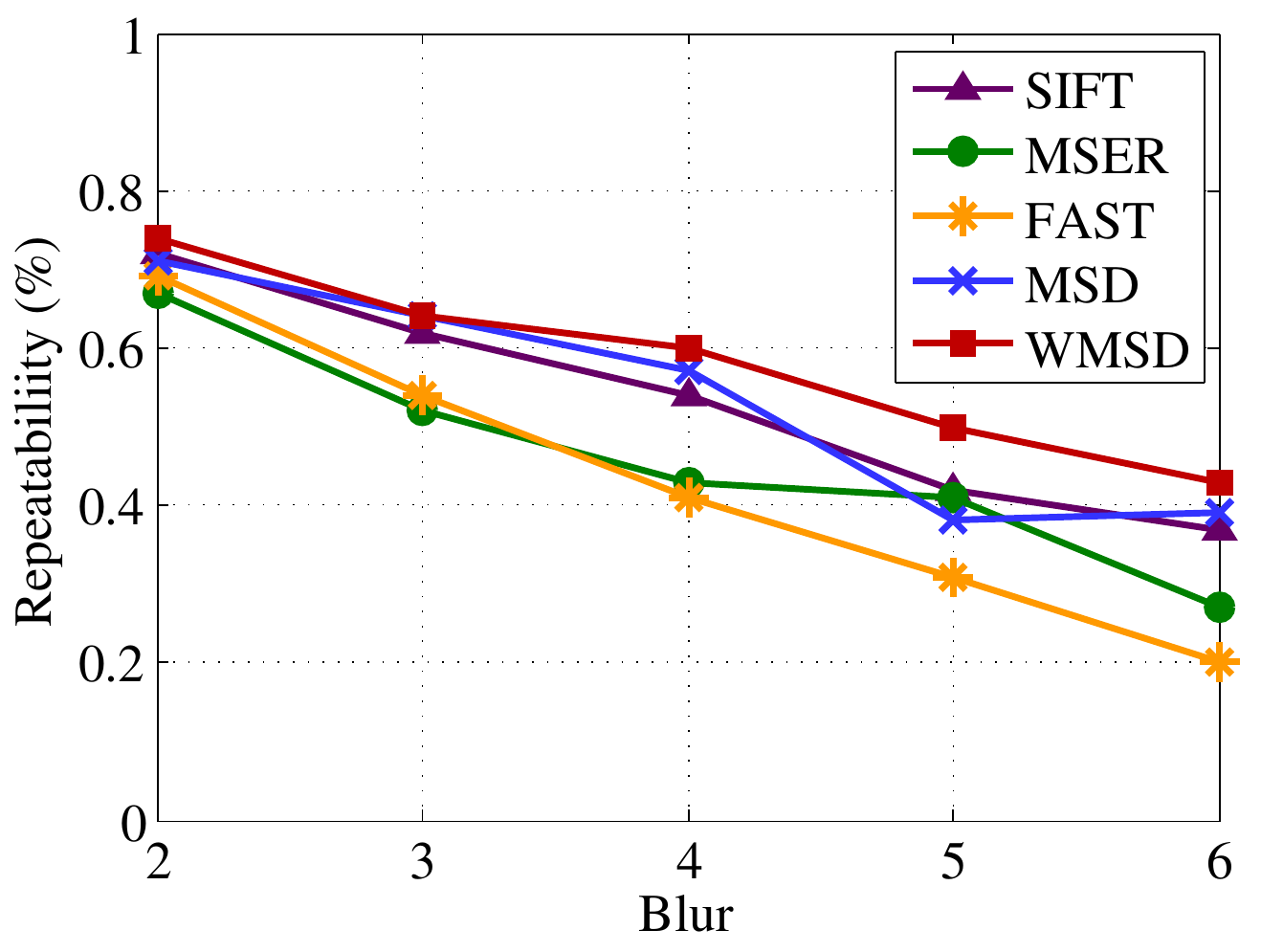}}\hfill
\subfigure[(d) Leuven]
{\includegraphics[width=0.5\linewidth]{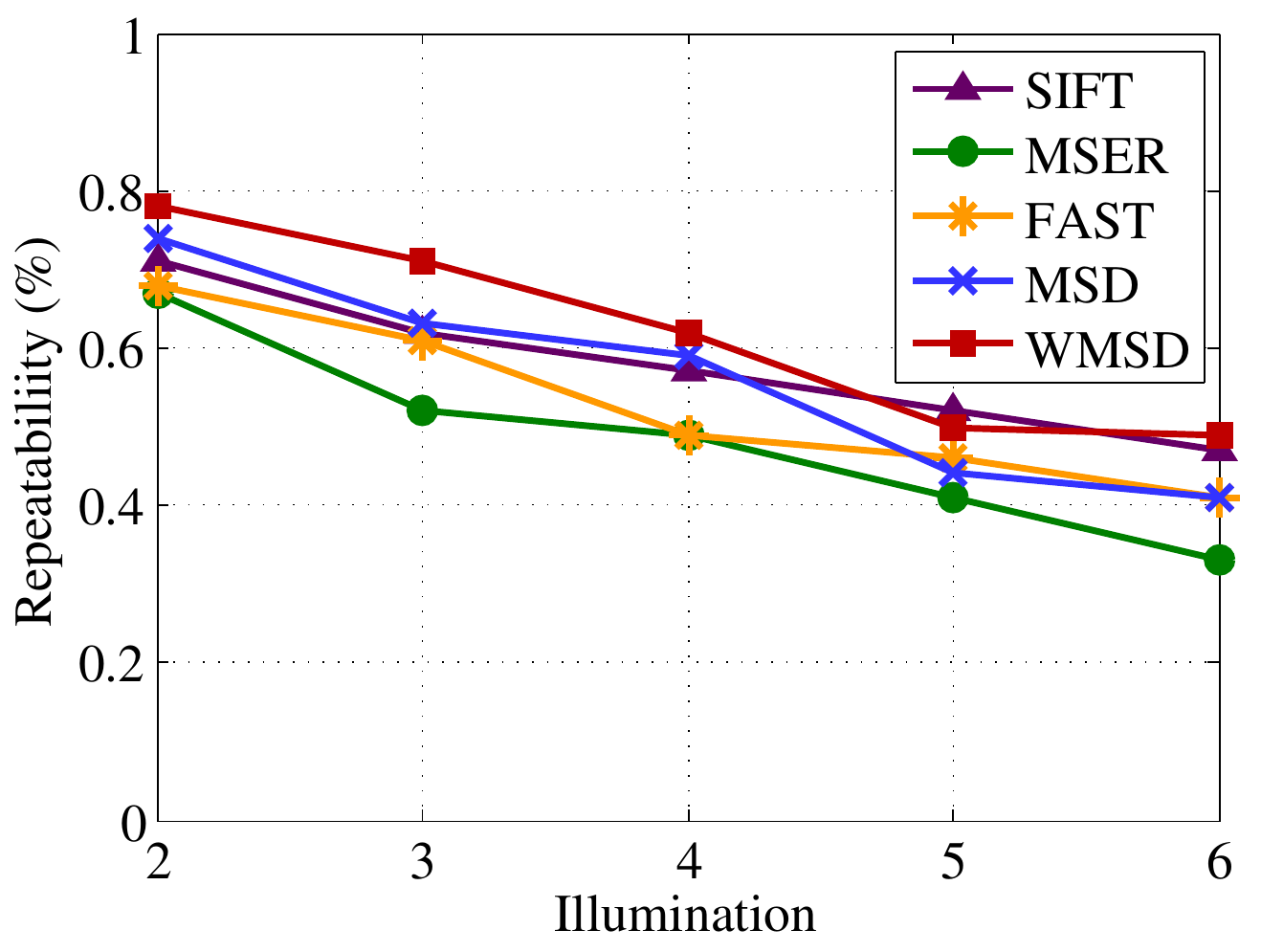}}\hfill
\vspace{-10pt}
\caption{Evaluation of the WMSD detection compared to conventional feature detections,
such as SIFT \cite{Lowe04}, MSER \cite{Matas02}, FAST \cite{Rosten10}, and MSD \cite{Tombari14}.
The WMSD provides reliable feature detection performance,
thus providing reliable hypothesis for initial sparse geometric fields.}\label{img:12}\vspace{-5pt}
\end{figure}
\begin{figure}[!t]
\centering
    \renewcommand{\thesubfigure}{}
    \subfigure[]
    {\includegraphics[width=0.5\linewidth]{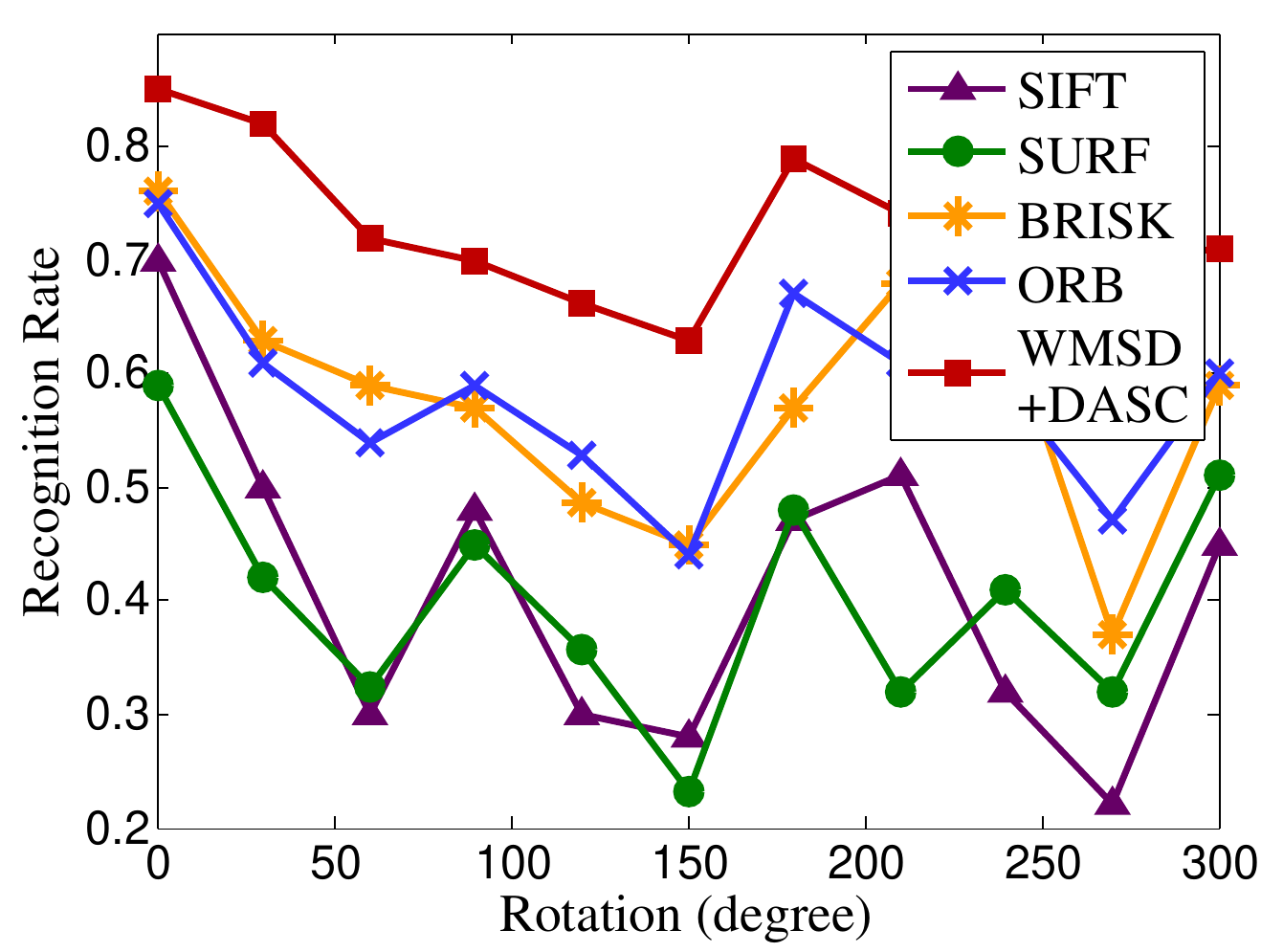}}
    \vspace{-20pt}
    \caption{Evaluation of the WMSD detection compared to conventional rotation estimations.
    Compared to conventional gradient-based rotation estimation (SIFT \cite{Lowe04} and SURF \cite{Bay06})
    or intensity-based rotation estimation (BRISK \cite{Leutenegger11} and ORB \cite{Rublee11}),
    our WMSD-based rotation estimation (with the DASC descriptor) shows the best performance.}\label{img:13}\vspace{-10pt}
\end{figure}
\figref{img:16} intensively analyzed the performance of the DASC
descriptor as varying associated parameters, including support
window size $M$, descriptor dimension $L^{\mathrm{dasc}}$, patch
size $N$, and the number of log-point circular point $N_c$. To
evaluate the quantitative performance, we measured an average
bad-pixel error rate on Middlebury benchmark \cite{middlebury}.
The larger the support window size $M$, the matching quality is
improved but the accuracy gain is saturated
around $31\times31$. Using a larger descriptor dimension
$L^{\mathrm{dasc}}$ yields a better performance since the descriptor
encodes more information. Considering the trade-off between efficiency and robustness,
$L^{\mathrm{dasc}}=128$ is set in experiments. When the patch size $N$ increases, the
matching quality is degraded since a series of similarity values
measured with large patches may lose locally discriminative
details. The number of log-polar circular point $N_c$ does not
affect the performance much, since optimal patterns can be
sampled even from small $N_c$.\vspace{-5pt}

\subsubsection{Component-wise performance gain analysis}\label{sec:622}
The DASC is originally motivated by the LSS concept from
\cite{Schechtman07}. The DASC consists of three key ingredients:
adaptive self-correlation (ASC), randomized pooling (RP), and
learning sampling pattern. In this context, we analyzed an accuracy gain
of the DASC over the LSS on the Middlebury benchmark as shown in
\figref{img:17}. Note that all experiments were done 
using LSS without max pooling, `LSS(wo/max)'. The original LSS method
\cite{Schechtman07} uses the SSD for measuring the patch similarity.
We replaced the patch similarity of the LSS method with
the ASC, named `LSS(ASC)', and then measured its matching accuracy.
As expected, the ASC improves the performance compared to the SSD
used in the original LSS. We also evaluated the LSS using a
randomized pooling with fixed center pixel, `LSS(ASC+RPF)', 
and the LSS using a learned randomized pooling with fixed center pixel, `LSS(ASC+LRPF)'. 
Unlike center-biased poolings, the DASC chooses sampling patterns randomly (`DASC+RP'), 
improving the performance.  Using learned
sampling patterns (`DASC+LRP') also leads to a performance gain. \vspace{-5pt}
\begin{table}[t]
	\caption{Evaluation of computational time. The brute-force and efficient computation of the DASC is denoted as \dag and \ddag, respectively.}\label{tab:1}\vspace{-15pt}
	\begin{center}
		\begin{tabular}{c>{\centering}m{0.12\linewidth}>{\centering}m{0.1\linewidth}>{\centering}m{0.1\linewidth}>{\centering}m{0.1\linewidth}>{\centering}m{0.1\linewidth}}
			\hlinewd{0.8pt}
			image size &SIFT &DAISY &LSS &DASC\dag &DASC\ddag \tabularnewline
			\hline
			\hline
			$463 \times 370$ &$130.3s$ &$2.5s$ &$31s$ &$128s$ &\bf{$1.3s$}\tabularnewline			
			$800 \times 600$ &$252s$ &$3.8s$ &$59s$ &$256s$ &\bf{$2.1s$}\tabularnewline
			\hlinewd{0.8pt}
		\end{tabular}
	\end{center}\vspace{-15pt}
\end{table}
\begin{figure}[t]
	\centering
	\renewcommand{\thesubfigure}{}
	\subfigure[(a) Illumination variation]
	{\includegraphics[width=0.5\linewidth]{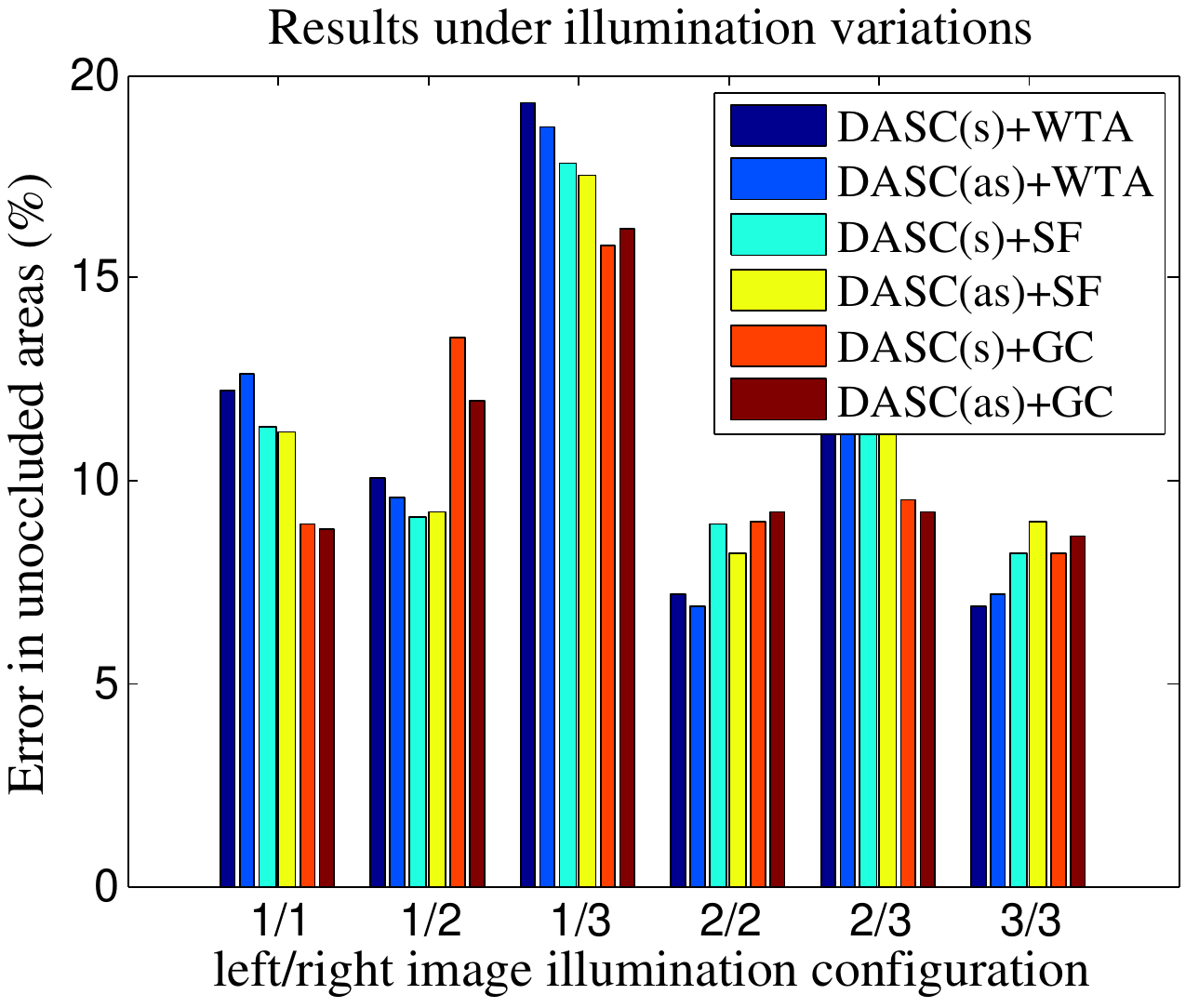}}\hfill
	\subfigure[(b) Exposure variation]
	{\includegraphics[width=0.5\linewidth]{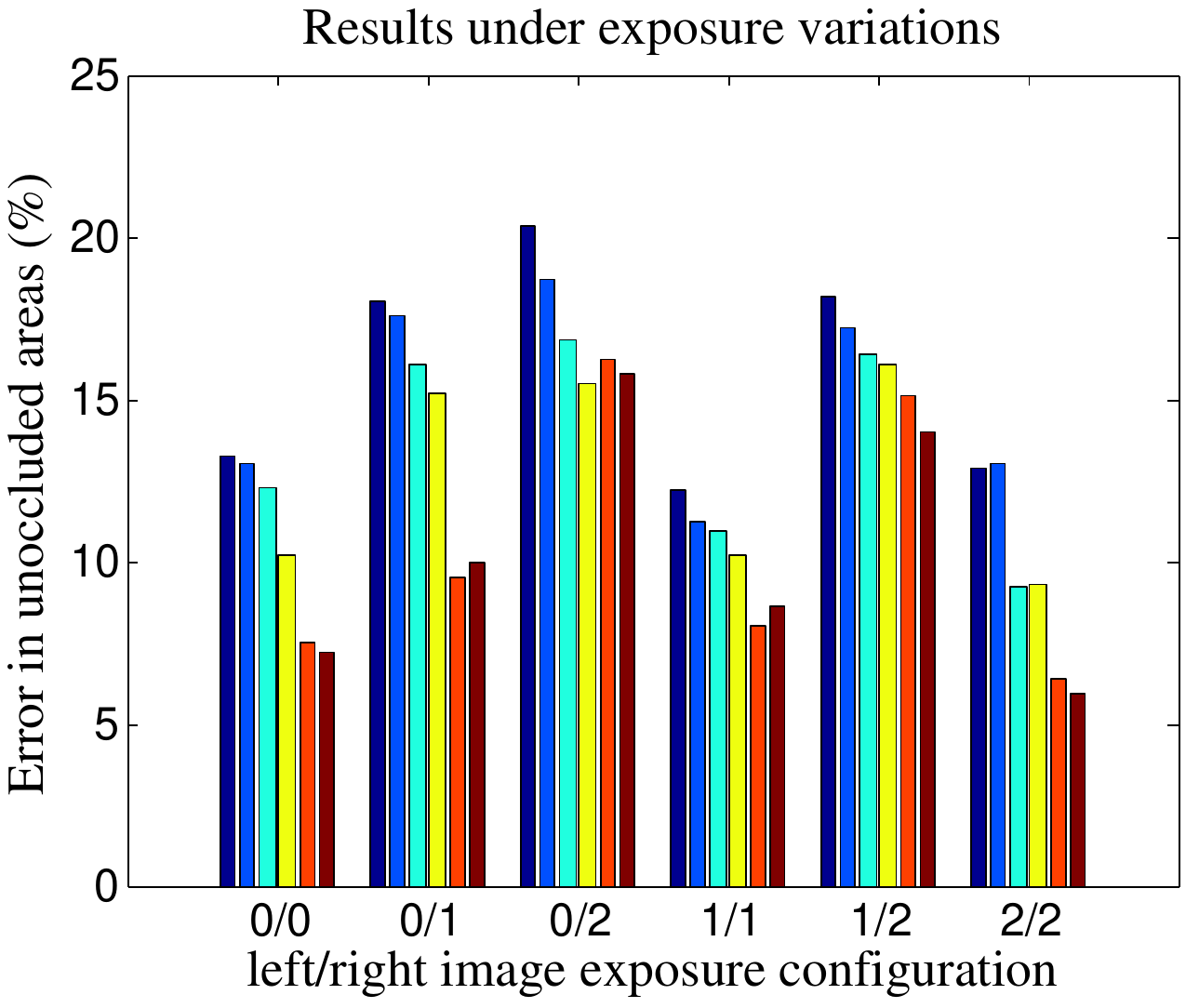}}\hfill
	\vspace{-10pt}
	\caption{Evaluation of a symmetric measure $\Psi (i,j)$ and an asymmetric measure $\tilde{\Psi} (i,j)$ in the DASC
		as varying optimization schemes with WTA, SF \cite{Liu11}, and GC \cite{Boykov01}.
		It shows that there are no significant performance gaps when using symmetric and asymmetric measure.}\label{img:8}\vspace{-10pt}
\end{figure}
\begin{figure}[!t]
	\centering
	\renewcommand{\thesubfigure}{}
	\subfigure[(a) Illumination variation]
	{\includegraphics[width=0.5\linewidth]{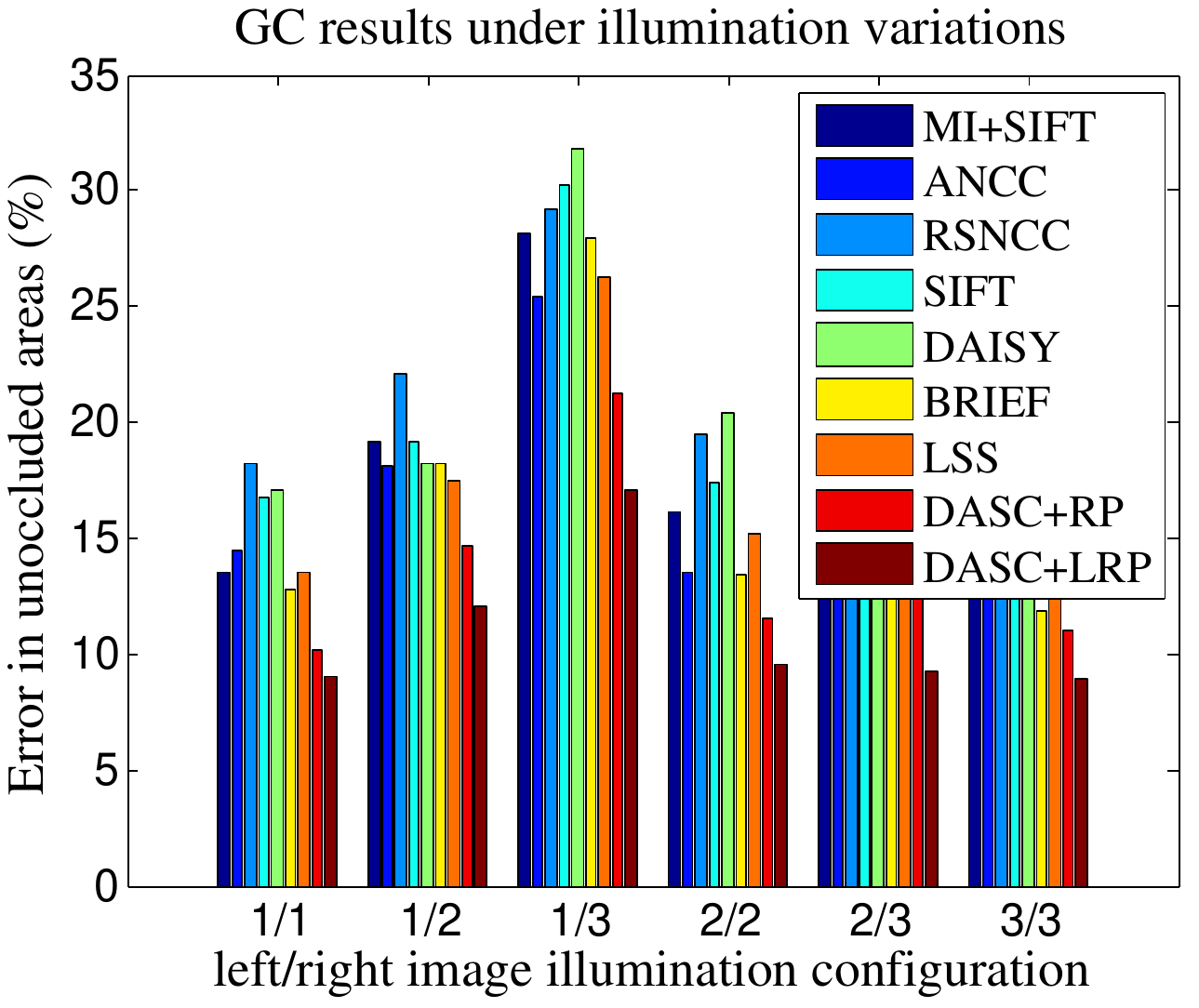}}\hfill
	\subfigure[(b) Exposure variation]
	{\includegraphics[width=0.5\linewidth]{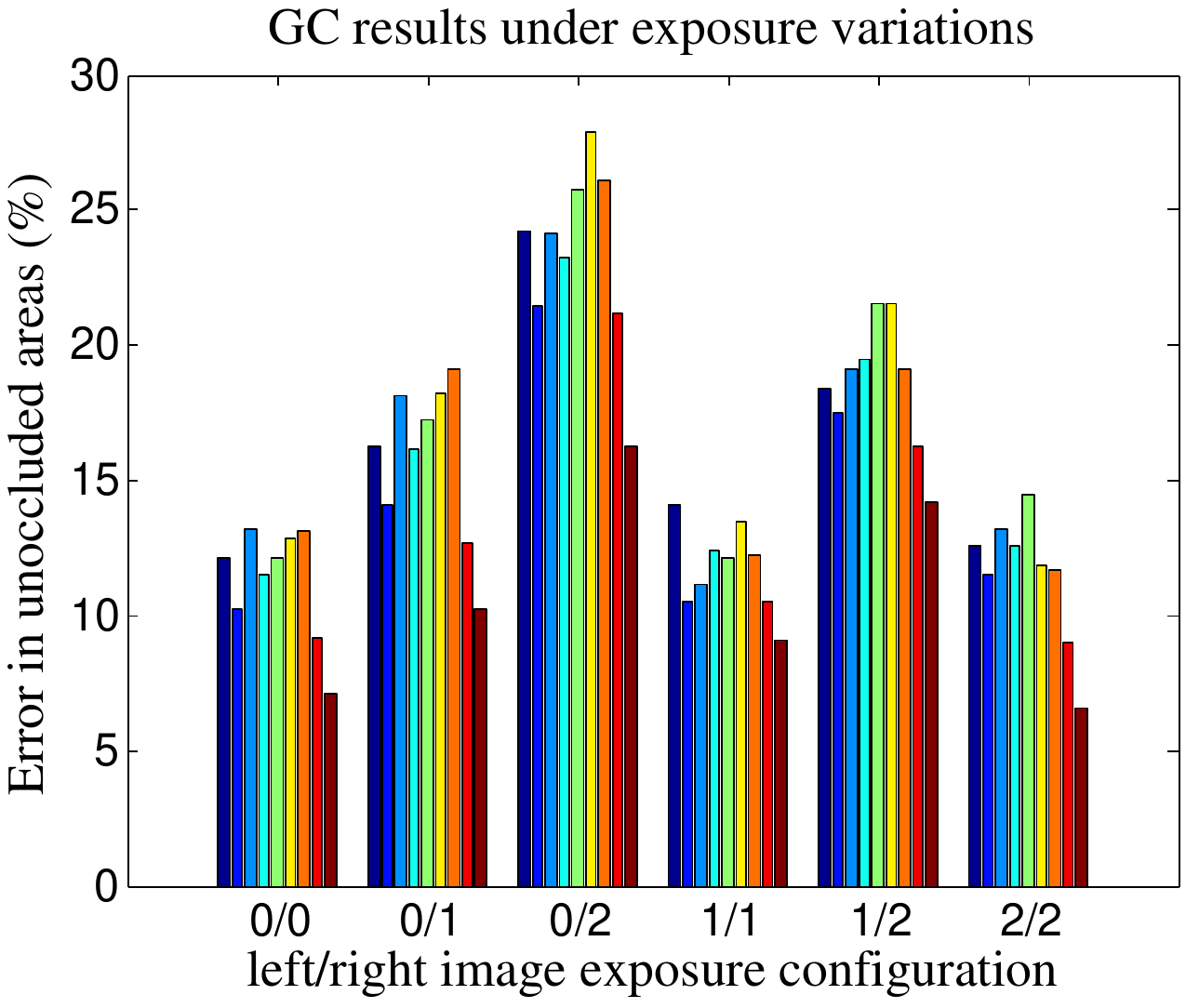}}\hfill
	\vspace{-8pt}
	\subfigure[(c) Illumination variation]
	{\includegraphics[width=0.5\linewidth]{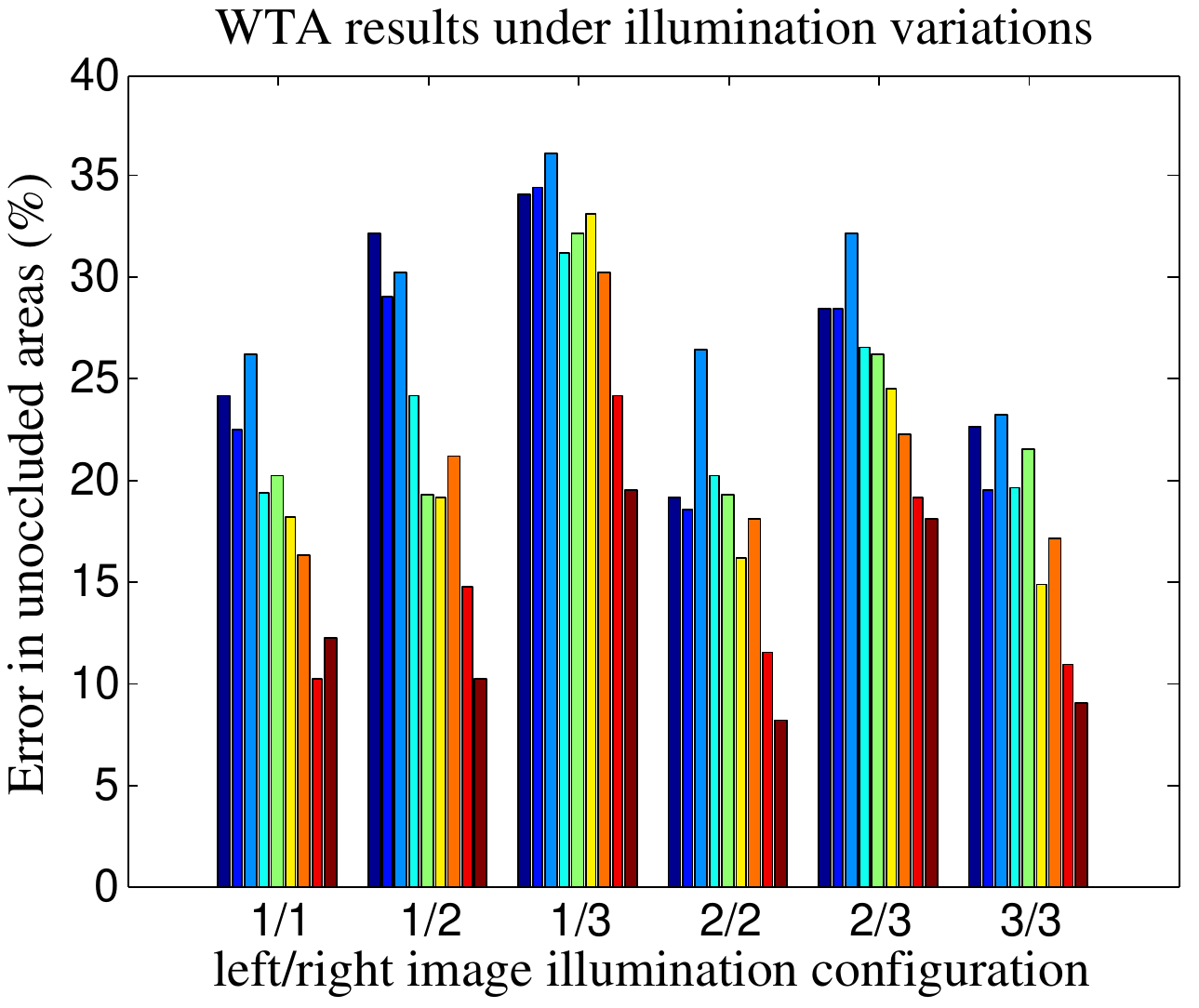}}\hfill
	\subfigure[(d) Exposure variation]
	{\includegraphics[width=0.5\linewidth]{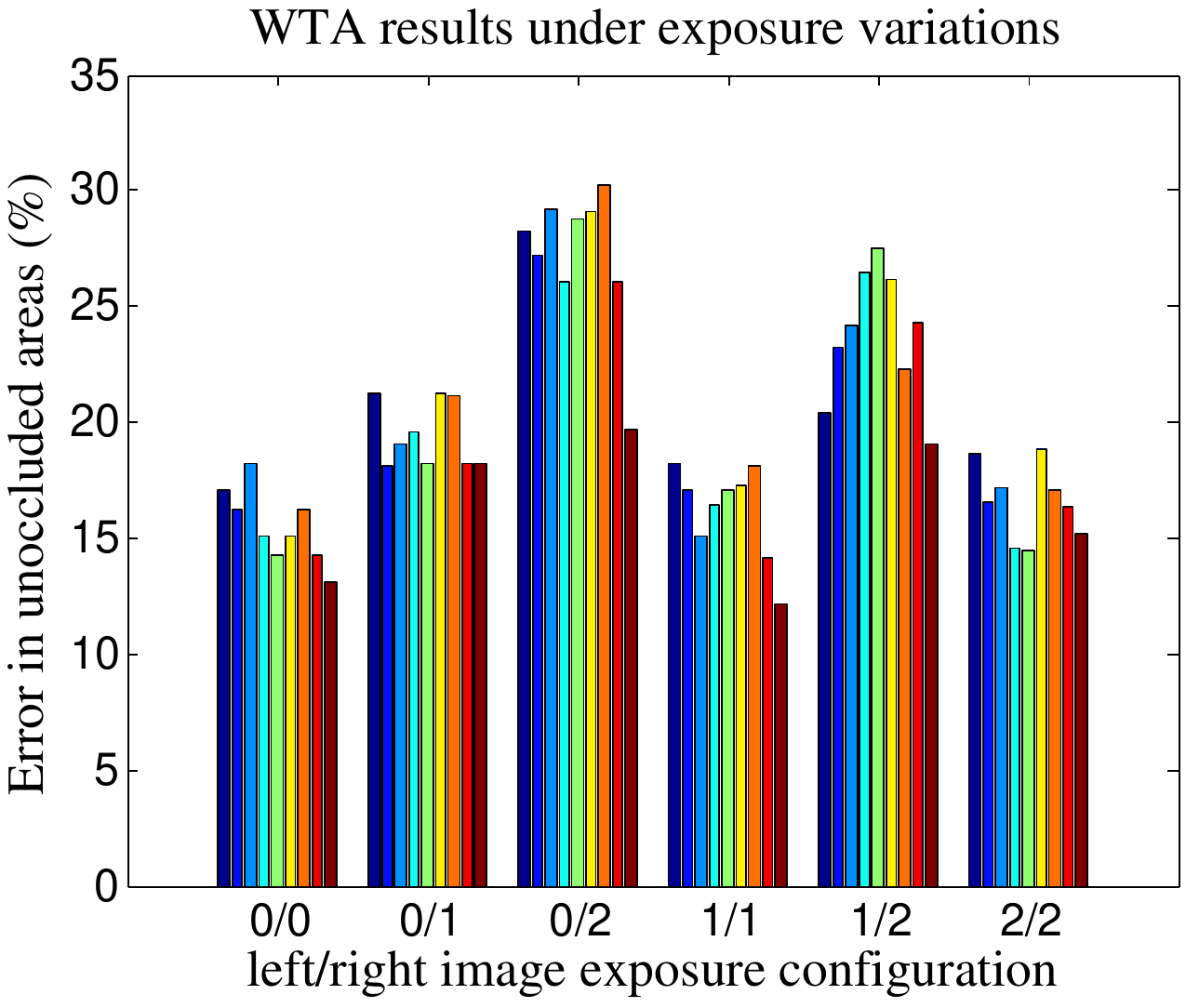}}\hfill
	\vspace{-10pt}
	\caption{Average bad-pixel error rate on Middlebury
		benchmark with illumination variations and exposure variations. The
		GC (first row) and WTA (second row) were used for optimization,
		respectively. Our DASC+LRP shows the best performance with the lowest error rate.}\label{img:19}\vspace{-10pt}
\end{figure}
\subsubsection{Edge-aware filtering analysis}\label{sec:623}
In \figref{img:18}, we analyzed the performance of the DASC
descriptor when different EAF is employed for computing
$\omega_{i,i'}$. When using a simple, unweighted `Box' filtering
($\omega_{i,i'}=1$), the patch similarity \equref{equ:asc} becomes a
normalized cross-correlation (NCC). In the Box and Gaussian
filtering case, there exists a performance limitation. In contrast,
all EAF methods show a satisfactory performance, including the
bilateral filter \cite{Tomasi98}, the fast bilateral filter
\cite{Paris09}, the domain transform \cite{Gastal11}, the fast GF
\cite{He15}, and GF \cite{He10}. 
In experiments, we utilized the GF \cite{He10}. \vspace{-5pt}
\begin{figure*}[t]
	\centering
	\renewcommand{\thesubfigure}{}
	\subfigure[]
	{\includegraphics[width=0.125\linewidth]{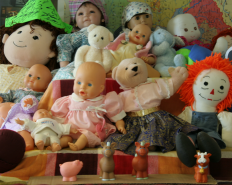}}\hfill
	\subfigure[]
	{\includegraphics[width=0.125\linewidth]{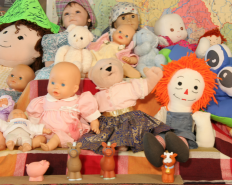}}\hfill
	\subfigure[]
	{\includegraphics[width=0.125\linewidth]{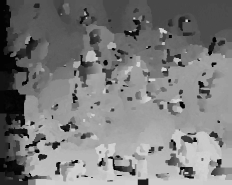}}\hfill
	\subfigure[]
	{\includegraphics[width=0.125\linewidth]{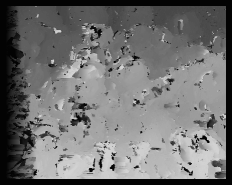}}\hfill
	\subfigure[]
	{\includegraphics[width=0.125\linewidth]{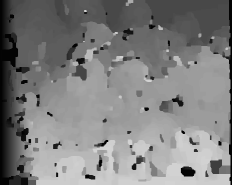}}\hfill
	\subfigure[]
	{\includegraphics[width=0.125\linewidth]{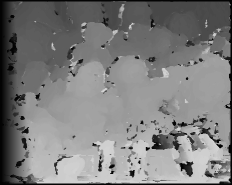}}\hfill
	\subfigure[]
	{\includegraphics[width=0.125\linewidth]{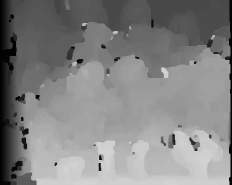}}\hfill
	\subfigure[]
	{\includegraphics[width=0.125\linewidth]{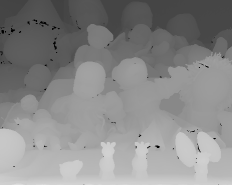}}\hfill
	\vspace{-22pt}
	\subfigure[(a) Left image]
	{\includegraphics[width=0.125\linewidth]{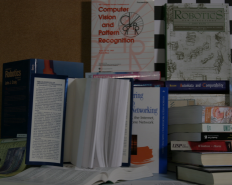}}\hfill
	\subfigure[(b) Right image]
	{\includegraphics[width=0.125\linewidth]{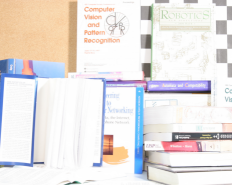}}\hfill
	\subfigure[(c) ANCC \cite{Heo}]
	{\includegraphics[width=0.125\linewidth]{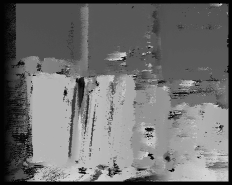}}\hfill
	\subfigure[(d) BRIEF \cite{Calonder11}]
	{\includegraphics[width=0.125\linewidth]{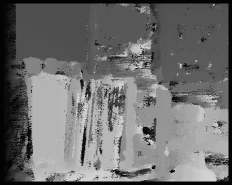}}\hfill
	\subfigure[(e) SIFT \cite{Lowe04}]
	{\includegraphics[width=0.125\linewidth]{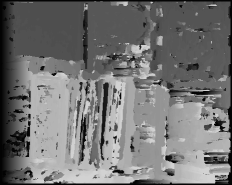}}\hfill
	\subfigure[(f) LSS \cite{Schechtman07}]
	{\includegraphics[width=0.125\linewidth]{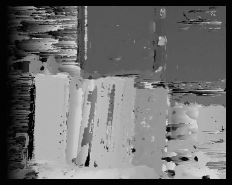}}\hfill
	\subfigure[(g) DASC+LRP]
	{\includegraphics[width=0.125\linewidth]{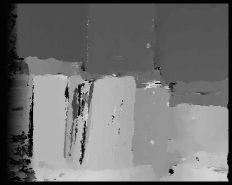}}\hfill
	\subfigure[(h) Ground Truth]
	{\includegraphics[width=0.125\linewidth]{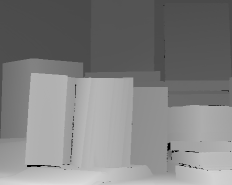}}\hfill
	\vspace{-10pt}
	\caption{Comparison of disparity estimation for \emph{Dolls} and \emph{Books} image pairs
		under illumination combination `1/3' and exposure combination `0/2', respectively. Compared to other approaches,
		our DASC descriptor estimates accurate and edge-preserved disparity maps while reducing artifacts.}\label{img:20}\vspace{-8pt}
\end{figure*}
\begin{figure*}[t!]
	\centering
	\renewcommand{\thesubfigure}{}
	\subfigure[]
	{\includegraphics[width=0.125\linewidth]{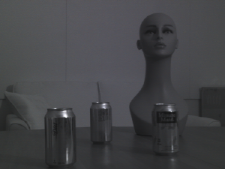}}\hfill
	\subfigure[]
	{\includegraphics[width=0.125\linewidth]{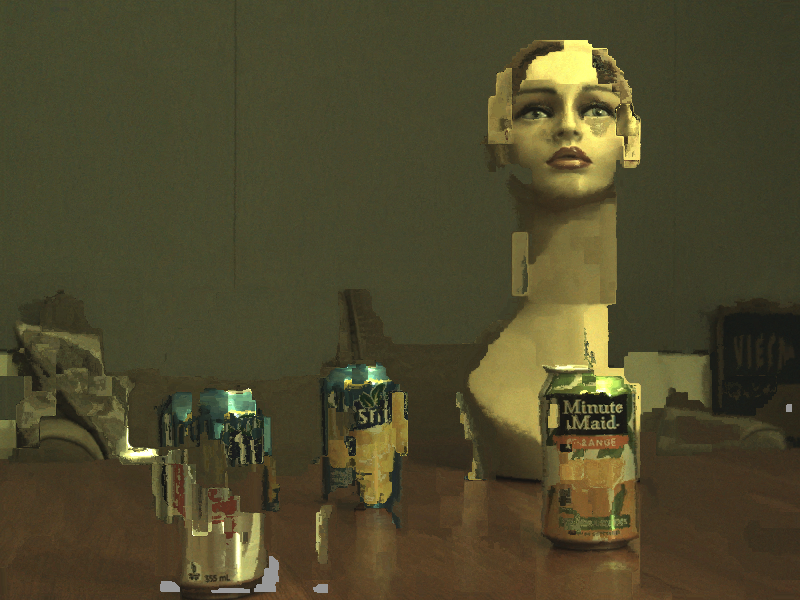}}\hfill
	\subfigure[]
	{\includegraphics[width=0.125\linewidth]{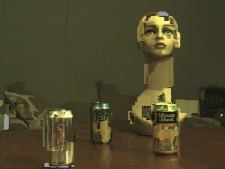}}\hfill
	\subfigure[]
	{\includegraphics[width=0.125\linewidth]{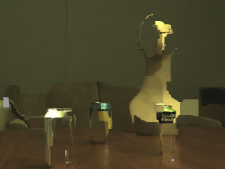}}\hfill
	\subfigure[]
	{\includegraphics[width=0.125\linewidth]{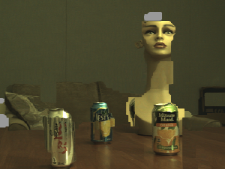}}\hfill
	\subfigure[]
	{\includegraphics[width=0.125\linewidth]{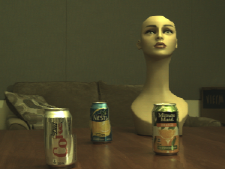}}\hfill
	\subfigure[]
	{\includegraphics[width=0.125\linewidth]{Figure/Figure21/RGB_NIR/f7_DASC.png}}\hfill
	\subfigure[]
	{\includegraphics[width=0.125\linewidth]{Figure/Figure21/RGB_NIR/f7_DASC.png}}\hfill
	\vspace{-22pt}
	\subfigure[]
	{\includegraphics[width=0.125\linewidth]{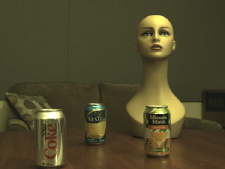}}\hfill
	\subfigure[]
	{\includegraphics[width=0.125\linewidth]{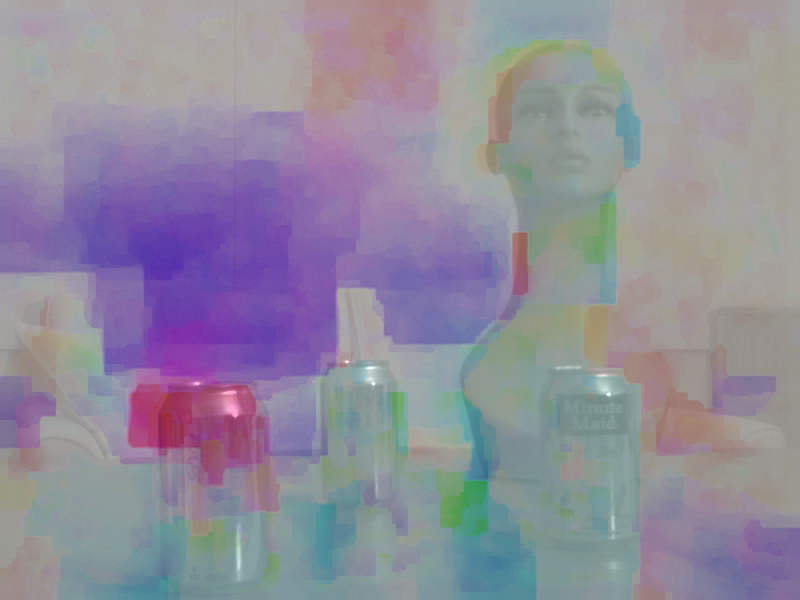}}\hfill
	\subfigure[]
	{\includegraphics[width=0.125\linewidth]{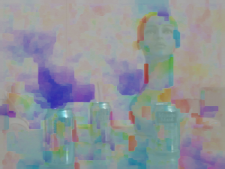}}\hfill
	\subfigure[]
	{\includegraphics[width=0.125\linewidth]{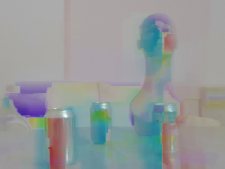}}\hfill
	\subfigure[]
	{\includegraphics[width=0.125\linewidth]{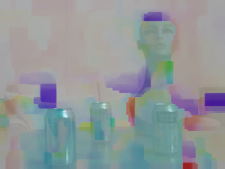}}\hfill
	\subfigure[]
	{\includegraphics[width=0.125\linewidth]{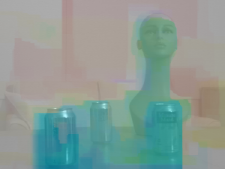}}\hfill
	\subfigure[]
	{\includegraphics[width=0.125\linewidth]{Figure/Figure21/RGB_NIR/f7_DASC_flow.png}}\hfill
	\subfigure[]
	{\includegraphics[width=0.125\linewidth]{Figure/Figure21/RGB_NIR/f7_DASC_flow.png}}\hfill
	\vspace{-22pt}
	\subfigure[]
	{\includegraphics[width=0.125\linewidth]{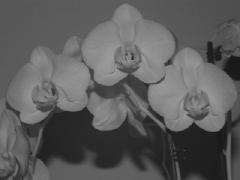}}\hfill
	\subfigure[]
	{\includegraphics[width=0.125\linewidth]{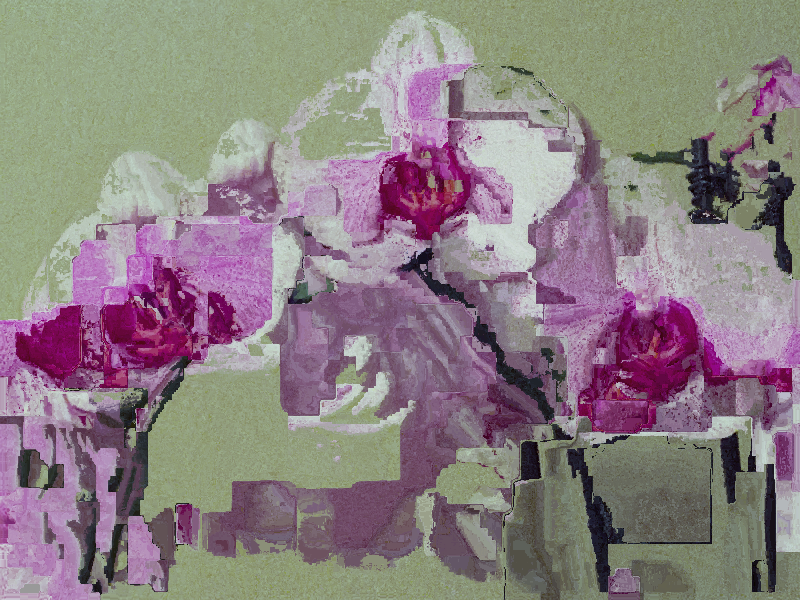}}\hfill
	\subfigure[]
	{\includegraphics[width=0.125\linewidth]{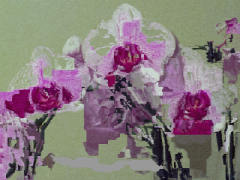}}\hfill
	\subfigure[]
	{\includegraphics[width=0.125\linewidth]{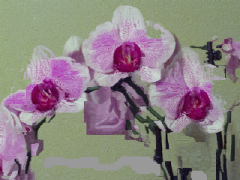}}\hfill
	\subfigure[]
	{\includegraphics[width=0.125\linewidth]{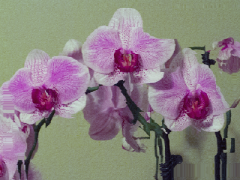}}\hfill
	\subfigure[]
	{\includegraphics[width=0.125\linewidth]{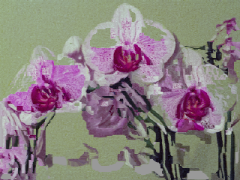}}\hfill
	\subfigure[]
	{\includegraphics[width=0.125\linewidth]{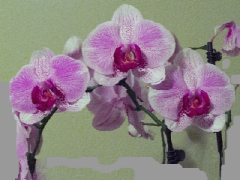}}\hfill
	\subfigure[]
	{\includegraphics[width=0.125\linewidth]{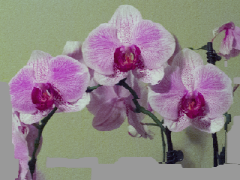}}\hfill
	\vspace{-22pt}
	\subfigure[]
	{\includegraphics[width=0.125\linewidth]{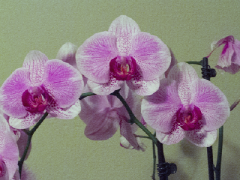}}\hfill
	\subfigure[]
	{\includegraphics[width=0.125\linewidth]{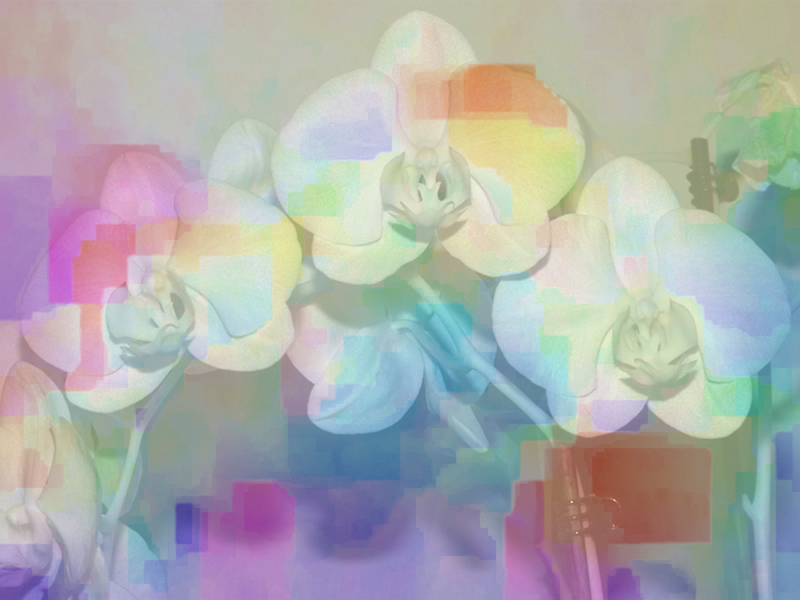}}\hfill
	\subfigure[]
	{\includegraphics[width=0.125\linewidth]{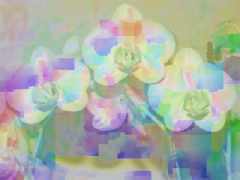}}\hfill
	\subfigure[]
	{\includegraphics[width=0.125\linewidth]{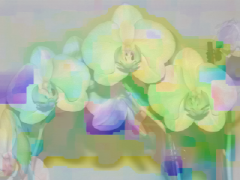}}\hfill
	\subfigure[]
	{\includegraphics[width=0.125\linewidth]{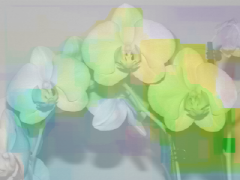}}\hfill
	\subfigure[]
	{\includegraphics[width=0.125\linewidth]{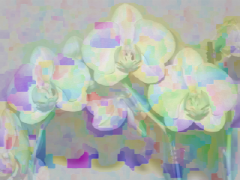}}\hfill
	\subfigure[]
	{\includegraphics[width=0.125\linewidth]{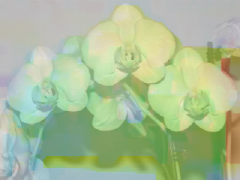}}\hfill
	\subfigure[]
	{\includegraphics[width=0.125\linewidth]{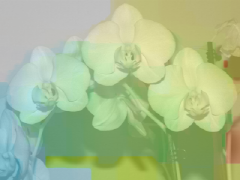}}\hfill
	\vspace{-22pt}
	\subfigure[]
	{\includegraphics[width=0.125\linewidth]{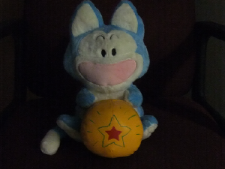}}\hfill
	\subfigure[]
	{\includegraphics[width=0.125\linewidth]{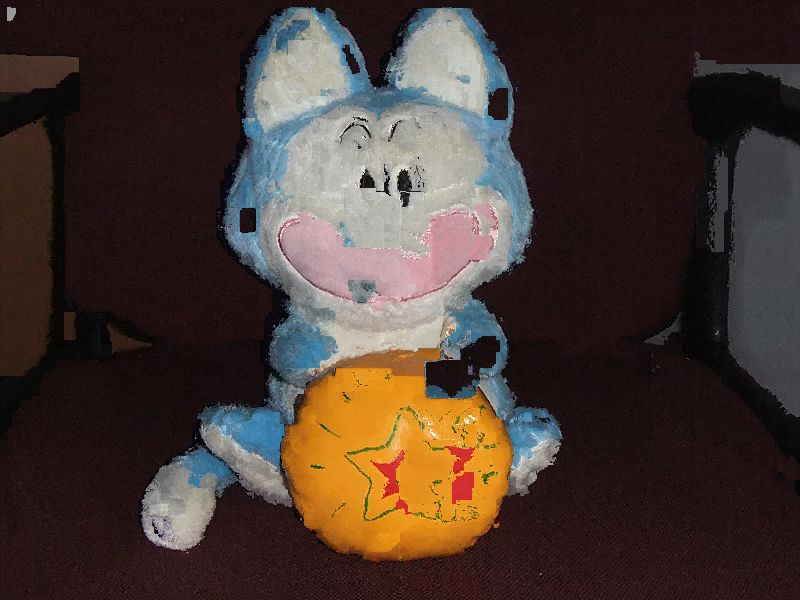}}\hfill
	\subfigure[]
	{\includegraphics[width=0.125\linewidth]{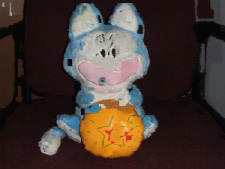}}\hfill
	\subfigure[]
	{\includegraphics[width=0.125\linewidth]{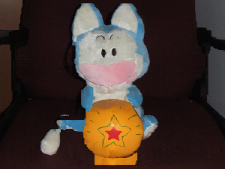}}\hfill
	\subfigure[]
	{\includegraphics[width=0.125\linewidth]{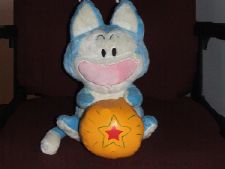}}\hfill
	\subfigure[]
	{\includegraphics[width=0.125\linewidth]{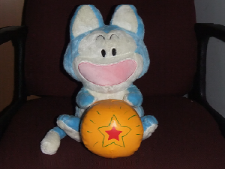}}\hfill
	\subfigure[]
	{\includegraphics[width=0.125\linewidth]{Figure/Figure21/Flash_noflash/f7_LSS.png}}\hfill
	\subfigure[]
	{\includegraphics[width=0.125\linewidth]{Figure/Figure21/Flash_noflash/f7_DASC.png}}\hfill
	\vspace{-22pt}
	\subfigure[]
	{\includegraphics[width=0.125\linewidth]{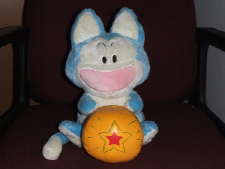}}\hfill
	\subfigure[]
	{\includegraphics[width=0.125\linewidth]{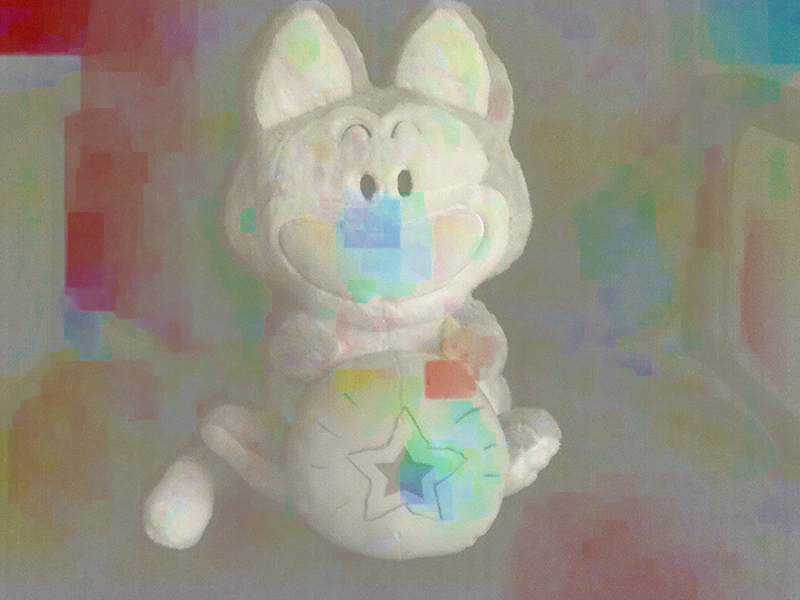}}\hfill
	\subfigure[]
	{\includegraphics[width=0.125\linewidth]{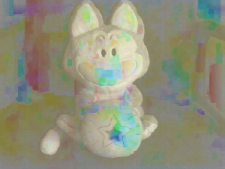}}\hfill
	\subfigure[]
	{\includegraphics[width=0.125\linewidth]{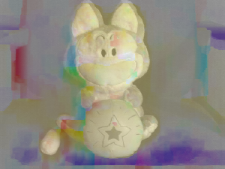}}\hfill
	\subfigure[]
	{\includegraphics[width=0.125\linewidth]{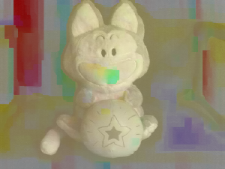}}\hfill
	\subfigure[]
	{\includegraphics[width=0.125\linewidth]{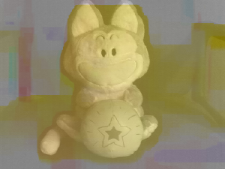}}\hfill
	\subfigure[]
	{\includegraphics[width=0.125\linewidth]{Figure/Figure21/Flash_noflash/f7_LSS_flow.png}}\hfill
	\subfigure[]
	{\includegraphics[width=0.125\linewidth]{Figure/Figure21/Flash_noflash/f7_DASC_flow.png}}\hfill
	\vspace{-22pt}
	\subfigure[]
	{\includegraphics[width=0.125\linewidth]{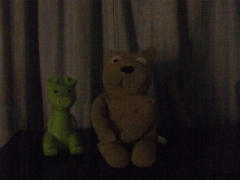}}\hfill
	\subfigure[]
	{\includegraphics[width=0.125\linewidth]{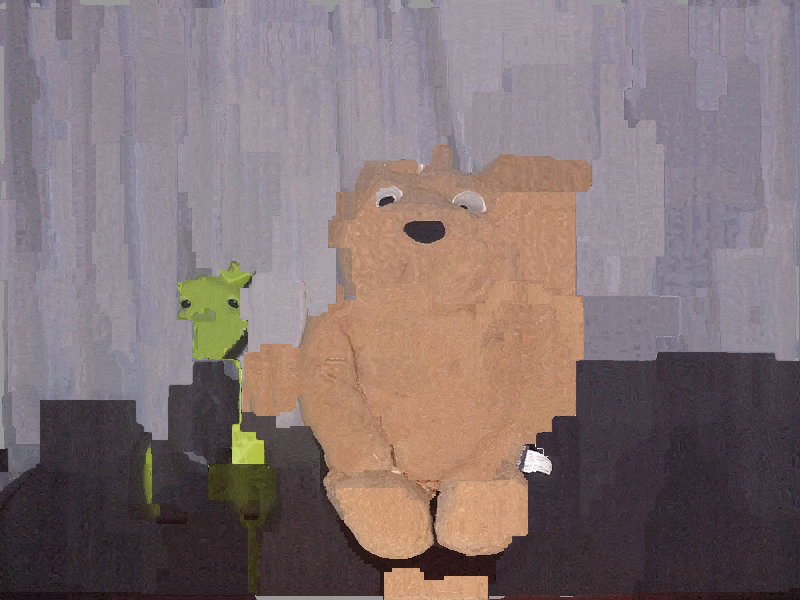}}\hfill
	\subfigure[]
	{\includegraphics[width=0.125\linewidth]{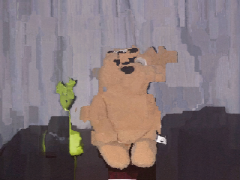}}\hfill
	\subfigure[]
	{\includegraphics[width=0.125\linewidth]{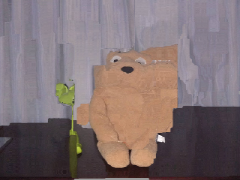}}\hfill
	\subfigure[]
	{\includegraphics[width=0.125\linewidth]{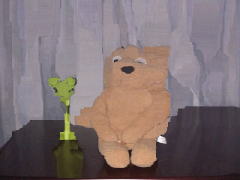}}\hfill
	\subfigure[]
	{\includegraphics[width=0.125\linewidth]{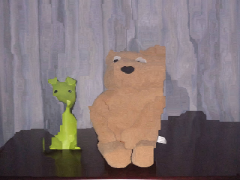}}\hfill
	\subfigure[]
	{\includegraphics[width=0.125\linewidth]{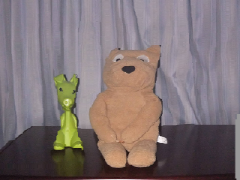}}\hfill
	\subfigure[]
	{\includegraphics[width=0.125\linewidth]{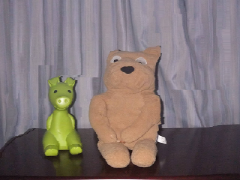}}\hfill
	\vspace{-22pt}
	\subfigure[(a) Image pairs]
	{\includegraphics[width=0.125\linewidth]{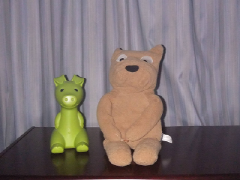}}\hfill
	\subfigure[(b) MI+SIFT \cite{Heo13}]
	{\includegraphics[width=0.125\linewidth]{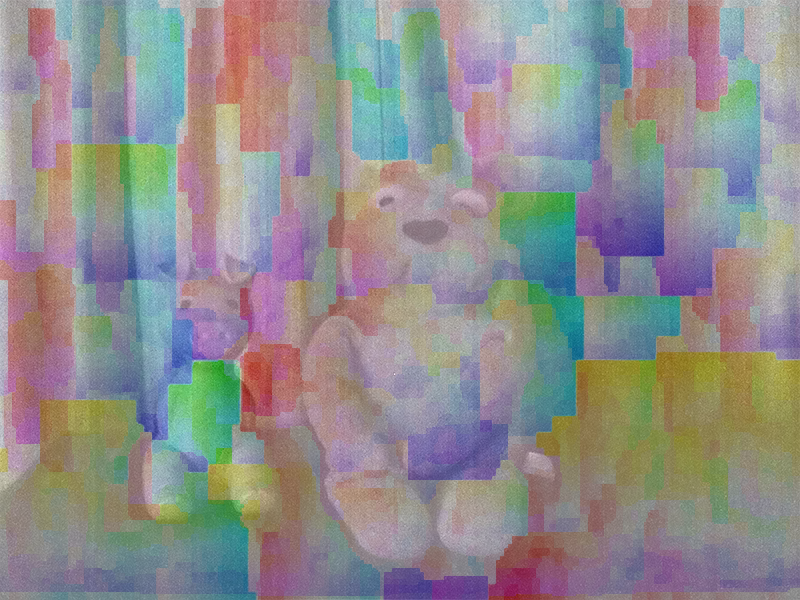}}\hfill
	\subfigure[(c) BRIEF \cite{Calonder11}]
	{\includegraphics[width=0.125\linewidth]{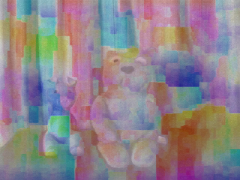}}\hfill
	\subfigure[(d) DAISY \cite{Tola10}]
	{\includegraphics[width=0.125\linewidth]{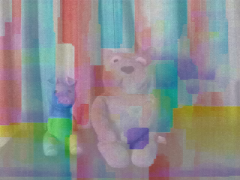}}\hfill
	\subfigure[(e) SIFT \cite{Lowe04}]
	{\includegraphics[width=0.125\linewidth]{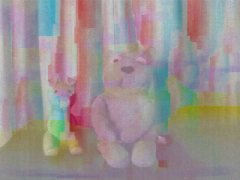}}\hfill
	\subfigure[(f) LSS \cite{Schechtman07}]
	{\includegraphics[width=0.125\linewidth]{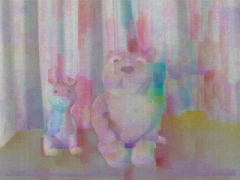}}\hfill
	\subfigure[(g) DASC+RP]
	{\includegraphics[width=0.125\linewidth]{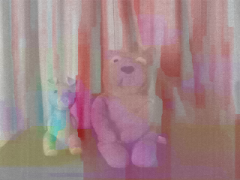}}\hfill
	\subfigure[(h) DASC+LRP]
	{\includegraphics[width=0.125\linewidth]{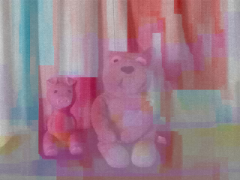}}\hfill
	\vspace{-10pt}
	\caption{Comparison of dense correspondence for
		(from top to bottom) RGB-NIR images and flash-noflash images. The results consist of warped color images
		and correspondence flow fields overlaid with reference images. Compared to other conventional approaches, our DASC+LRP descriptor estimates reliable dense
		correspondence fields for challenging multi-modal and multi-spectral image pairs.}\label{img:21}\vspace{-10pt}
\end{figure*}
\begin{figure*}[t]
	\centering
	\renewcommand{\thesubfigure}{}
	\subfigure[]
	{\includegraphics[width=0.125\linewidth]{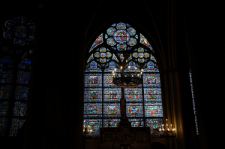}}\hfill
	\subfigure[]
	{\includegraphics[width=0.125\linewidth]{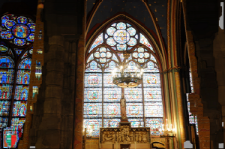}}\hfill
	\subfigure[]
	{\includegraphics[width=0.125\linewidth]{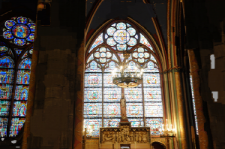}}\hfill
	\subfigure[]
	{\includegraphics[width=0.125\linewidth]{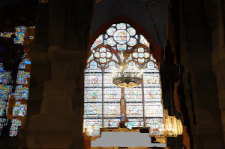}}\hfill
	\subfigure[]
	{\includegraphics[width=0.125\linewidth]{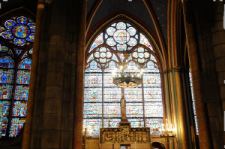}}\hfill
	\subfigure[]
	{\includegraphics[width=0.125\linewidth]{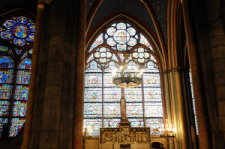}}\hfill
	\subfigure[]
	{\includegraphics[width=0.125\linewidth]{Figure/Figure22/Different_Exposure/f7_LSS.png}}\hfill
	\subfigure[]
	{\includegraphics[width=0.125\linewidth]{Figure/Figure22/Different_Exposure/f7_DASC.png}}\hfill
	\vspace{-22pt}
	\subfigure[]
	{\includegraphics[width=0.125\linewidth]{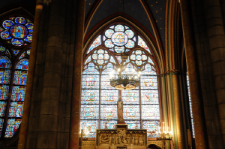}}\hfill
	\subfigure[]
	{\includegraphics[width=0.125\linewidth]{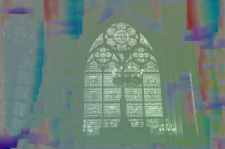}}\hfill
	\subfigure[]
	{\includegraphics[width=0.125\linewidth]{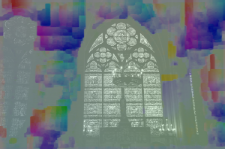}}\hfill
	\subfigure[]
	{\includegraphics[width=0.125\linewidth]{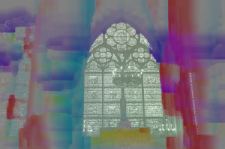}}\hfill
	\subfigure[]
	{\includegraphics[width=0.125\linewidth]{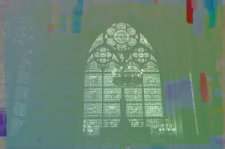}}\hfill
	\subfigure[]
	{\includegraphics[width=0.125\linewidth]{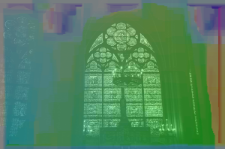}}\hfill
	\subfigure[]
	{\includegraphics[width=0.125\linewidth]{Figure/Figure22/Different_Exposure/f7_LSS_flow.png}}\hfill
	\subfigure[]
	{\includegraphics[width=0.125\linewidth]{Figure/Figure22/Different_Exposure/f7_DASC_flow.png}}\hfill
	\vspace{-22pt}
	\subfigure[]
	{\includegraphics[width=0.125\linewidth]{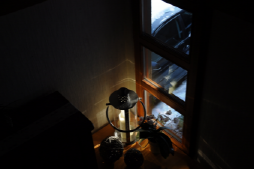}}\hfill
	\subfigure[]
	{\includegraphics[width=0.125\linewidth]{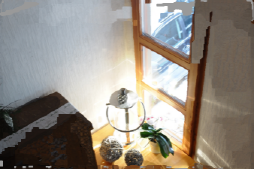}}\hfill
	\subfigure[]
	{\includegraphics[width=0.125\linewidth]{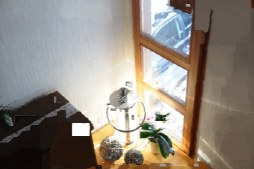}}\hfill
	\subfigure[]
	{\includegraphics[width=0.125\linewidth]{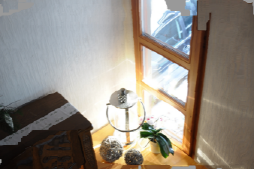}}\hfill
	\subfigure[]
	{\includegraphics[width=0.125\linewidth]{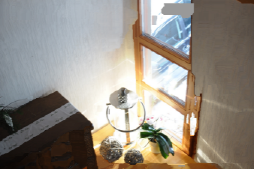}}\hfill
	\subfigure[]
	{\includegraphics[width=0.125\linewidth]{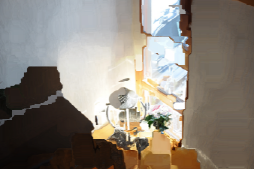}}\hfill
	\subfigure[]
	{\includegraphics[width=0.125\linewidth]{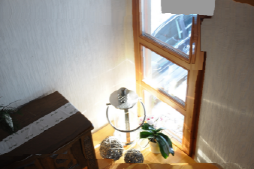}}\hfill
	\subfigure[]
	{\includegraphics[width=0.125\linewidth]{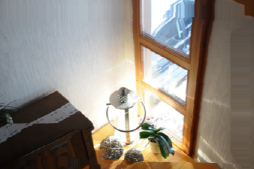}}\hfill
	\vspace{-22pt}
	\subfigure[]
	{\includegraphics[width=0.125\linewidth]{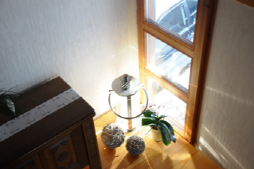}}\hfill
	\subfigure[]
	{\includegraphics[width=0.125\linewidth]{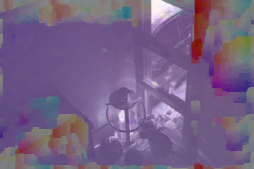}}\hfill
	\subfigure[]
	{\includegraphics[width=0.125\linewidth]{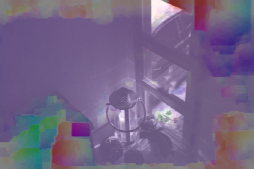}}\hfill
	\subfigure[]
	{\includegraphics[width=0.125\linewidth]{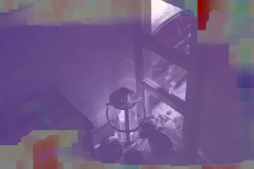}}\hfill
	\subfigure[]
	{\includegraphics[width=0.125\linewidth]{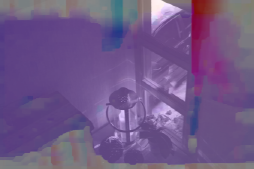}}\hfill
	\subfigure[]
	{\includegraphics[width=0.125\linewidth]{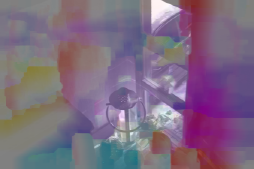}}\hfill
	\subfigure[]
	{\includegraphics[width=0.125\linewidth]{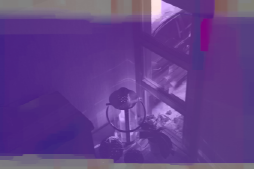}}\hfill
	\subfigure[]
	{\includegraphics[width=0.125\linewidth]{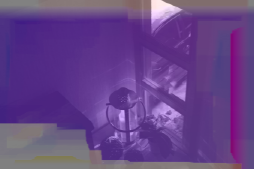}}\hfill
	\vspace{-22pt}
	
	\subfigure[]
	{\includegraphics[width=0.125\linewidth]{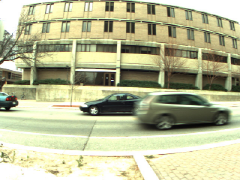}}\hfill
	\subfigure[]
	{\includegraphics[width=0.125\linewidth]{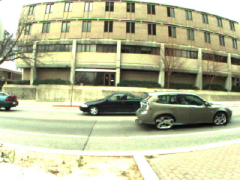}}\hfill
	\subfigure[]
	{\includegraphics[width=0.125\linewidth]{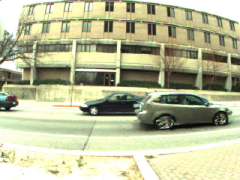}}\hfill
	\subfigure[]
	{\includegraphics[width=0.125\linewidth]{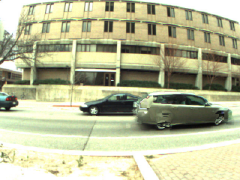}}\hfill
	\subfigure[]
	{\includegraphics[width=0.125\linewidth]{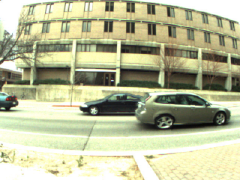}}\hfill
	\subfigure[]
	{\includegraphics[width=0.125\linewidth]{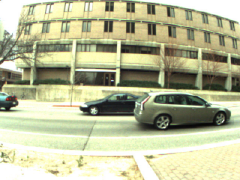}}\hfill
	\subfigure[]
	{\includegraphics[width=0.125\linewidth]{Figure/Figure22/Blur_Sharp/f7_LSS.png}}\hfill
	\subfigure[]
	{\includegraphics[width=0.125\linewidth]{Figure/Figure22/Blur_Sharp/f7_DASC.png}}\hfill
	\vspace{-22pt}
	\subfigure[]
	{\includegraphics[width=0.125\linewidth]{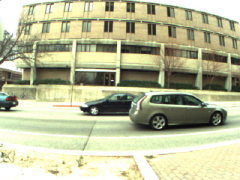}}\hfill
	\subfigure[]
	{\includegraphics[width=0.125\linewidth]{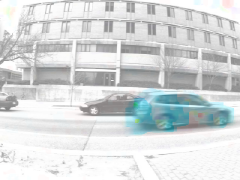}}\hfill
	\subfigure[]
	{\includegraphics[width=0.125\linewidth]{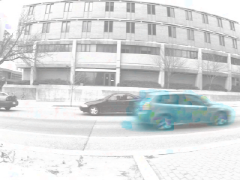}}\hfill
	\subfigure[]
	{\includegraphics[width=0.125\linewidth]{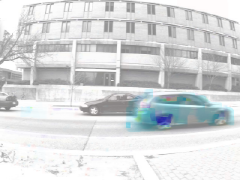}}\hfill
	\subfigure[]
	{\includegraphics[width=0.125\linewidth]{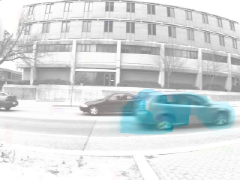}}\hfill
	\subfigure[]
	{\includegraphics[width=0.125\linewidth]{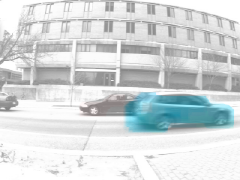}}\hfill
	\subfigure[]
	{\includegraphics[width=0.125\linewidth]{Figure/Figure22/Blur_Sharp/f7_LSS_flow.png}}\hfill
	\subfigure[]
	{\includegraphics[width=0.125\linewidth]{Figure/Figure22/Blur_Sharp/f7_DASC_flow.png}}\hfill
	\vspace{-22pt}
	\subfigure[]
	{\includegraphics[width=0.125\linewidth]{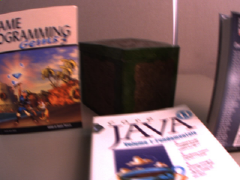}}\hfill
	\subfigure[]
	{\includegraphics[width=0.125\linewidth]{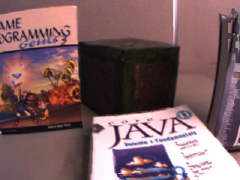}}\hfill
	\subfigure[]
	{\includegraphics[width=0.125\linewidth]{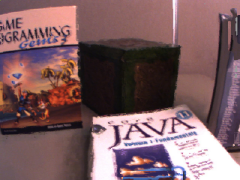}}\hfill
	\subfigure[]
	{\includegraphics[width=0.125\linewidth]{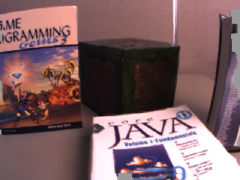}}\hfill
	\subfigure[]
	{\includegraphics[width=0.125\linewidth]{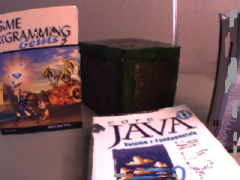}}\hfill
	\subfigure[]
	{\includegraphics[width=0.125\linewidth]{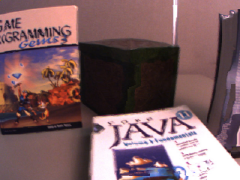}}\hfill
	\subfigure[]
	{\includegraphics[width=0.125\linewidth]{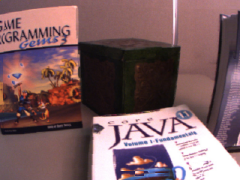}}\hfill
	\subfigure[]
	{\includegraphics[width=0.125\linewidth]{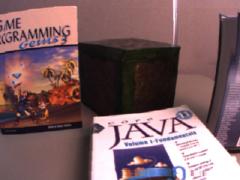}}\hfill
	\vspace{-22pt}
	\subfigure[(a) Image pairs]
	{\includegraphics[width=0.125\linewidth]{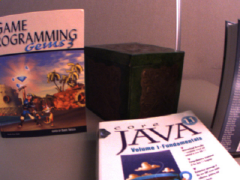}}\hfill
	\subfigure[(b) RSNCC \cite{Shen14}]
	{\includegraphics[width=0.125\linewidth]{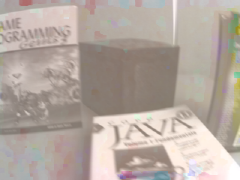}}\hfill
	\subfigure[(c) BRIEF \cite{Calonder11}]
	{\includegraphics[width=0.125\linewidth]{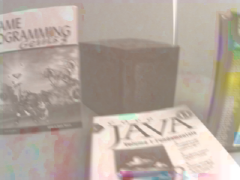}}\hfill
	\subfigure[(d) DAISY \cite{Tola10}]
	{\includegraphics[width=0.125\linewidth]{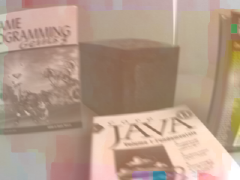}}\hfill
	\subfigure[(e) SIFT \cite{Lowe04}]
	{\includegraphics[width=0.125\linewidth]{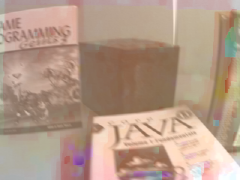}}\hfill
	\subfigure[(f) LSS \cite{Schechtman07}]
	{\includegraphics[width=0.125\linewidth]{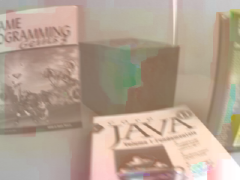}}\hfill
	\subfigure[(g) DASC+RP]
	{\includegraphics[width=0.125\linewidth]{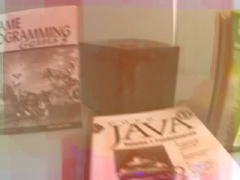}}\hfill
	\subfigure[(h) DASC+LRP]
	{\includegraphics[width=0.125\linewidth]{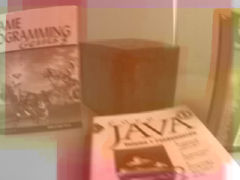}}\hfill
	\vspace{-10pt}
	\caption{Comparison of dense correspondence for
		(from top to bottom) different exposure images and
		blurred-sharpen images. The results consist of warped color images
		and correspondence flow fields overlaid with reference images. Compared to other conventional approaches, our DASC+LRP descriptor estimates reliable dense
		correspondence fields for challenging multi-modal and multi-spectral
		image pairs.}\label{img:22}\vspace{-10pt}
\end{figure*}
\subsubsection{WMSD feature detector analysis}\label{sec:624}
In \figref{img:12} and \figref{img:13}, we analyzed the feature
detection performance of the WMSD detector with a repeatability \cite{Rosten10} and
recognition rate measure \cite{Calonder11} in Mikolajczyk
dataset \cite{mikolajczyk05}. Compared to conventional feature detection approaches
\cite{Lowe04,Matas02,Rosten10,Tombari14}, the WMSD detector extracts
reliable and distinctive points with a high repeatability thanks to
its robustness for modality variations including blur artifacts and illumination changes. 
Furthermore, compared to conventional
gradient-based \cite{Lowe04,Bay06} or intensity-based rotation
estimations \cite{Leutenegger11,Rublee11}, our WMSD-based rotation
estimation combined with the DASC descriptor shows the best performance with
a high recognition rate. \vspace{-5pt}
\subsubsection{Symmetric and asymmetric measure analysis}\label{sec:441}
As shown in \figref{img:8}, a performance gap between using the
asymmetric measure $\tilde{\Psi} (i,j)$ in (9) and 
the symmetric measure $\Psi (i,j)$ in (7) is negligible, while using the
asymmetric measure is much faster. \vspace{-5pt}
\subsubsection{Runtime analysis}\label{sec:625} 
In \tabref{tab:1}, we compared the computational speed of DASC
descriptor with state-of-the-art local descriptors, SIFT
\cite{Lowe04}, DAISY \cite{Tola10}, and LSS \cite{Schechtman07}. The
DASC provides state-of-the-art computational speed. It should be
noted that through recent more efficient edge-aware filters
\cite{He15}, the runtime of DASC can be further reduced.
\vspace{-5pt}
\subsection{Middlebury Stereo Benchmark}\label{sec:63}
We evaluated our DASC+LRP descriptor compared to other approaches in
Middlebury stereo benchmark containing illumination and exposure
variations \cite{middlebury}. In experiments, the illumination (or
exposure) combination `1/3' indicates that two images were captured
under $1^{st}$ and $3^{rd}$ illumination (exposure) conditions,
respectively \cite{middlebury}. \figref{img:19} shows average bad
matching errors in un-occluded areas of depth maps obtained under
illumination or exposure variations with the graph-cut (GC)
\cite{Boykov01} and winner-takes-all (WTA) optimization.
\figref{img:20} shows disparity maps for severe illumination
variations obtained by varying cost functions with the WTA
optimization. Our DASC+LRP descriptor achieves the best results both
quantitatively and qualitatively. Area-based approaches,
\emph{e.g.}, MI+SIFT \cite{Heo13}, ANCC \cite{Heo}, and RSNCC
\cite{Shen14}, are very sensitive to severe radiometric variations,
especially when local variations frequently occur. Contrarily,
descriptor-based approaches perform better than the area-based
approaches. Interestingly, the BRIEF \cite{Calonder11} is better
than other gradient-based descriptors (SIFT \cite{Lowe04} and DAISY
\cite{Tola10}) thanks to an ordering robustness. \vspace{-5pt}
\subsection{Multi-modal and Multi-spectral Benchmark}\label{sec:64}
Next, we evaluated our DASC+LRP descriptor with images under
modality variations, \emph{e.g.}, RGB-NIR \cite{Shen14,Brown11},
different exposure \cite{Shen14,Sen12}, flash-noflash \cite{Sen12},
and blurred artifacts \cite{HaCohen13,Lee13}. As varying
descriptors and similarity measures, we use the WTA and SIFT flow
optimization using the hierarchical dual-layer belief propagation
(BP) \cite{Liu11}, whose code is publicly available. Unlike the
Middlebury stereo benchmark, these datasets have no ground truth
correspondence maps, and thus we manually obtained ground truth
displacement vectors for $100$ corner points for all images, and used
them for an objective evaluation similar to \cite{Shen14}.

Area-based approaches, \emph{e.g.}, MI+SIFT \cite{Heo13}, ANCC \cite{Heo}, and RSNCC
\cite{Shen14}, are very sensitive to
local variations. As already described in literatures \cite{Shen14},
gradient-based approaches, \emph{e.g.}, SIFT \cite{Lowe04} and DAISY
\cite{Tola10}, have shown limited performance in RGB-NIR pairs where
the gradient reversal and inversion frequently appear. The BRIEF
\cite{Calonder11} cannot deal with noisy and modality varying
regions since it considers a pixel difference only. It should be noted
that some efforts have been made to estimate reliable flow maps in the motion blur, \emph{e.g.}, blur-flow \cite{Portz12},
but they typically employ an iterative matching framework, which relies heavily on an initial estimate.
Additionally, they do not scale well to general purpose matching scenarios.
Unlike these approaches, the LSS \cite{Schechtman07} and our descriptor consider
the local self-similarities, but the LSS still lacks a
discriminative power for dense matching. Our DASC+RP descriptor
leveraging patch-wise pooling with adaptive self-correlation provides
satisfactory results under modality variations. By employing the
optimal sampling pattern via discriminative learning (DASC+LRP), the matching accuracy was
further improved. \figref{img:21} and \figref{img:22} show qualitative evaluation,
clearly demonstrating the outstanding performance of our descriptor.
\tabref{tab:2} shows an objective evaluation of DASC+LRP descriptor
and other state-of-the-art methods on these datasets.
\vspace{-5pt}
\begin{table*}[t]
	\caption{Comparison of quantitative evaluation on multi-spectral and
		multi-modal images.}\label{tab:2}\vspace{-10pt}
	\centering
	\begin{tabular}{ >{\raggedright}m{0.1\linewidth}
			>{\centering}m{0.05\linewidth}  >{\centering}m{0.05\linewidth}
			>{\centering}m{0.05\linewidth}  >{\centering}m{0.05\linewidth}  >{\centering}m{0.05\linewidth}
			>{\centering}m{0.05\linewidth}  >{\centering}m{0.05\linewidth}
			>{\centering}m{0.05\linewidth}  >{\centering}m{0.05\linewidth}  >{\centering}m{0.05\linewidth} }
		\hlinewd{0.8pt}
		\multirow{2}{*}{} & \multicolumn{5}{ c }{WTA optimization} & \multicolumn{5}{ c }{SF optimization \cite{Liu11}} \tabularnewline
		\cline{2-11}
		&RGB-NIR &{\footnotesize flash-noflash} &{\footnotesize diff. expo.} &blur-sharp &Average
		&RGB-NIR &{\footnotesize flash-noflash} &{\footnotesize diff. expo.} &blur-sharp &Average \tabularnewline
		\hline
		\hline
		MI+SIFT \cite{Heo13} &25.13 &27.12 &28.23 &24.21 &27.12 &17.21 &13.24 &14.16 &20.14 &16.87 \tabularnewline
		ANCC \cite{Heo} &23.21 &20.42 &25.19 &26.14 &23.74 &18.45 &14.14 &11.96 &19.24 &15.94 \tabularnewline
		RSNCC \cite{Shen14} &27.51 &25.12 &18.21 &27.91 &24.68 &13.41 &15.87 &9.15 &18.21 &14.16 \tabularnewline
		SIFT \cite{Lowe04} &24.11 &18.72 &19.42 &27.18 &22.36 &18.51 &11.06 &14.87 &20.78 &16.35 \tabularnewline
		DAISY \cite{Tola10} &27.61 &26.30 &20.72 &27.41 &25.51 &20.42 &10.84 &12.71 &22.91 &16.72 \tabularnewline
		BRIEF \cite{Calonder11} &29.14 &18.29\cellcolor[gray]{0.9} &17.13\cellcolor[gray]{0.9} &26.43 &22.75\cellcolor[gray]{0.9} &17.54 &9.21\cellcolor[gray]{0.9} &9.54 &19.72 &14.05 \tabularnewline
		LSS \cite{Schechtman07} &27.82\cellcolor[gray]{0.9} &19.18 &18.21 &26.14\cellcolor[gray]{0.9} &22.84 &16.14\cellcolor[gray]{0.9} &11.88 &9.11\cellcolor[gray]{0.9} &18.51\cellcolor[gray]{0.9} &13.91\cellcolor[gray]{0.9} \tabularnewline
		\hline
		\textbf{DASC+RP} &18.21\cellcolor[gray]{0.7} &14.28\cellcolor[gray]{0.7} &12.12\cellcolor[gray]{0.7} &17.11\cellcolor[gray]{0.7} &12.18\cellcolor[gray]{0.7} &15.43\cellcolor[gray]{0.7} &7.51\cellcolor[gray]{0.7} &7.32\cellcolor[gray]{0.7} &12.21\cellcolor[gray]{0.7} &9.68\cellcolor[gray]{0.7} \tabularnewline
		\textbf{DASC+LRP} &\textbf{13.42}\cellcolor[gray]{0.5} &\textbf{11.28}\cellcolor[gray]{0.5} &\textbf{9.23}\cellcolor[gray]{0.5} &\textbf{13.28}\cellcolor[gray]{0.5} &\textbf{11.80}\cellcolor[gray]{0.5} &\textbf{8.10}\cellcolor[gray]{0.5} &\textbf{5.41}\cellcolor[gray]{0.5} &\textbf{6.24}\cellcolor[gray]{0.5} &\textbf{10.81}\cellcolor[gray]{0.5} &\textbf{7.64}\cellcolor[gray]{0.5} \tabularnewline
		\hlinewd{0.8pt}
	\end{tabular}\vspace{-15pt}
\end{table*}
\subsection{DIML Multi-modal Benchmark}\label{sec:65}
\begin{figure}[t]
	\centering
	\renewcommand{\thesubfigure}{}
	\subfigure[]
	{\includegraphics[width=0.2\linewidth]{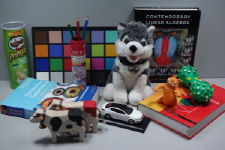}}\hfill
	\subfigure[]
	{\includegraphics[width=0.2\linewidth]{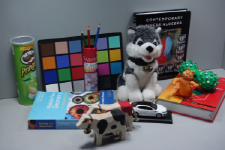}}\hfill
	\subfigure[]
	{\includegraphics[width=0.2\linewidth]{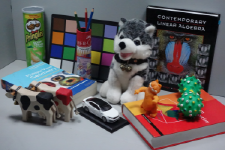}}\hfill
	\subfigure[]
	{\includegraphics[width=0.2\linewidth]{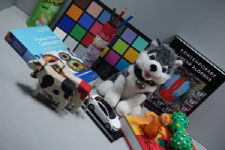}}\hfill
	\subfigure[]
	{\includegraphics[width=0.2\linewidth]{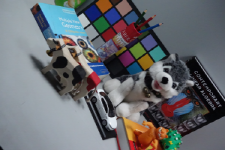}}\hfill
	\vspace{-22pt}
	\subfigure[]
	{\includegraphics[width=0.2\linewidth]{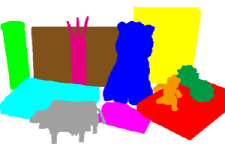}}\hfill
	\subfigure[]
	{\includegraphics[width=0.2\linewidth]{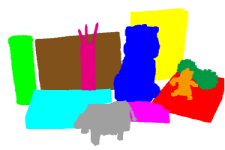}}\hfill
	\subfigure[]
	{\includegraphics[width=0.2\linewidth]{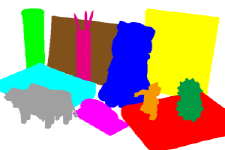}}\hfill
	\subfigure[]
	{\includegraphics[width=0.2\linewidth]{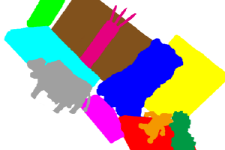}}\hfill
	\subfigure[]
	{\includegraphics[width=0.2\linewidth]{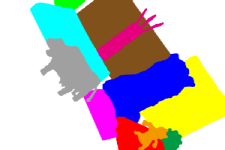}}\hfill
	\vspace{-22pt}
	\subfigure[]
	{\includegraphics[width=0.2\linewidth]{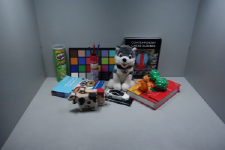}}\hfill
	\subfigure[]
	{\includegraphics[width=0.2\linewidth]{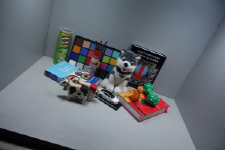}}\hfill
	\subfigure[]
	{\includegraphics[width=0.2\linewidth]{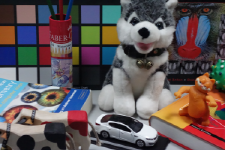}}\hfill
	\subfigure[]
	{\includegraphics[width=0.2\linewidth]{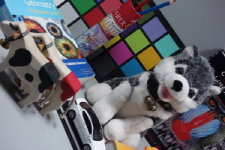}}\hfill
	\subfigure[]
	{\includegraphics[width=0.2\linewidth]{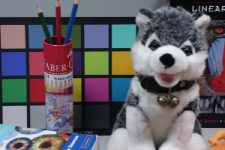}}\hfill
	\vspace{-22pt}
	\subfigure[]
	{\includegraphics[width=0.2\linewidth]{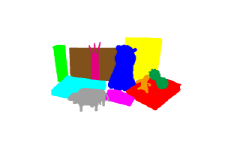}}\hfill
	\subfigure[]
	{\includegraphics[width=0.2\linewidth]{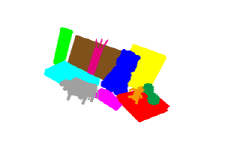}}\hfill
	\subfigure[]
	{\includegraphics[width=0.2\linewidth]{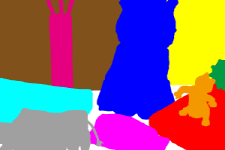}}\hfill
	\subfigure[]
	{\includegraphics[width=0.2\linewidth]{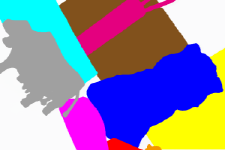}}\hfill
	\subfigure[]
	{\includegraphics[width=0.2\linewidth]{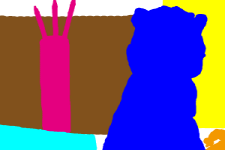}}\hfill
	\vspace{-20pt}
	\caption{Examples of DIML multi-modal benchmark.
		It consists of images taken under $10$ different geometric conditions such as viewpoint, scale, rotation, and scale-rotation
		with ground truth annotation.}\label{img:23}\vspace{-10pt}
\end{figure}

\begin{figure}[t]
	\centering
	\renewcommand{\thesubfigure}{}
	\subfigure[]
	{\includegraphics[width=0.2\linewidth]{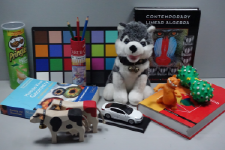}}\hfill
	\subfigure[]
	{\includegraphics[width=0.2\linewidth]{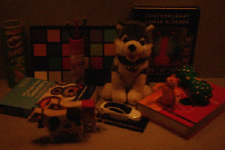}}\hfill
	\subfigure[]
	{\includegraphics[width=0.2\linewidth]{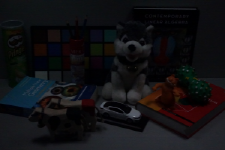}}\hfill
	\subfigure[]
	{\includegraphics[width=0.2\linewidth]{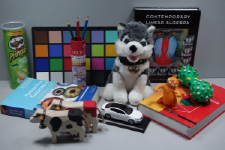}}\hfill
	\subfigure[]
	{\includegraphics[width=0.2\linewidth]{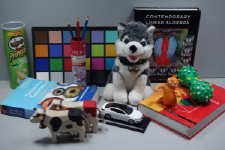}}\hfill
	\vspace{-22pt}
	\subfigure[]
	{\includegraphics[width=0.2\linewidth]{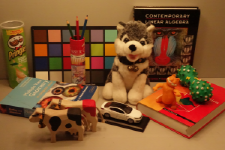}}\hfill
	\subfigure[]
	{\includegraphics[width=0.2\linewidth]{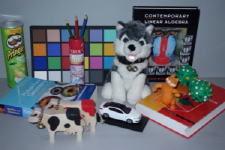}}\hfill
	\subfigure[]
	{\includegraphics[width=0.2\linewidth]{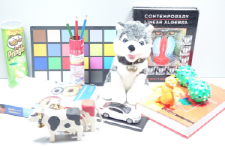}}\hfill
	\subfigure[]
	{\includegraphics[width=0.2\linewidth]{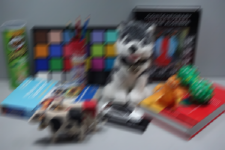}}\hfill
	\subfigure[]
	{\includegraphics[width=0.2\linewidth]{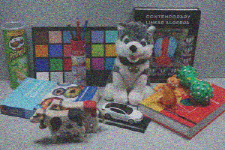}}\hfill
	\vspace{-20pt}
	\caption{Examples of DIML multi-modal benchmark. Each geometry image sets in \figref{img:23}
		consists of $5$ different photometric variations such as illumination, exposure, flash-noflash, blur, and noise.}\label{img:24}\vspace{-10pt}
\end{figure}

Since there have been no database with both photometric and
geometric variations, we built the DIML multi-modal benchmark \cite{dasc}. 
All databases were taken by SONY Cyber-Shot DSC-RX100
camera in a darkroom with the lighting booth GretagMacbeth
SpectraLight III. In terms of geometric deformations, we captured 10
geometry image sets by combining geometric variations of viewpoint,
scale, and rotation as shown in \figref{img:23}, and each image set
consists of images taken under $5$ different photometric variation
pairs including illumination, exposure, flash-noflash, blur, and
noise as shown in \figref{img:24}. Therefore, the DIML multi-modal
benchmark consists of $100$ images with the size of $1200 \times
800$. Furthermore, by following \cite{Liu11}, we manually built
ground truth object annotation maps to evaluate the performance
quantitatively, and computed the label transfer accuracy (LTA)
$\mathcal{A}^{\mathrm{LTA}}$ such that
\begin{equation}\label{equ:LTA}
\mathcal{A}^{\mathrm{LTA}} = \frac{1}{\mathcal{T}_a}\sum\nolimits_{i \in {\mathcal{I}}} {1({e_i} \neq {a_i},{a_i} > 0)}
\end{equation}
where the ground-truth annotation is ${a_i}$, estimated annotation
is ${e_i}$, and $\mathcal{T}_a = \sum\nolimits_{i \in
{\mathcal{I}}}{1({a_i} > 0)}$ is the number of labeled pixels.
This metric has been widely used in wide-baseline matching tasks
\cite{Kim13}. Though $\mathcal{A}^{\mathrm{LTA}}$ does
not measure a matching performance in a pixel precision, it was
shown in \cite{Liu11} that this metric is an excellent alternative
enough to evaluate the performance
of descriptors in case that there are no ground truth correspondence
maps available.
\begin{figure}[t]
	\centering
	\renewcommand{\thesubfigure}{}
	\subfigure[(a) DAISY \cite{Tola10}]
	{\includegraphics[width=0.5\linewidth]{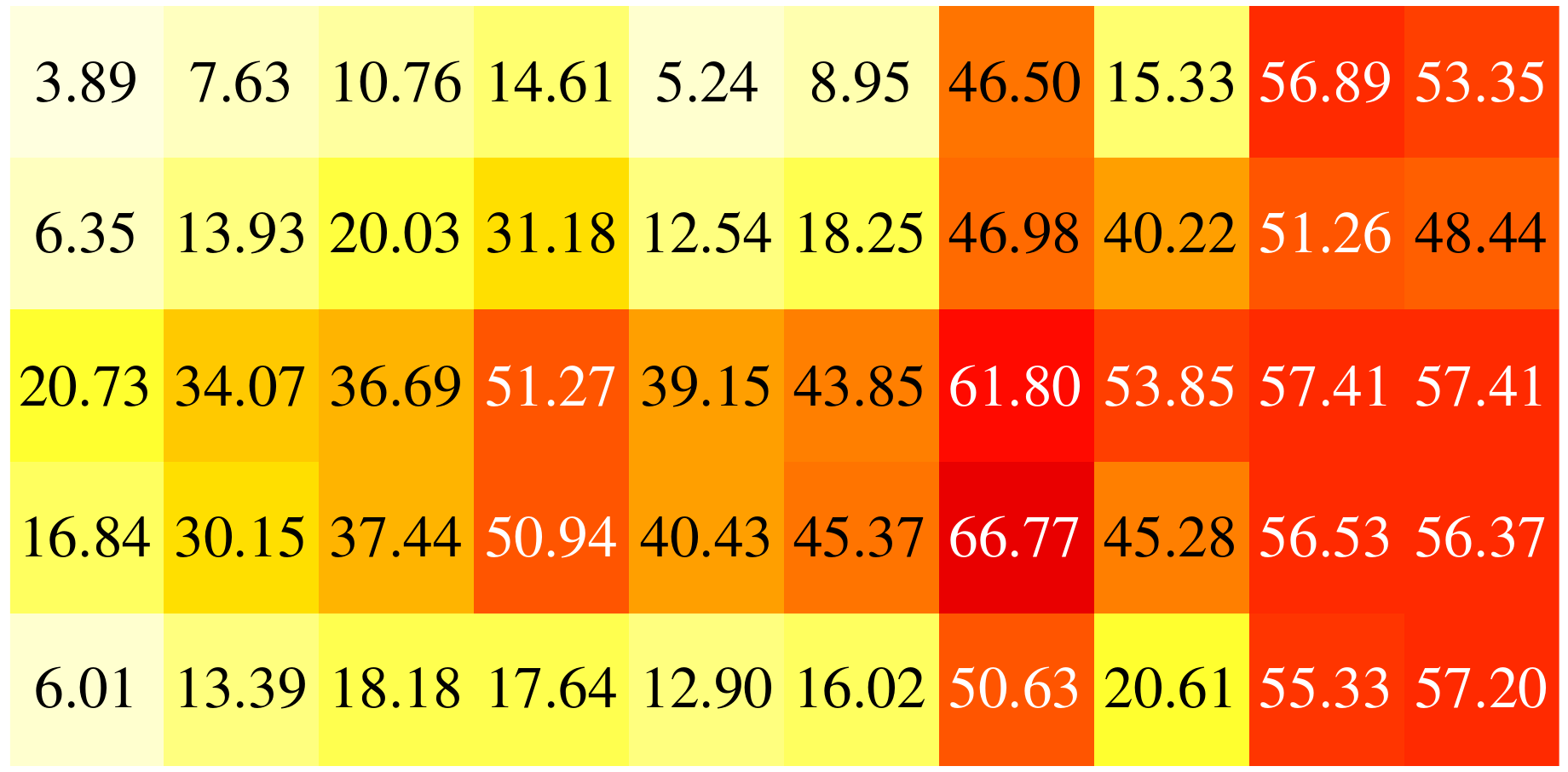}}\hfill
	\subfigure[(b) LSS \cite{Schechtman07}]
	{\includegraphics[width=0.5\linewidth]{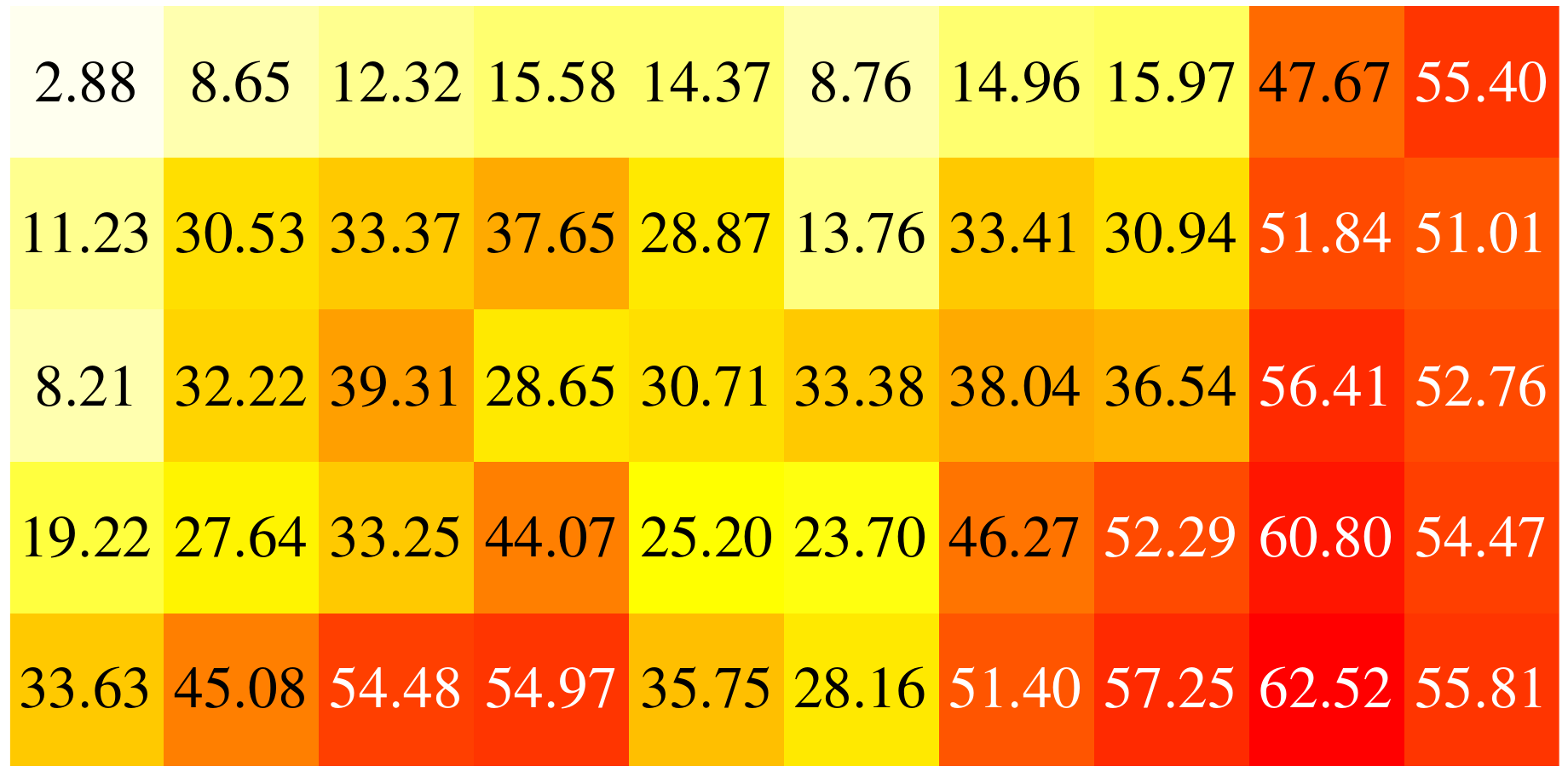}}\hfill
	\vspace{-8pt} 
	\subfigure[(c) SegSIFT \cite{Trulls13}]
	{\includegraphics[width=0.5\linewidth]{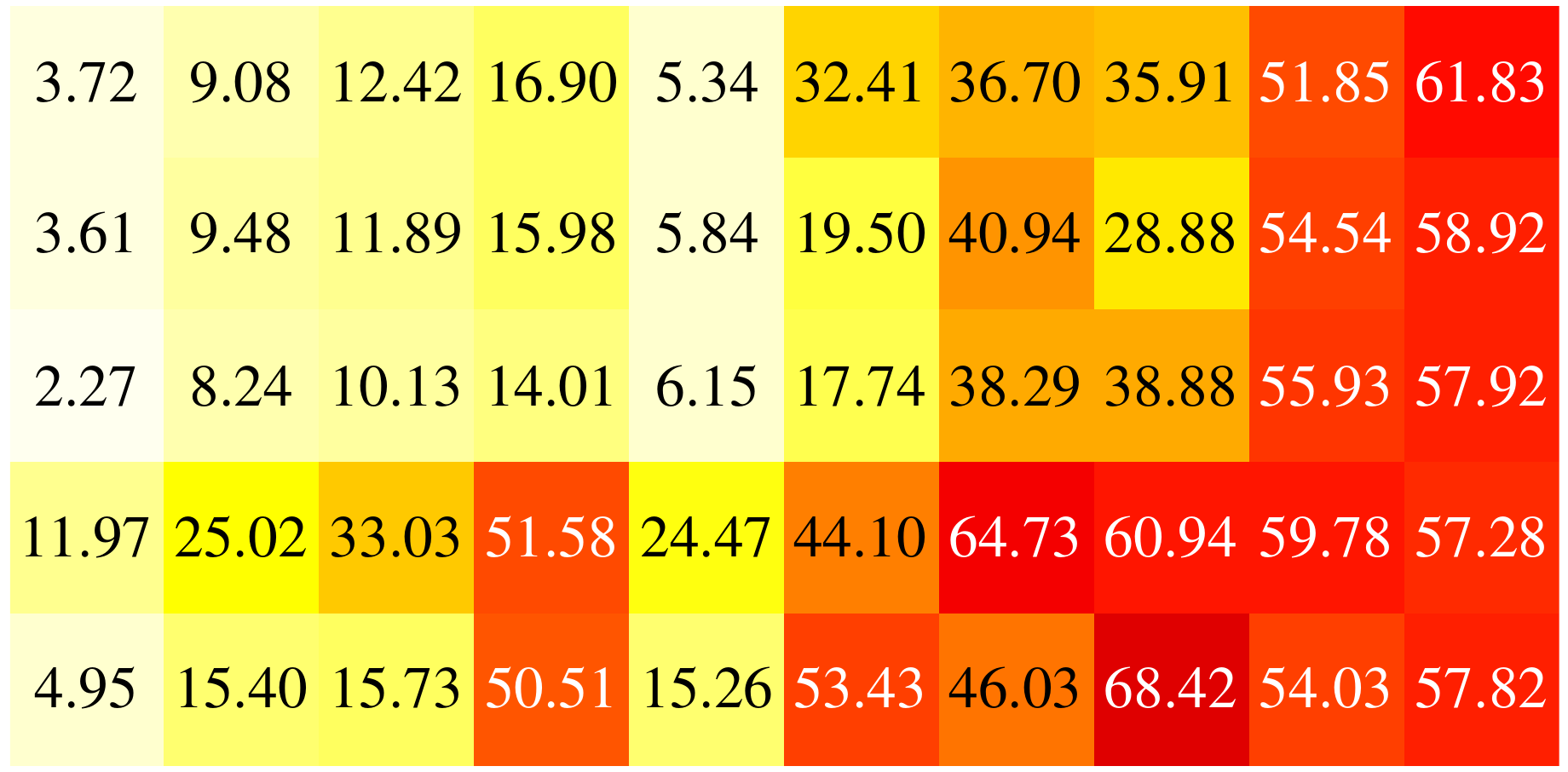}}\hfill
	\subfigure[(d) SegSID \cite{Trulls13}]
	{\includegraphics[width=0.5\linewidth]{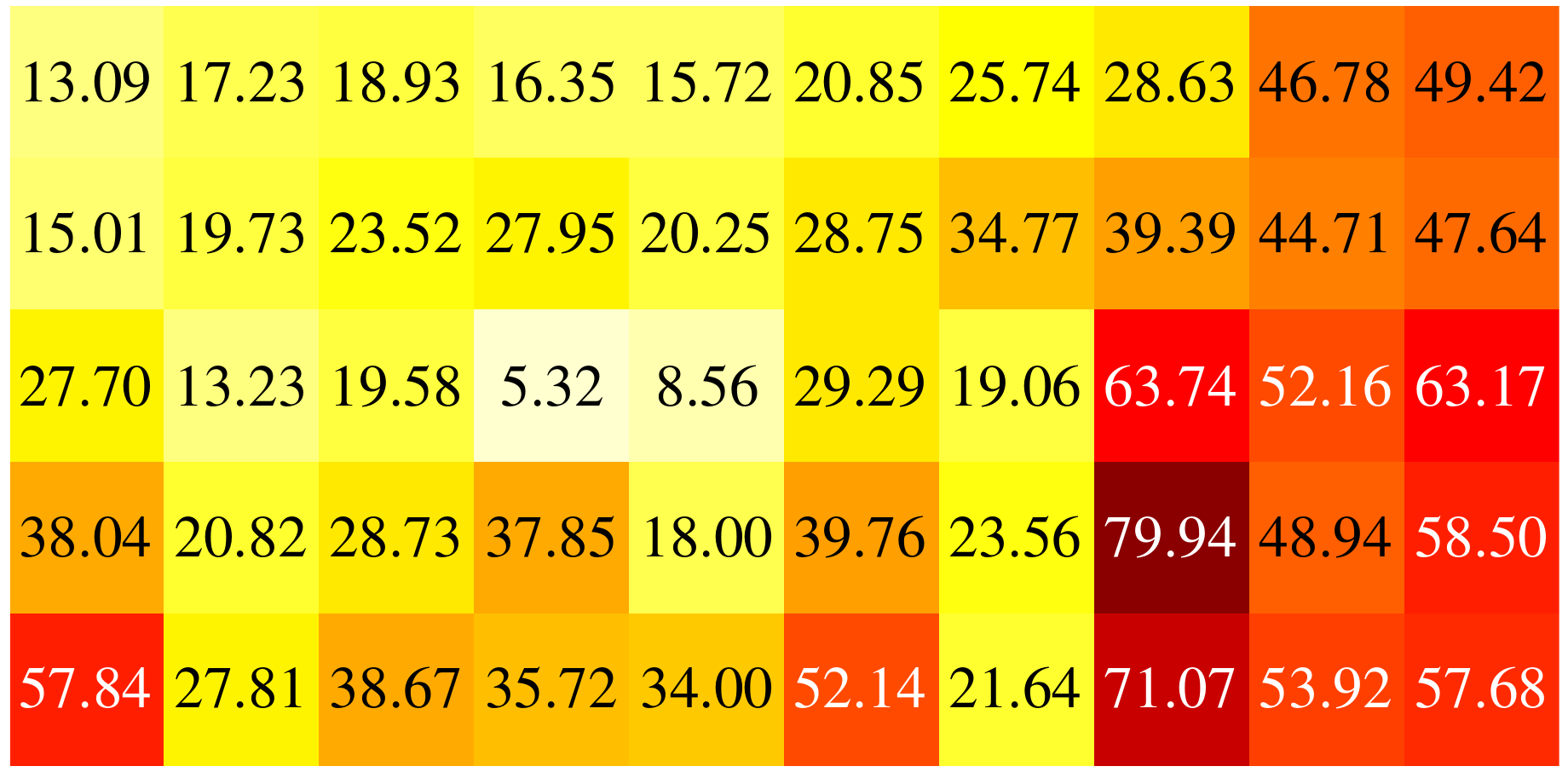}}\hfill
	\vspace{-8pt} 
	\subfigure[(e) DSP \cite{Kim13}]
	{\includegraphics[width=0.5\linewidth]{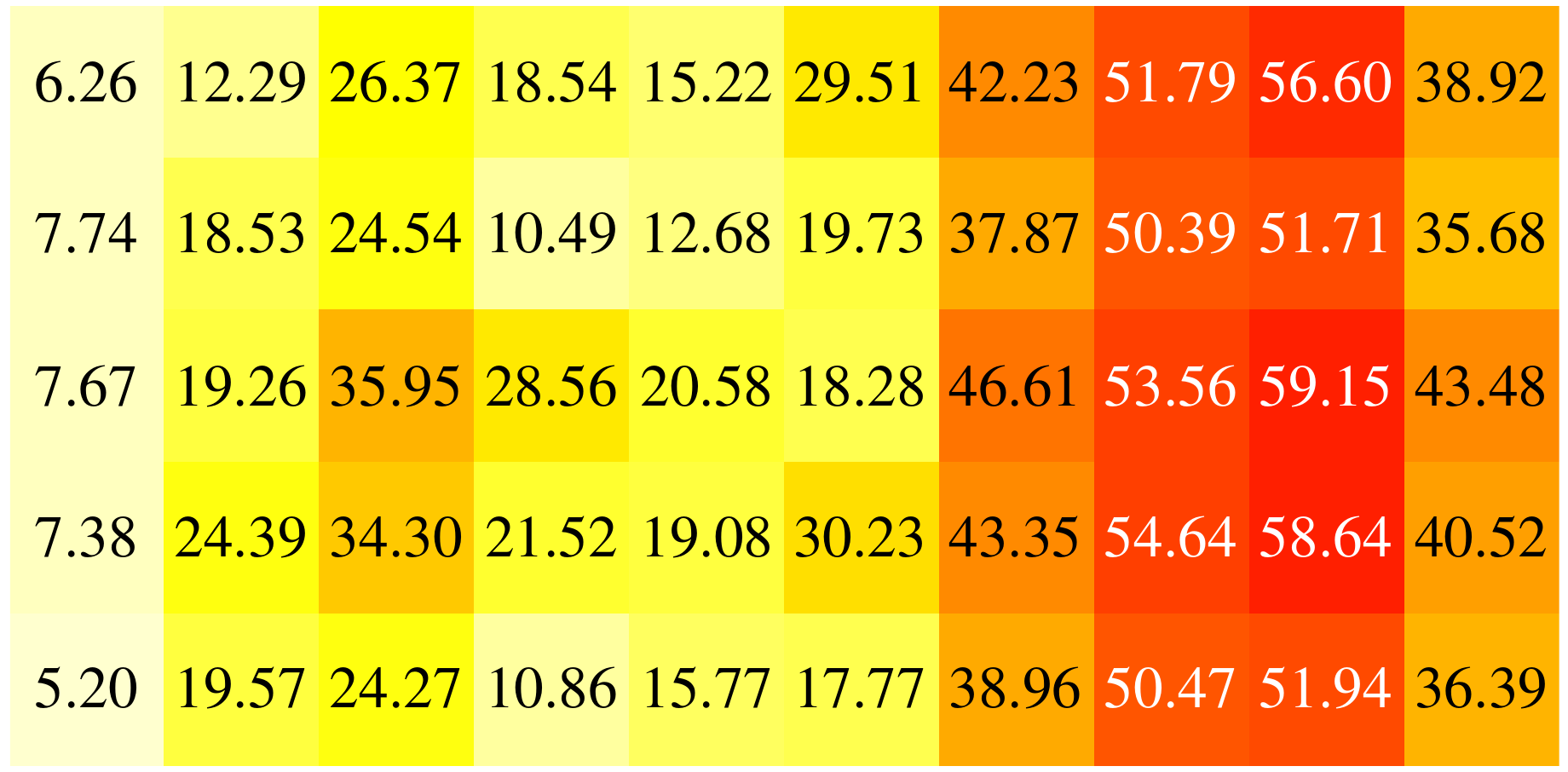}}\hfill
	\subfigure[(f) SSF \cite{Qiu14}]
	{\includegraphics[width=0.5\linewidth]{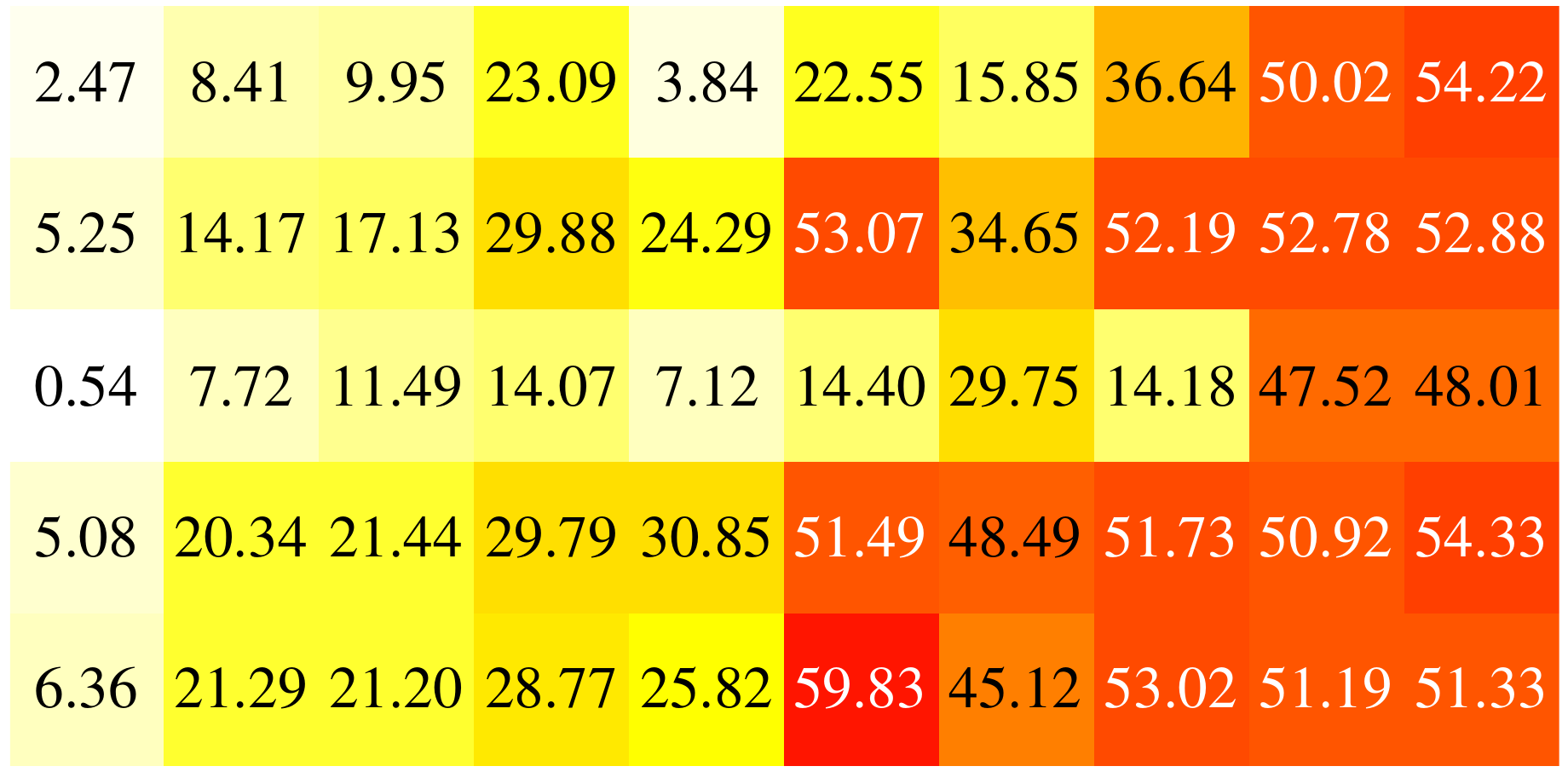}}\hfill
	\vspace{-8pt} 
	\subfigure[(g) DASC]
	{\includegraphics[width=0.5\linewidth]{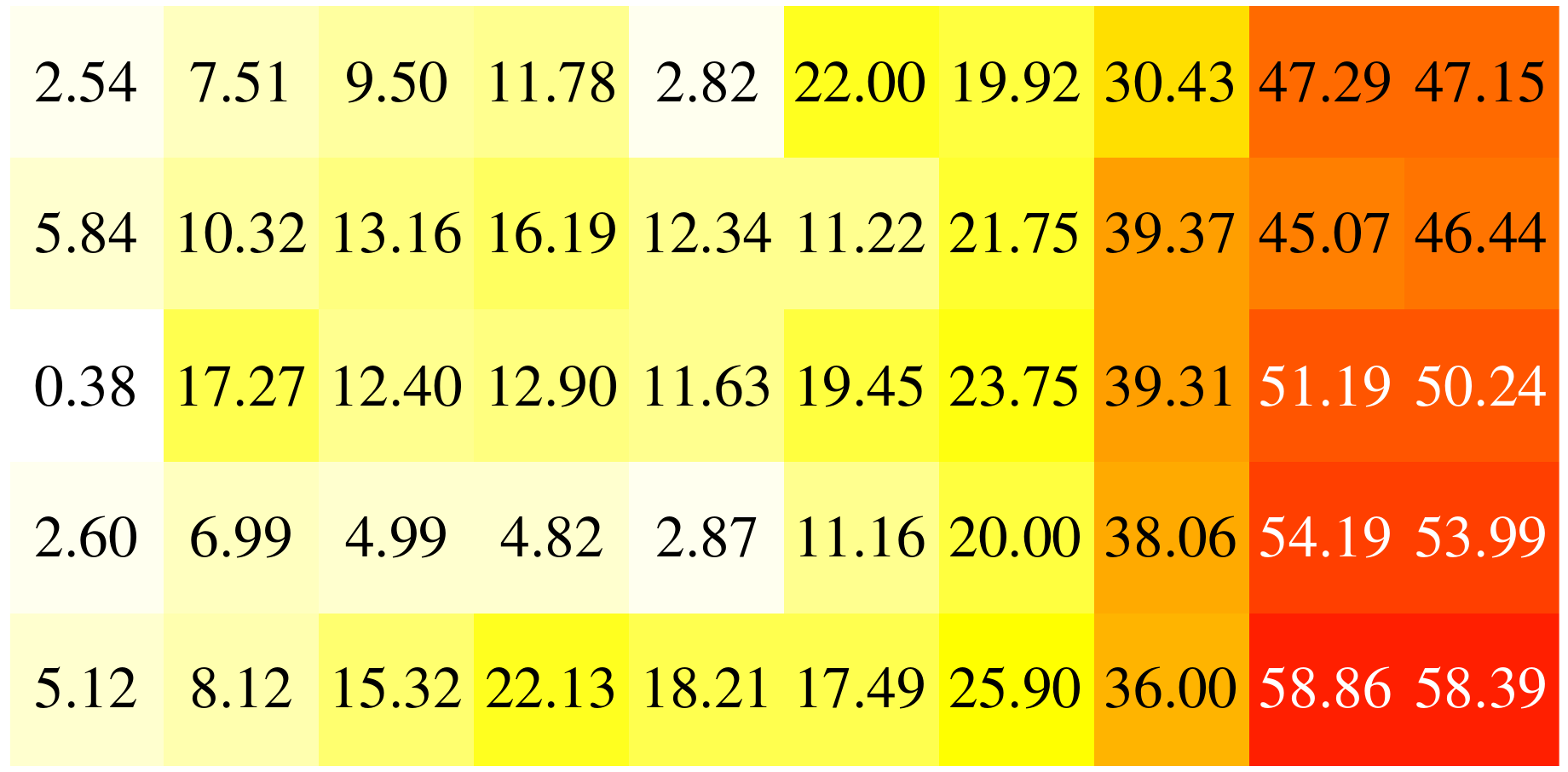}}\hfill
	\subfigure[(h) GI-DASC]
	{\includegraphics[width=0.5\linewidth]{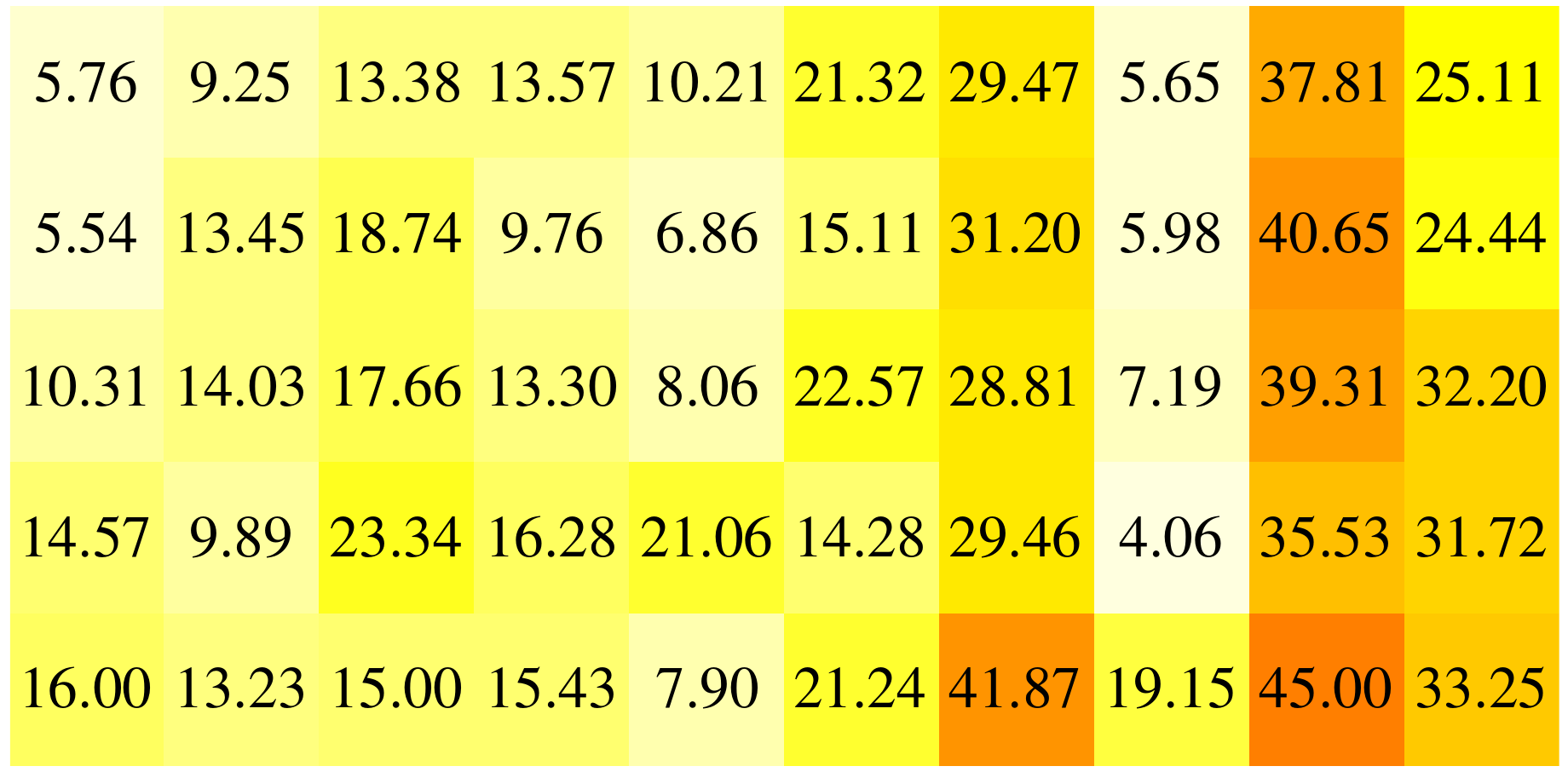}}\hfill
	\vspace{-10pt} \caption{Comparison of quantitative evaluation on
		DIML benchmark \cite{dasc}. Each result represents the
		LTA for geometric (x-axis) and photometric
		(y-axis) variations, respectively. The DASC outperforms
		conventional descriptors such as DAISY
		\cite{Tola10} and LSS \cite{Schechtman07}.
		Interestingly, its accuracy is also higher than those of
		state-of-the-art geometry-invariant methods including SegSIFT \cite{Trulls13}, SegSID \cite{Trulls13}, DSP \cite{Kim13}, and SSF \cite{Qiu14}. The
		GI-DASC shows the best performance under varying
		photometric and geometric conditions.}\label{img:25}\vspace{-10pt}
\end{figure}

For an image from the reference geometry image set (the
first image in \figref{img:23}), we estimated visual correspondence
maps with images from other geometry image set, and then computed the LTA.
Furthermore, visual correspondence maps were estimated for each photometric pair.
Here, matching results at occluded pixels should be excluded in the
evaluation as they have no corresponding pixels. We hence warped an
image taken from near into an image taken at a distance, when
computing the LTA. The experimental setup for DIML multi-modal
benchmark was given in detail at our project page \cite{dasc}.
\begin{figure*}[t]
	\centering
	\renewcommand{\thesubfigure}{}
	\subfigure[]
	{\includegraphics[width=0.125\linewidth]{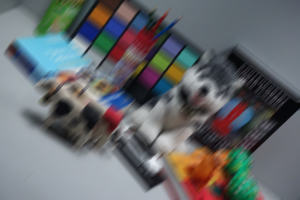}}\hfill
	\subfigure[]
	{\includegraphics[width=0.125\linewidth]{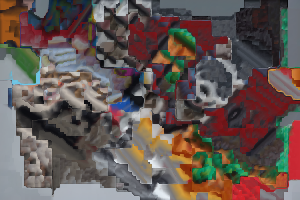}}\hfill
	\subfigure[]
	{\includegraphics[width=0.125\linewidth]{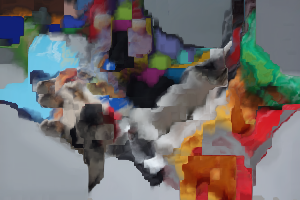}}\hfill
	\subfigure[]
	{\includegraphics[width=0.125\linewidth]{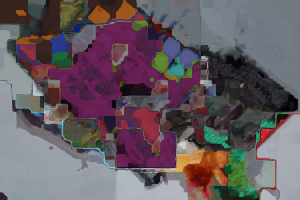}}\hfill
	\subfigure[]
	{\includegraphics[width=0.125\linewidth]{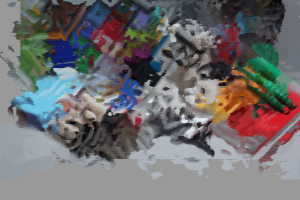}}\hfill
	\subfigure[]
	{\includegraphics[width=0.125\linewidth]{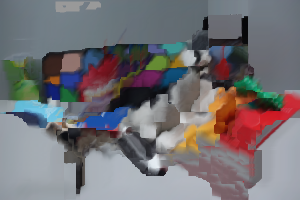}}\hfill
	\subfigure[]
	{\includegraphics[width=0.125\linewidth]{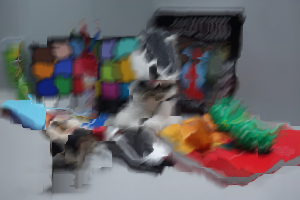}}\hfill
	\subfigure[]
	{\includegraphics[width=0.125\linewidth]{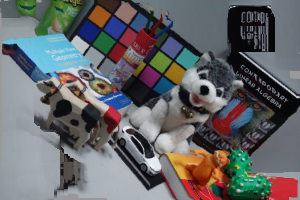}}\hfill
	\vspace{-22pt}
	\subfigure[]
	{\includegraphics[width=0.125\linewidth]{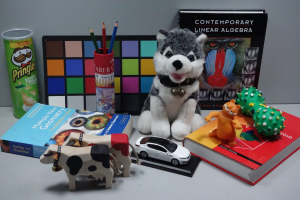}}\hfill
	\subfigure[]
	{\includegraphics[width=0.125\linewidth]{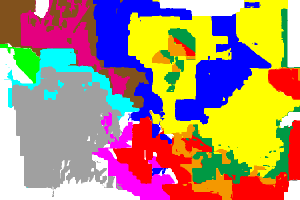}}\hfill
	\subfigure[]
	{\includegraphics[width=0.125\linewidth]{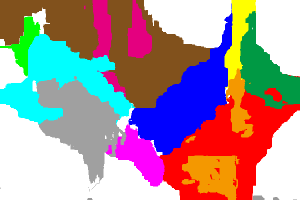}}\hfill
	\subfigure[]
	{\includegraphics[width=0.125\linewidth]{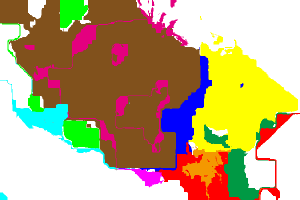}}\hfill
	\subfigure[]
	{\includegraphics[width=0.125\linewidth]{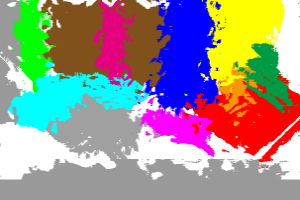}}\hfill
	\subfigure[]
	{\includegraphics[width=0.125\linewidth]{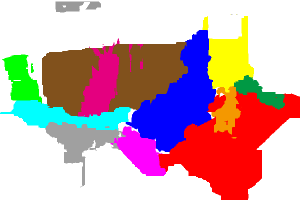}}\hfill
	\subfigure[]
	{\includegraphics[width=0.125\linewidth]{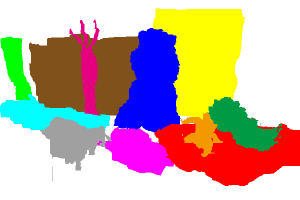}}\hfill
	\subfigure[]
	{\includegraphics[width=0.125\linewidth]{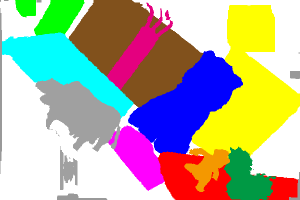}}\hfill
	\vspace{-22pt}
	\subfigure[]
	{\includegraphics[width=0.125\linewidth]{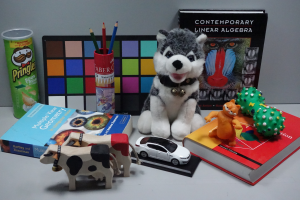}}\hfill
	\subfigure[]
	{\includegraphics[width=0.125\linewidth]{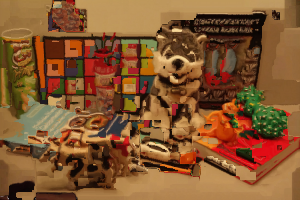}}\hfill
	\subfigure[]
	{\includegraphics[width=0.125\linewidth]{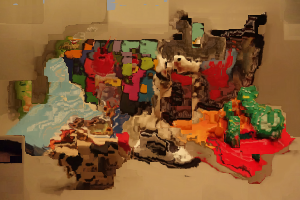}}\hfill
	\subfigure[]
	{\includegraphics[width=0.125\linewidth]{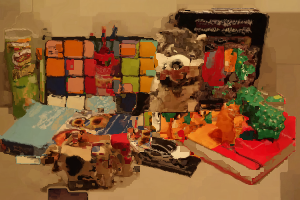}}\hfill
	\subfigure[]
	{\includegraphics[width=0.125\linewidth]{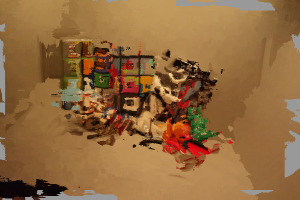}}\hfill
	\subfigure[]
	{\includegraphics[width=0.125\linewidth]{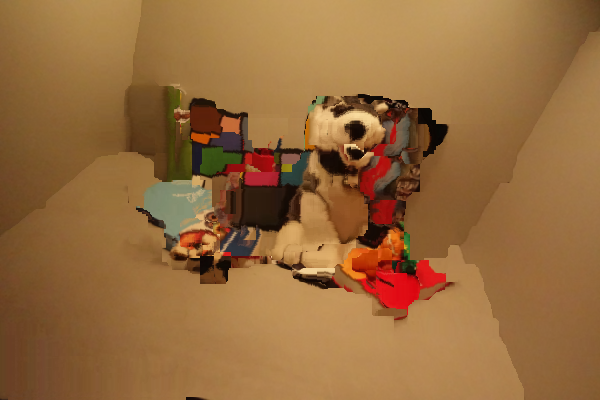}}\hfill
	\subfigure[]
	{\includegraphics[width=0.125\linewidth]{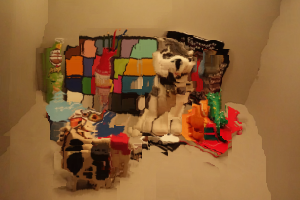}}\hfill
	\subfigure[]
	{\includegraphics[width=0.125\linewidth]{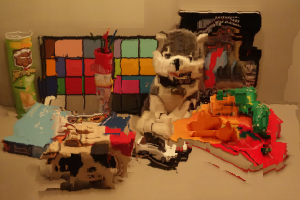}}\hfill
	\vspace{-22pt}
	\subfigure[(a) Image pairs]
	{\includegraphics[width=0.125\linewidth]{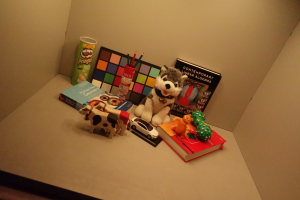}}\hfill
	\subfigure[(b) DAISY \cite{Tola10}]
	{\includegraphics[width=0.125\linewidth]{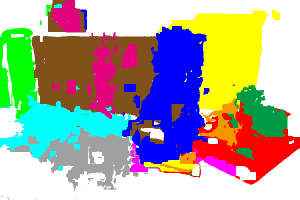}}\hfill
	\subfigure[(c) Seg-SIFT \cite{Trulls13}]
	{\includegraphics[width=0.125\linewidth]{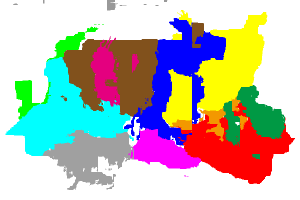}}\hfill
	\subfigure[(d) Seg-SID \cite{Kokkinos08}]
	{\includegraphics[width=0.125\linewidth]{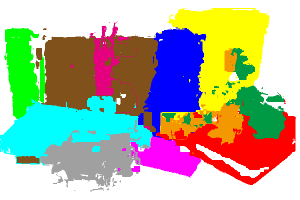}}\hfill
	\subfigure[(e) DSP \cite{Kim13}]
	{\includegraphics[width=0.125\linewidth]{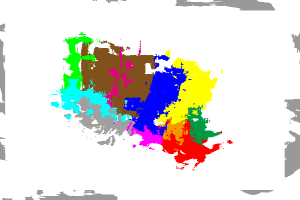}}\hfill
	\subfigure[(f) SSF \cite{Qiu14}]
	{\includegraphics[width=0.125\linewidth]{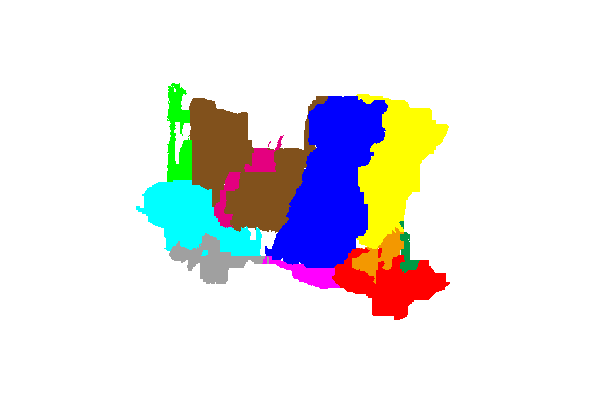}}\hfill
	\subfigure[(g) DASC]
	{\includegraphics[width=0.125\linewidth]{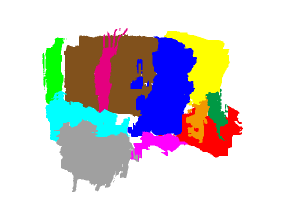}}\hfill
	\subfigure[(h) GI-DASC]
	{\includegraphics[width=0.125\linewidth]{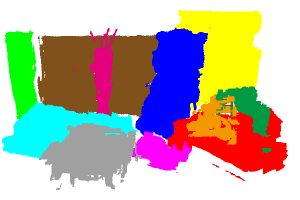}}\hfill
	\vspace{-10pt}
	\caption{Comparison of qualitative evaluation on DIML multi-modal benchmark.
		The results consist of warped color images and warped ground truth annotations.
		Compared to other conventional descriptors and geometry-invariant approaches, our DASC descriptor estimates reliable dense
		correspondence fields for image pairs across varying geometric and photometric conditions.}\label{img:26}\vspace{-12pt}
\end{figure*}
\begin{figure}[t]
	\centering
	\renewcommand{\thesubfigure}{}
	\subfigure[]
	{\includegraphics[width=0.9\linewidth]{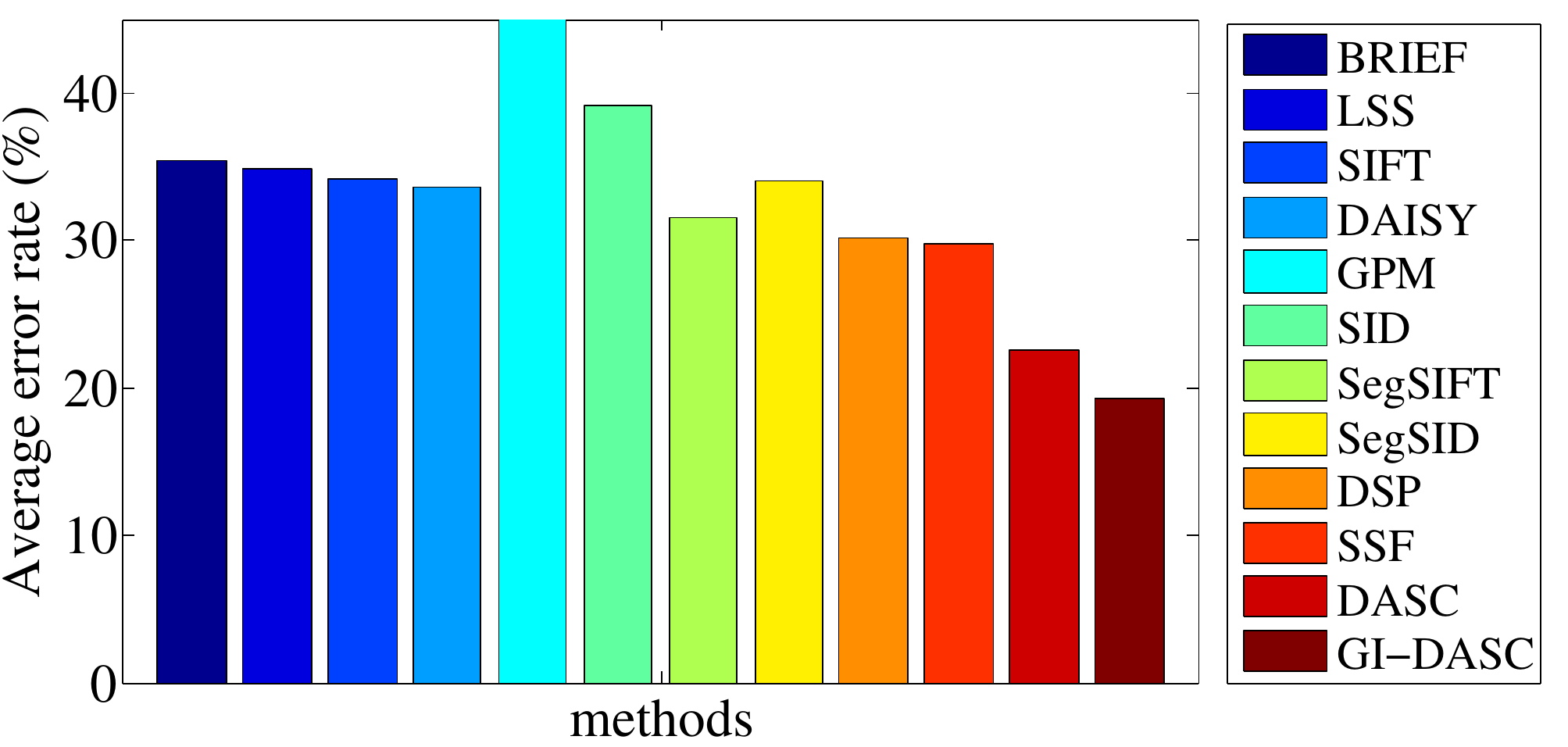}}
	\vspace{-20pt}
	\caption{Average error rates on DIML multi-modal benchmark.}\label{img:27}\vspace{-10pt}
\end{figure}

We compared our two descriptors, DASC and GI-DASC, with conventional
descriptors such as SIFT \cite{Lowe04}, DAISY \cite{Tola10}, BRIEF
\cite{Calonder11}, and LSS \cite{Schechtman07}, and
state-of-the-arts geometry-invariant approaches such as SID
\cite{Kokkinos08}, SegSIFT \cite{Trulls13}, SegSID \cite{Trulls13},
GPM \cite{Barnes10}, DSP \cite{Kim13}, and SSF \cite{Qiu14}. For the
sake of simplicity, we omit `LRP' in the DASC-LRP and GI-DASC-LRP.
\figref{img:25} shows the LTA error rates as varying photometric and
geometric deformations. \figref{img:26} shows qualitative evaluation
results. As expected, feature descriptors such as SIFT
\cite{Lowe04}, DAISY \cite{Tola10}, BRIEF \cite{Calonder11}, and LSS
\cite{Schechtman07}, though using a powerful global optimization,
\emph{i.e.}, hierarchical dual-layer BP \cite{Liu11},
exhibit limitations on severe geometric variations,
while they provide robustness to some extent for
photometric variations. Our DASC descriptor in \figref{img:25}(k)
shows a better performance than other descriptors, but it also shows
the limitation for severe geometric variations. The GPM
\cite{Barnes10} had very low performance in terms of flow estimation
although it provides plausible warping results. The SID
\cite{Kokkinos08} have been proposed to provide geometric
robustness, but it is unable to address photometric variations.
Segmentation-aware description \cite{Trulls13} could improve the
matching accuracy of SIFT and SID for geometric variations, but it
also has limitation since it also reduces a discriminative power
of descriptor itself as shown in \figref{img:25}(g) and (h).
The DSP \cite{Kim13} provides limited performances,
since it just uses the SIFT with a fixed scale
and rotation. The SSF \cite{Qiu14} estimates visual
correspondence by repeatedly applying the SIFT on the scale-space
while enduring a huge computational complexity, but it still has
limitations in terms of computational complexity.
Contrarily, the GI-DASC descriptor optimized
by hierarchical dual-layer BP \cite{Liu11} provides the robustness for
both photometric and geometric deformations as shown in
\figref{img:25}(l). \figref{img:27} shows the average error rates on
DIML multi-modal benchmark. \vspace{-5pt}
\subsection{MPI Optical Flow Benchmark}\label{sec:66}
Optical flow methods typically assume only a
small displacement between consecutive frames.
Several approaches have been proposed to estimate a large displacement flow
vector \cite{Brox11}. However, motion blur and illumination
variation can degenerate the performance of these approaches. In
order to handle such challenging issues simultaneously, we applied
the DASC to the large displacement optical flow (LDOF)
approach \cite{Brox11}. It was evaluated on the MPI SINTEL
database \cite{Butler12} containing large non-rigid motion as well
as specular reflections, motion blur, and defocus blur. The dataset
consists of two kind of rendering frames, named clean and final
pass, and each set contains $12$ sequences with over $500$ frames in
total \cite{Butler12}. \tabref{tab:3} shows average end-point error
(EPE) results on MPI SINTEL. The DASC achieves a higher
gain, compared to other descriptors. \vspace{-5pt}
\begin{table}[!t]
	\caption{Comparison of average EPE on the MPI SINTEL \cite{Butler12}.}\label{tab:3}\vspace{-10pt}
	\centering
	\begin{tabular}{  >{\raggedright}m{0.26\linewidth}  >{\centering}m{0.09\linewidth}  >{\centering}m{0.14\linewidth}  >{\centering}m{0.09\linewidth}  >{\centering}m{0.14\linewidth} }
		\hlinewd{0.8pt}
		\multirow{2}{*}{} & \multicolumn{2}{ c }{Clean Pass} & \multicolumn{2}{ c }{Final Pass} \tabularnewline
		\cline{2-5}
		&\emph{all} &\emph{unmatched} &\emph{all} &\emph{unmatched} \tabularnewline
		\hline
		\hline
		Classic-NL \cite{Sun10} &7.940 &39.821 &9.439 &43.123 \tabularnewline
		LDOF \cite{Brox11} &7.180 &38.124 &8.422 &42.892 \tabularnewline
		LDOF+BRIEF \cite{Calonder11} &6.281 &37.841 &7.741 &41.875 \tabularnewline
		LDOF+LSS \cite{Schechtman07} &6.182\cellcolor[gray]{0.7} &37.514\cellcolor[gray]{0.7} &7.152\cellcolor[gray]{0.7} &40.332\cellcolor[gray]{0.7} \tabularnewline
		\hline
		\textbf{LDOF+DASC} &\textbf{5.578}\cellcolor[gray]{0.5} &\textbf{36.975}\cellcolor[gray]{0.5} &\textbf{6.384}\cellcolor[gray]{0.5} &\textbf{38.932}\cellcolor[gray]{0.5} \tabularnewline
		\hlinewd{0.8pt}
	\end{tabular}\vspace{-10pt}
\end{table}
\begin{figure}[t]
	\centering
	\renewcommand{\thesubfigure}{}
	\subfigure[(a) image 1]
	{\includegraphics[width=0.25\linewidth]{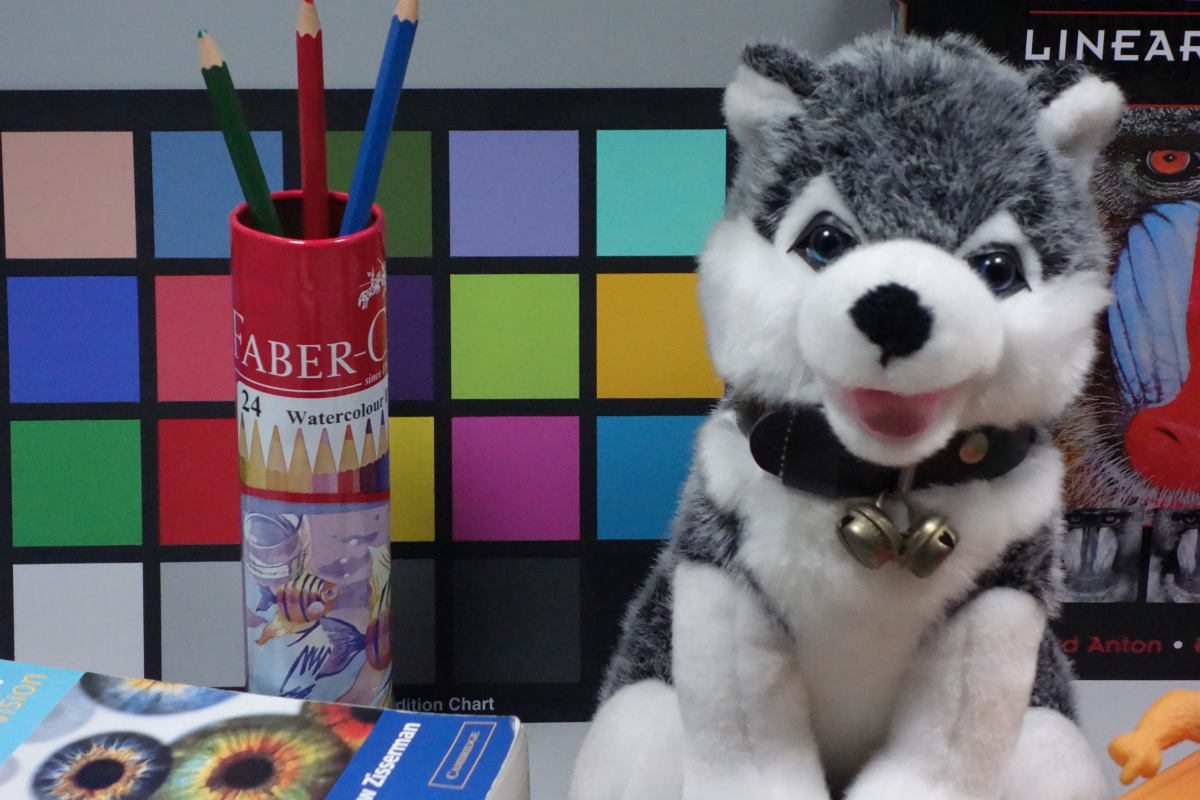}}\hfill
	\subfigure[(b) image 2]
	{\includegraphics[width=0.25\linewidth]{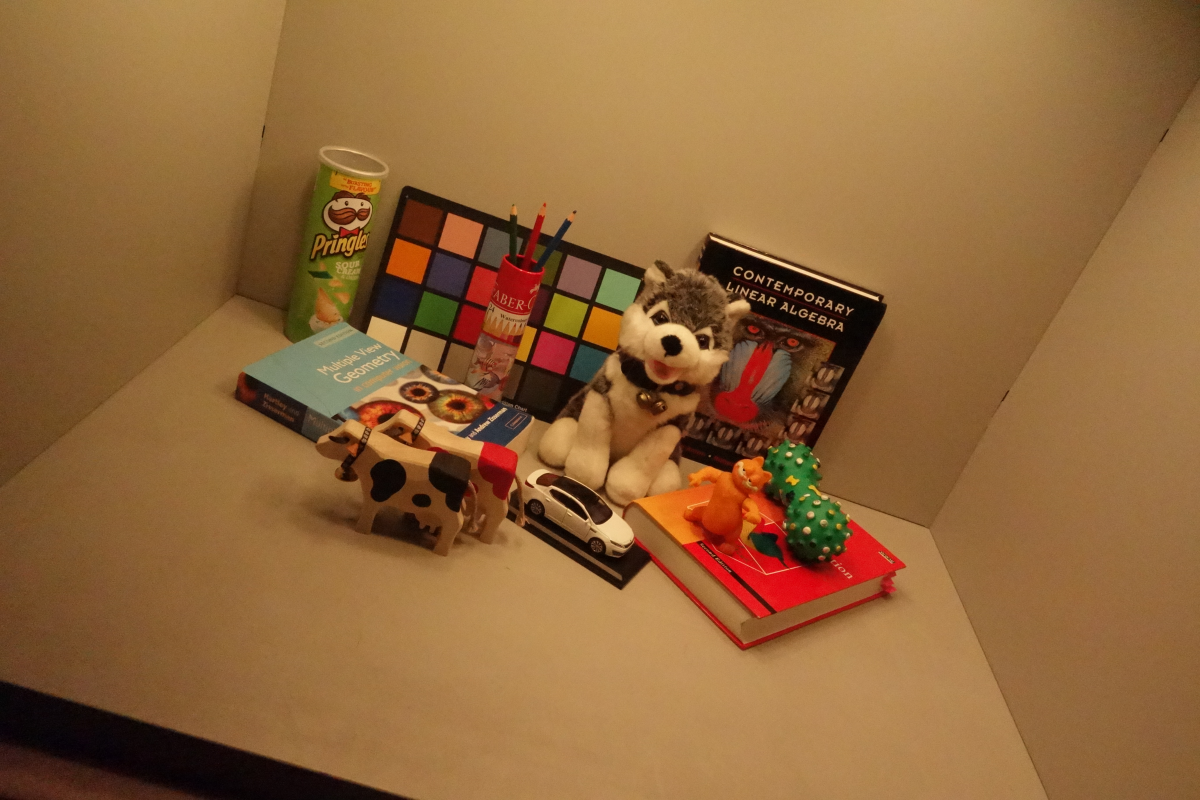}}\hfill
	\subfigure[(c) SSF \cite{Qiu14}]
	{\includegraphics[width=0.25\linewidth]{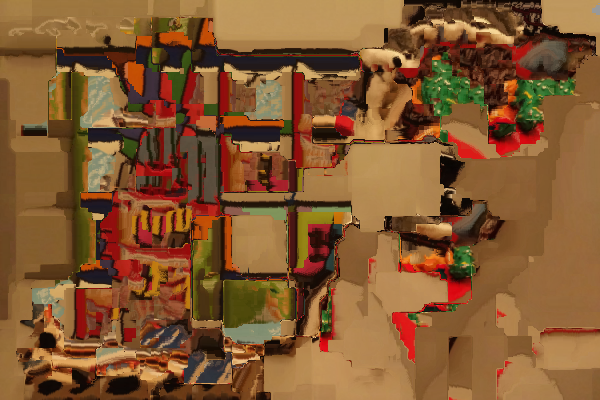}}\hfill
	\subfigure[(d) GI-DASC]
	{\includegraphics[width=0.25\linewidth]{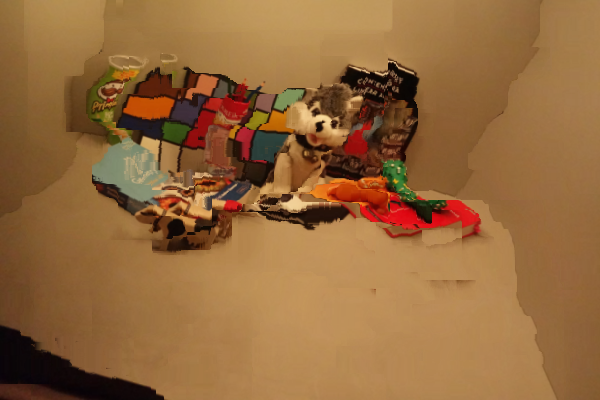}}\hfill
	\vspace{-10pt}
	\caption{Limitations for images under severe geometric variations.}\label{img:27}\vspace{-12pt}
\end{figure}
\subsection{Limitations}\label{sec:67}
Similar to \cite{Hassner12,Qiu14,Yang14}, our GI-DASC approximately determines a
relative scale using successive Gaussian smoothing, which
might work in only a limited range of scale variation as in \figref{img:27}. 
By leveraging an octave structure based on sub-sampling \cite{Lowe04}, 
a wider range of scale may be covered. \vspace{-5pt}
\section{Conclusion}\label{sec:7}
The robust novel dense descriptor called the DASC has been proposed
for dense multi-spectral and multi-modal correspondences. It
leverages an adaptive self-correlation measure and a randomized
receptive field pooling learned by linear discriminative learning.
Moreover, by making use of fast edge-aware filters, our DASC
descriptor is capable of computing the dense descriptor very
efficiently. In order to address geometric variations, the GI-DASC
descriptor also has been proposed by leveraging the efficiency and
effectiveness of the DASC through a superpixel-based representation.
The DASC and GI-DASC descriptor demonstrated its robustness in
establishing dense correspondence between challenging image pairs
taken under different modality conditions, \emph{e.g.}, RGB-NIR,
different illumination and exposure, flash-noflash, blurring
artifacts. We believe our method will serve as an essential tool for
several applications using multi-modal and multi-spectral
images.\vspace{-5pt}

\ifCLASSOPTIONcompsoc
\section*{Acknowledgments}

\else
\section*{Acknowledgment}

\fi

This work was supported by the National Research Foundation of Korea(NRF) grant funded by the Korea government(MSIP) (NRF-2013R1A2A2A01068338). \vspace{-5pt}

\ifCLASSOPTIONcaptionsoff
  \newpage
\fi

\bibliographystyle{IEEEtran}
\bibliography{egbib}

\end{document}